%% file: egbib.tex
\documentclass[final]{cvpr}

\usepackage{times}
\usepackage{epsfig}
\usepackage{graphicx}
\usepackage{amsmath}
\usepackage{amssymb}
\usepackage{pifont}%
\newcommand{\cmark}{\ding{51}}%
\newcommand{\xmark}{\ding{55}}%

\usepackage{booktabs}
\usepackage{tabularx}
\usepackage{colortbl}
\usepackage{multirow}
\usepackage[caption=false]{subfig}
\usepackage{comment}
\usepackage{xfrac}
\usepackage{hyphenat}
\usepackage{nicefrac}

\DeclareMathOperator{\atantwo}{atan2}
\def\360{\degree{360}}
\newcommand{\degree}[1]{${#1}^o$}

\definecolor{brightpink}{rgb}{1.0, 0.0, 0.5}
\definecolor{brightgreen}{rgb}{0.4, 1.0, 0.0}
\definecolor{cadmiumyellow}{rgb}{1.0, 0.96, 0.0}
\definecolor{canaryyellow}{rgb}{1.0, 0.94, 0.0}
\definecolor{caribbeangreen}{rgb}{0.0, 0.8, 0.6}
\definecolor{citrine}{rgb}{0.89, 0.82, 0.04}
\definecolor{chromeyellow}{rgb}{1.0, 0.65, 0.0}
\definecolor{classicrose}{rgb}{0.98, 0.8, 0.91}
\definecolor{cherryblossompink}{rgb}{1.0, 0.72, 0.77}
\definecolor{carnationpink}{rgb}{1.0, 0.65, 0.79}
\definecolor{candypink}{rgb}{0.89, 0.44, 0.48}

\definecolor{lightgreen}{RGB}{197,224,180}
\definecolor{lightblue}{RGB}{222,235,247}
\definecolor{lightpurple}{RGB}{238,229,241}
\definecolor{lightorg}{RGB}{251,229,214}
\definecolor{lightorange}{rgb}{1.0, 0.8, 0.6}

\newcommand{\first}[1]{\textbf{#1} \cellcolor{lightgreen}}
\newcommand{\second}[1]{#1 \cellcolor{lightblue}}
\newcommand{\third}[1]{#1 \cellcolor{lightpurple}}

\usepackage[export]{adjustbox}
\usepackage{sepfootnotes}

\definecolor{s3d_table}{RGB}{251,178,154}%
\definecolor{m3d_old_table}{RGB}{253,221,121}%
\definecolor{s3d}{RGB}{251,128,114}%
\definecolor{m3d_old}{RGB}{253,191,111}%
\definecolor{pano3d}{RGB}{117,112,179}%
\definecolor{gv2_medium}{RGB}{31,120,180}%
\definecolor{gv2_tiny}{RGB}{27,158,119}%
\definecolor{gv2_filmic}{RGB}{231,41,138}%
\definecolor{gv2_fullplus}{RGB}{179,222,105}

\newcommand*{\affaddr}[1]{#1} 
\newcommand*{\affmark}[1][*]{\textsuperscript{#1}}
\newcommand*{\email}[1]{\small{\texttt{#1}}}

\usepackage[pagebackref=true,breaklinks=true,colorlinks,bookmarks=false]{hyperref}

\def\cvprPaperID{12} %
\def\confYear{CVPR 2021}

\usepackage{nopageno}
\begin{document}

\title{Pano3D: A Holistic Benchmark and a Solid Baseline for \360 Depth Estimation}

\author{Georgios Albanis\thanks{Indicates equal contribution.} \! \affmark[1] %
Nikolaos Zioulis\footnotemark[1] \! \affmark[1,2] %
Petros Drakoulis\affmark[1] 
Vasileios Gkitsas\affmark[1] 
Vladimiros Sterzentsenko\affmark[1]
\\
Federico Alvarez\affmark[2]
Dimitrios Zarpalas\affmark[1]
Petros Daras\affmark[1]
\\
\affaddr{\affmark[1] Centre for Research and Technology Hellas, Thessaloniki, Greece} \\ \affaddr{\affmark[2] Universidad Polit\'{e}cnica de Madrid, Madrid, Spain}\\
\email{\{galbanis,nzioulis,petros.drakoulis,gkitsasv,vladster\}@iti.gr}\\
\email{fag@gatv.ssr.upm.es} \quad \email{\{zarpalas,daras\}@iti.gr}\\
\vspace{10pt}
\normalsize\href{https://vcl3d.github.io/Pano3D/}{vcl3d.github.io/Pano3D}
}

\maketitle

\begin{abstract}
   Pano3D is a new benchmark for depth estimation from spherical panoramas.
   It aims to assess performance across all depth estimation traits, the primary direct depth estimation performance targeting precision and accuracy, and also the secondary traits, boundary preservation and smoothness.
   Moreover, Pano3D moves beyond typical intra-dataset evaluation to inter-dataset performance assessment.
   By disentangling the capacity to generalize in unseen data into different test splits, Pano3D represents a holistic benchmark for \360 depth estimation.
   We use it as a basis for an extended analysis seeking to offer insights into classical choices for depth estimation.
   This results into a solid baseline for panoramic depth that followup works can built upon to steer future progress.
\end{abstract}

\section{Introduction}
\label{sec:introduction}
\input{Introduction.tex}

\section{Related Work}
\label{sec:related_work}
\input{RelatedWork.tex}

\section{Methodology}
\label{sec:methodology}
\input{Methodology.tex}

\section{Analysis}
\label{sec:results}
\input{Results.tex}

\section{Summary}
\label{sec:conclusion}
\input{Discussion}

\section*{Acknowledgements}
This work was supported by the EC funded H2020 project ATLANTIS [GA 951900].

{\small
\bibliographystyle{ieee_fullname}
\bibliography{egbib}
}

\appendix
\section{Supplementary Material}
This supplementary material complements our original manuscript with additional quantitative results, offering extra ablation experiments, providing qualitative results on real data and comparisons between the different architectures. 

\subsection{Quantitative Results}
Table~\ref{tab:Allloses} complements Table 1 of the main document, presenting the performance of all remaining metrics, namely the spherical direct depth metrics, the boundary preservation metrics, and the smoothness metrics.
In addition, Figure~\ref{fig:losses} presents the different models' performance in terms of three indicators, one for each trait.
These indicators combine an error and an accuracy metric:
\begin{align}
    i_d &= \frac{1}{(1 - \delta_{1.25}) \times RMSE},\\
    i_b &= \frac{1}{(1 - \nicefrac{(F_{0.25} + F_{0.5} + F_{1.0})}{3}) \times dbe^{acc}},\\
    i_s &= \frac{1}{(1 - \nicefrac{(\alpha_{11.25^o} + \alpha_{22.5^o} + \alpha_{30^o})}{3}) \times RMSE^o},
\end{align}
with $i_d, i_b,$ and $i_s$ the depth, boundary and smoothness performance indicators.
Evidently, UNet performs significantly better than the other models, especially in the boundary consistency metrics, while all models benefit of the addition of extra losses. 
The addition, of skip connections in a common ResNet architecture offers better performance.
While $\mathcal{L}_{grad}$ offers better depth performance for ResNet\textsubscript{\textit{skip}}, the variant trained with $\mathcal{L}_{comb}$ offers higher performance across the two secondary traits.

In addition, we complement the main's paper spherical metrics Table~\ref{tab:table2_oldandsperical} by collating the traditional ones for a straightforward comparison.

\input{Tables/Table1_supp}
\input{Tables/Table2_supp}
\input{Figures/sup_fig_1.tex}
\input{Tables/Table3_supp_updt}

Finally, Table~\ref{tab:lossesablation} reproduces the grounds upon our methodology was designed, namely the efficacy of pre-trained models \cite{ranftl2020towards} and the L1 loss \cite{carvalho2018regression}.
We use the DenseNet and Pnas models with the encoders initialized using weights pre-trained on ImageNet.
Both claims stand, with all pre-trained models achieving better performance than the model trained from scratch.
In addition, the L1 loss outperforms both berHu \cite{laina2016deeper} and log loss. 
Interestingly, the performance drops significantly in DenseNet when trained with other losses, while for Pnas the performance gap is smaller.
Therefore, when benchmarking different models, this needs to be taken into account as well.
Only through consistent experimentation across different aspects measurable and explainable progress will be possible.

\subsection{Qualitative Results}
Finally we present additional qualitative results for different models.
Apart from the collation of the predicted depth maps between the different models, we provide an advantage visualisation technique similar to that presented in HoHoNet \cite{sun2020hohonet}. 
The visualisation is the MAE difference between two comparable models. 

To that end, Figure~\ref{fig:adv_res_reskip} demonstrates the comparison of ResNet and ResNet\textsubscript{\textit{skip}} architectures, Figure~\ref{fig:adv_unet_pnas} that of the UNet and Pnas architectures, and, finally Figure~\ref{fig:adv_unet_reskip} presents the differences between the UNet and ResNet\textsubscript{\textit{skip}} architectures.

Additionally, Figure~\ref{fig:edges} presents comparative results regarding the boundary preservation performance across models. 
Once again, UNet is able to capture finer-grained details while the Pnas model produces smoother results. Similarly, the differences between ResNet and ResNet\textsubscript{\textit{skip}}, attributed to the addition of the skip connections are apparent across all samples. 

Nonetheless, Pnas better captures the global context as seen in Figure~\ref{fig:unet_mesh}, where the scene's dominant planar surfaces are better preserved by it than UNet. 

\begin{figure*}[!htbp]
\includegraphics[]{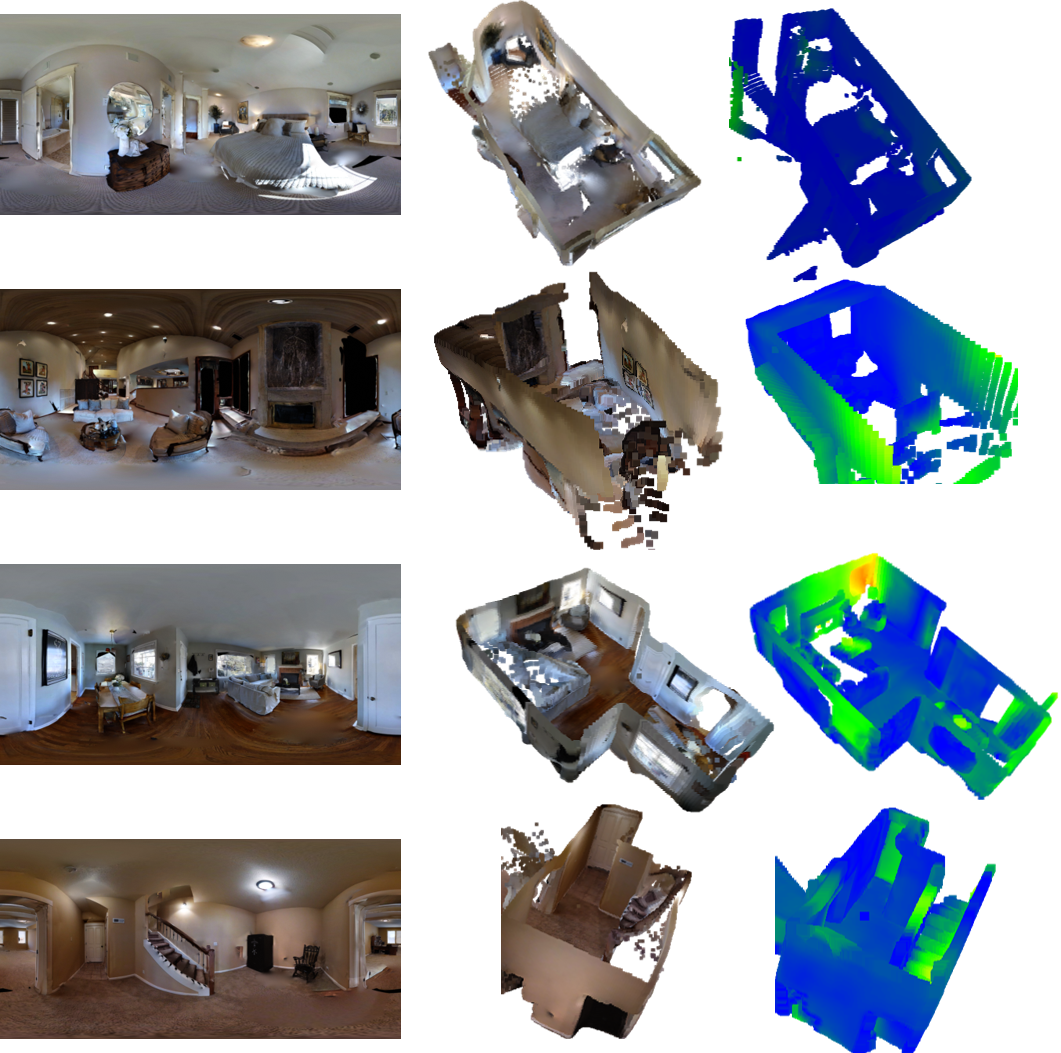}
\caption{UNet qualitative results. 
From left to right: \textbf{i)} Input color panorama, \textbf{ii)} colored predicted point cloud, and \textbf{iii)} heatmap visualization of the \textit{c2c} error on the ground truth point cloud.} 
\label{fig:unet_qual}
\end{figure*}

\begin{figure*}[!htbp]
\includegraphics[]{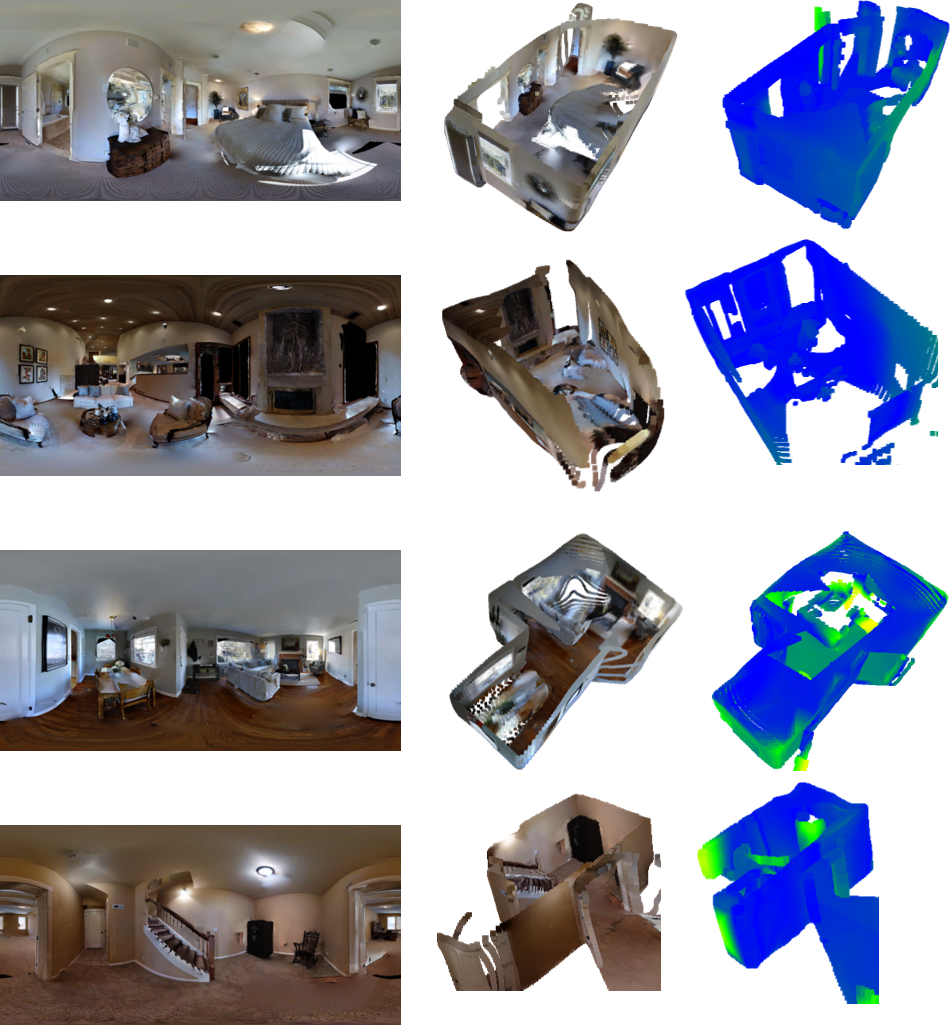}
\caption{Pnas qualitative results.
From left to right: \textbf{i)} Input color panorama, \textbf{ii)} colored predicted point cloud, and \textbf{iii)} heatmap visualization of the \textit{c2c} error on the ground truth point cloud.} 
\label{fig:pnas_qual}
\end{figure*}

\begin{figure*}[!htbp]
\includegraphics[]{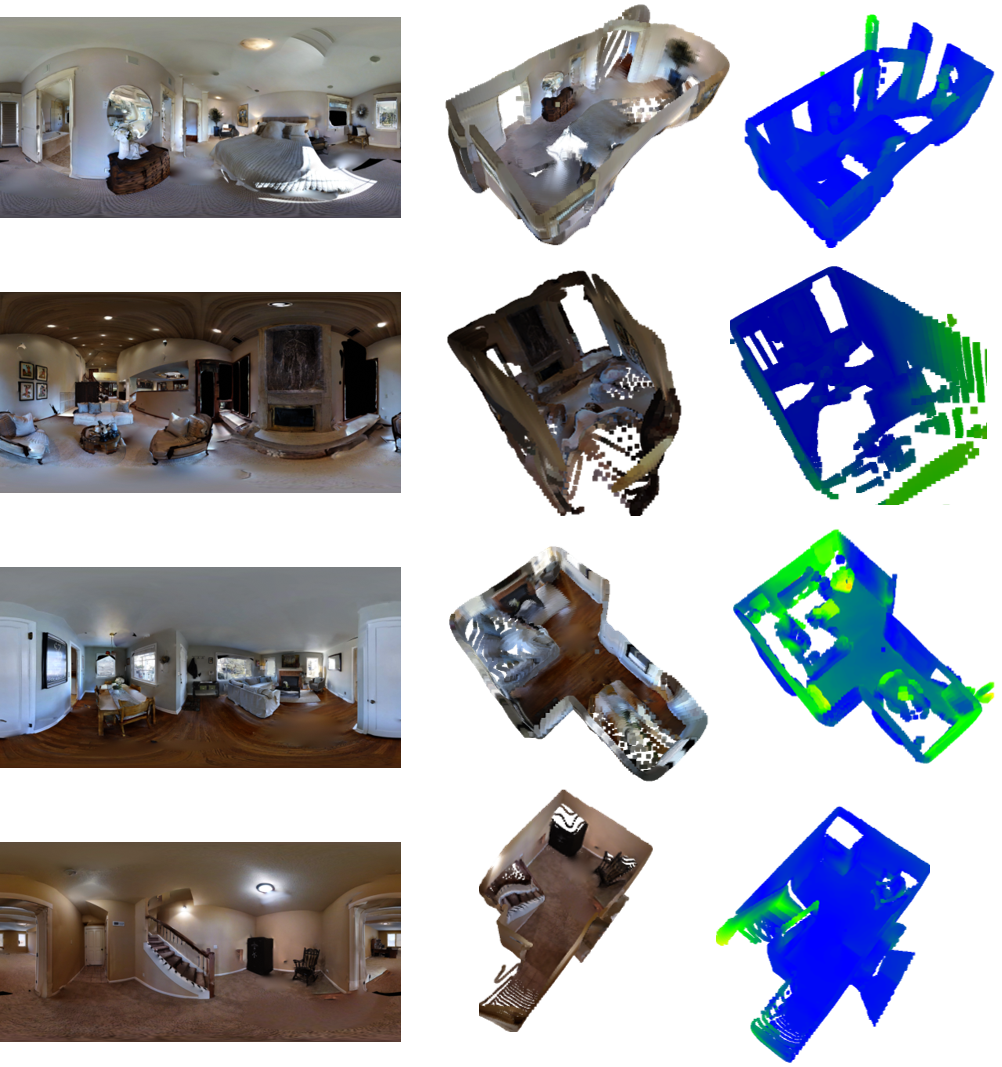}
\caption{ResNet\textsubscript{skip} qualitative results.
From left to right: \textbf{i)} Input color panorama, \textbf{ii)} colored predicted point cloud, and \textbf{iii)} heatmap visualization of the \textit{c2c} error on the ground truth point cloud.} 
\label{fig:skip_qual}
\end{figure*}

Figures~\ref{fig:unet_qual}, \ref{fig:pnas_qual}, \ref{fig:skip_qual} demonstrate qualitative results in GV2 \textit{tiny} split for the UNet, Pnas, and ResNet\textsubscript{\textit{skip}} architectures respectively.
Apart from the predicted point cloud we visualise the \textit{c2c} error on the ground truth point cloud, with a blue-green-red colormap denoting the error's magnitude.

\begin{figure*}[!htbp]
\includegraphics[scale=0.35]{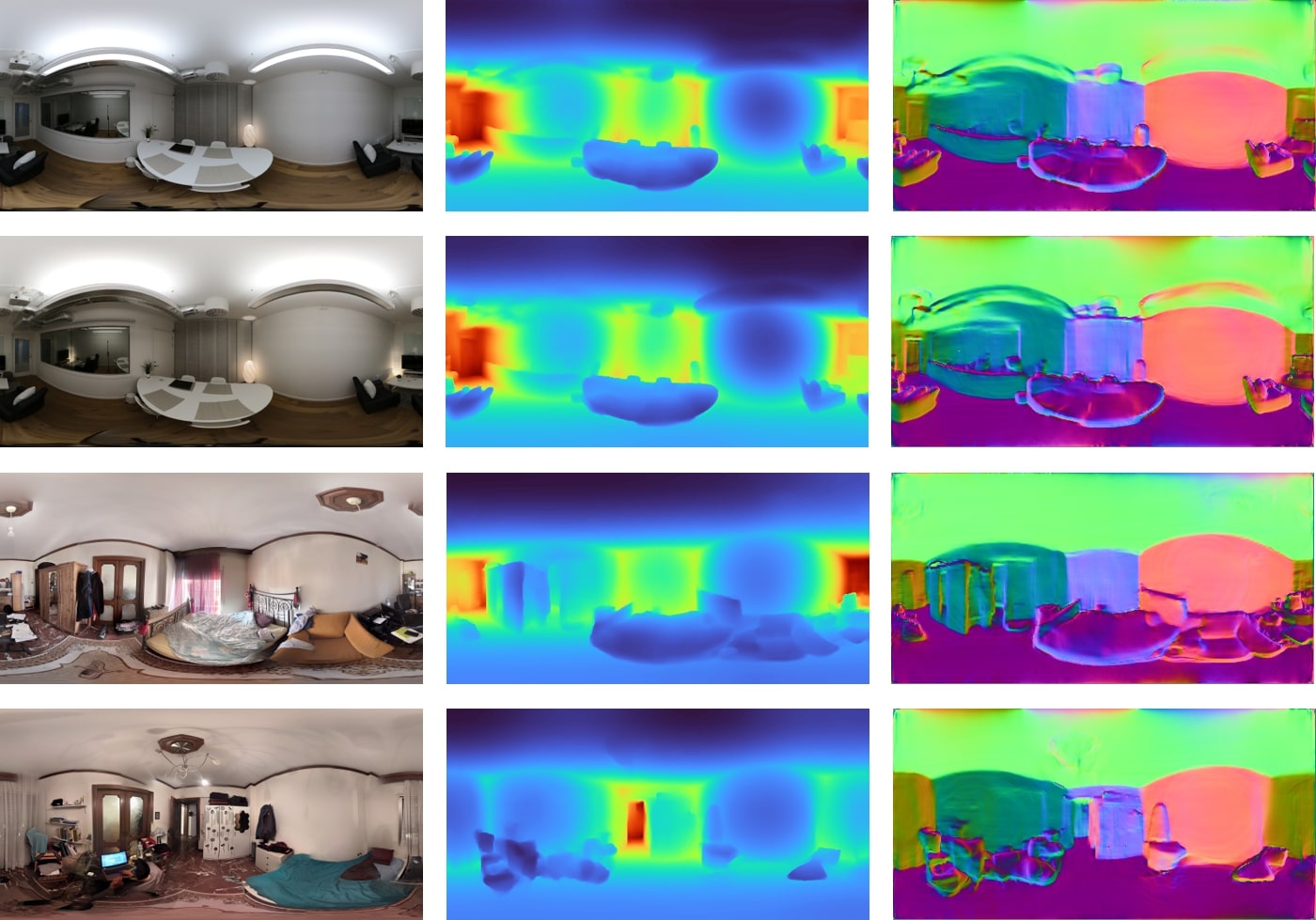}
\caption{Qualitative results using the UNet model applied to in-the-wild real data captures.
The top two rows are captures with a \360 camera, while the bottom two rows are stitched panoramas from a mobile phone.
From left to right: \textbf{i)} Input color panorama, \textbf{ii)} predicted depth, and \textbf{iii)} normals derived from the predicted depth.} 
\label{fig:inthewildnormals}
\end{figure*}

Finally, Figures~\ref{fig:inthewildnormals} and \ref{fig:inthewild} offer qualitative results of our best performing method in real world, in-the-wild, data captures. 
We also qualitatively compare our predictions with a state-of-the-art \360 depth estimation model (i.e. BiFuse \cite{wang2020bifuse}). 
It is worth highlighting that even the two of the three \360 images are captured by a panorama camera, the last two images are captured by a smartphone camera, and as such there are artifacts. 
Yet, it seems that this does not greatly affect the performance of models.
The UNet model produces higher quality depth estimates than BiFuse, albeit trained only on the train split of M3D, while the publicly available BiFuse model, as reported in UniFuse \cite{jiang2021unifuse}, has been trained on the \textit{entire} M3D dataset. 

\input{Figures/adv_res_reskip}
\input{Figures/adv_unet_pnas}
\input{Figures/adv_unet_reskip}
\input{Figures/supp/edges/edgescomparisonupdt}
\input{Figures/supp/unet_mesh}

\input{Figures/supp/inthewild}

\end{document}

%% file: Introduction.tex
Benchmarks are the drivers of progress as they facilitate measurable technical increments, and can also provide explainable insights for diverging technical approaches.
They must be unbiased, especially given the emergence of data-driven methods that can easily exploit any hidden bias in the data.
Besides, the expressiveness of deep models necessitates the enrichment of benchmarks with varying data distributions to allow for the assessment of their generalization and their capacity to exploit different sources of data.

\input{Figures/intro}

The recent availability of \360 depth datasets out of stitched raw sensor data \cite{armeni2017joint,chang2018matterport3d}, 3D reconstruction renderings \cite{zioulis2018omnidepth,zioulis2019spherical}, and photorealistic synthetic scenes \cite{jin2020geometric,Structured3D} has stimulated research in monocular \360 depth estimation \cite{tateno2018distortion,eder2019mapped,wang2020bifuse,jiang2021unifuse,zeng2020joint,sun2020hohonet}.
Still, the progress in monocular depth estimation has been mainly driven by research for traditional cameras, and assessed on perspective datasets, starting with the pioneering work of \cite{eigen2014depth}.
Even though other approaches exist (\textit{e.g.}~ordinal regression \cite{fu2018deep,bhat2020adabins}), depth estimation is most typically addressed as a dense regression objective.
Various estimator choices are available for the direct objective like L1, L2, or robust versions like the reverse Huber (berHu) loss \cite{laina2016deeper}.
Complementary errors have also been introduced like the virtual normal loss \cite{yin2019enforcing} which captures longer range depth relations.
Additional smoothness ensuring losses can be used to enforce a reasonable and established prior of depth maps, which is their piece-wise smoothly spatially varying nature \cite{huang2000statistics}.

Depth maps also exhibit sharp edges at object boundaries \cite{huang2000statistics}, whose preservation is important for various downstream applications.
Recent works which focus explicitly on improving the estimated boundaries introduced new metrics to measure boundary preservation performance \cite{hu2019revisiting,ramamonjisoa2020predicting}.
Since convolutional data-driven methods spatially downscale the encoded representations, predicting neighboring values relies on neighborhood information, leading to interpolation blurriness.
Counteracting approaches, like encoder-decoder skip connections or guided filters \cite{wu2018fast}, can lead to texture transfer artifacts, hurting the predictions' smoothness.
The latter (smoothness) is also an important trait for some tasks like scene-scale 3D reconstruction which usually relies on surface orientation information \cite{kazhdan2013screened} to preserve structural planarities, while the former (boundaries) are necessary for applications like view synthesis \cite{attal2020matryodshka} or object retrieval \cite{karsch2013boundary}.
Still, smoothness related metrics are usually presented on surface \cite{wang2020vplnet,karakottas2019360} or plane \cite{eder2019pano} estimation works.
Further, the balance between them needs to be tuned as they are conflicting objectives.

In this work we set to deliver an unbiased and holistic benchmark for monocular \360 depth estimation that provides performance analysis across all traits, \textbf{i)} depth estimation, \textbf{ii)} boundary preservation, \textbf{iii)} smoothness.
We also consider an orthogonal evaluation strategy that seeks to assess the models' generalization as well, across its different facets, \textbf{i)} varying depth distributions, \textbf{ii)} adaptation to the scenes' contexts, and \textbf{iii)} different camera domains.
To support the benchmark, we design a set of solid baselines that respect best practises as reported in the literature and rely on standard architectures.
Our results, data, code, configurations and trained models are publicly available at \href{https://vcl3d.github.io/Pano3D/}{vcl3d.github.io/Pano3D/}.

\vspace{-0.15cm}
\begin{itemize}
    \item We show that recently made available datasets contain significant biases or artifacts that prevent them from being suitable as solid benchmarks.
    \vspace{-0.15cm}
    \item We provide \360 depth estimation performance results for all different traits, across different domains, contexts, distributions and resolutions, while also taking depth refinement advances into account.
    \vspace{-0.15cm}
    \item We demonstrate the effectiveness of skip connections, a rare architectural choice for (\360) depth estimation.
\end{itemize}

%% file: Figures/intro.tex
\begin{figure}[!htbp]
    \includegraphics[width=\linewidth]{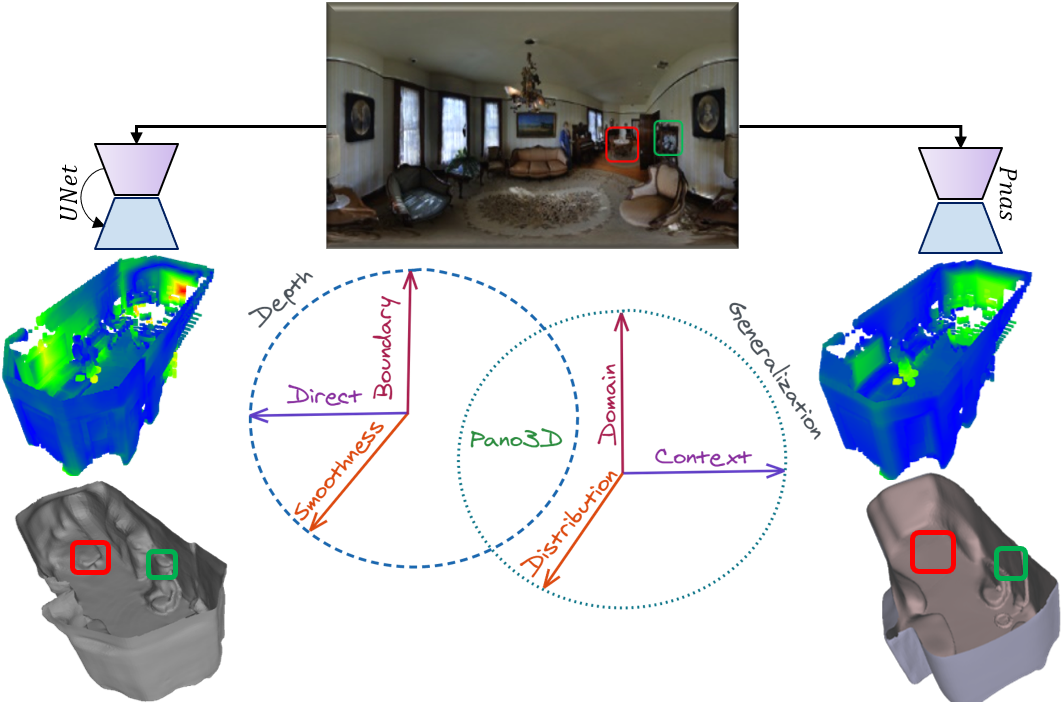}
    \caption{Preserving depth's piece-wise smoothness should be the primary goal of data-driven depth estimation models.
    Yet most works only assess direct depth performance neglecting secondary traits like smoothness or boundary preservation.
    Different architectures (UNet -- left, or Pnas -- right) exhibit different inference characteristics skewed towards boundaries (UNet) or smoothness (Pnas).
    The Pano3D benchmark descends from a holistic perspective taking into account all performance traits, and additionally considers an orthogonal performance assessment, generalization to unseen data from different distributions, contexts or domains. }
    \label{fig:teaser}
    \vspace{-0.45cm}
\end{figure}

%% file: RelatedWork.tex
\textbf{Monocular Omnidirectional Depth Estimation.}
\label{sec:related_omnidirectional}
\input{OmniDirectionalDepth.tex}

\input{Figures/datasetfig}

\textbf{Monocular Perspective Depth Estimation.}
\label{sec:related_perspective}
\input{PerspectiveDepth.tex}

%% file: OmnidirectionalDepth.tex
The first works addressing the monocular data-driven omnidirectional depth estimation task were \cite{tateno2018distortion} and \cite{zioulis2018omnidepth}.
The former applied traditional CNNs trained on perspective images in a distortion-aware manner to spherical images, while the latter introduced a rendered spherical dataset of paired color and depth images, in addition to a simplistic rectangular filtering preprocessing block.
Pano Popups \cite{eder2019pano} simultaneously predict depth and surface orientation to construct planar 3D models, showing the insuffiency of depth estimates along to approximate planar regions.

The generalized Mapped Convolutions \cite{eder2019mapped} were applied to omnidirectional depth estimation, showing how accounting for the distortion when using equirectangular projection increases performance in the image regions closer to the equator.
Although these spatially imbalanced predictions are an important issue to address for \360 depth estimation methods, the usual evaluation methodologies do not address this apart from \cite{zioulis2019spherical}.
The omnidirectional extension networks \cite{cheng2020omnidirectional} employ a near field-of-view (NFoV) perspective depth camera to accompany the spherical one, offering a necessary, albeit not full FoV, constraint 
to enhance the preservation of details in the inferred depth map.

Recent omnidirectional depth estimation works diverged in two paths.
One route is to exploit the nature of the spherical images within network architectures, with BiFuse \cite{wang2020bifuse} fusing features from a cubemap and an equirectangular representation, while UniFuse \cite{jiang2021unifuse} shows that the fusion of cubemap features to the equirectangular ones is more effective.
HoHoNet \cite{sun2020hohonet} adapts classical CNNs to operate on \360 images by flattening the meridians to DCT coefficients, allowing for efficient dense feature reconstruction, and applying it to monocular depth estimation from spherical panoramas.
Other recent works \cite{jin2020geometric,zeng2020joint} explore the connection between the layout and depth estimation tasks, while \cite{feng2020deep} relies on the joint optimization between depth and surface orientation estimates using a UNet model \cite{ronneberger2015u}.

%% file: Figures/datasetfig.tex
\begin{figure*}[ht!]
  \def\mycolspace{0.5mm}
  \def\datasetheight{1.45cm}
  \centering
\begin{tabular}{@{}c@{\hspace{\mycolspace}}c@{\hspace{\mycolspace}}c@{\hspace{\mycolspace}}c@{\hspace{\mycolspace}}c@{\hspace{\mycolspace}}c@{}}
\includegraphics[height=\datasetheight,clip]{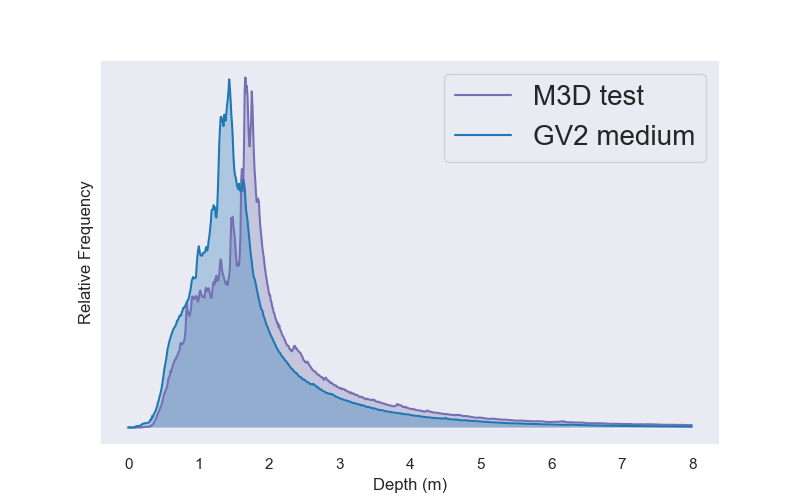} &
\includegraphics[height=\datasetheight,clip,cfbox=s3d 1.1pt 0pt]{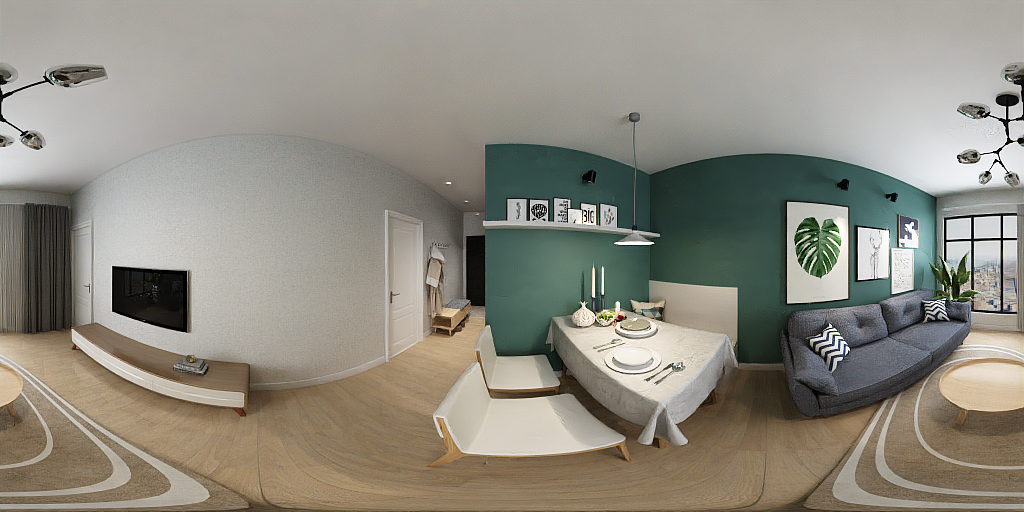} &
\includegraphics[height=\datasetheight, clip,cfbox=gv2_medium 1.1pt 0pt]{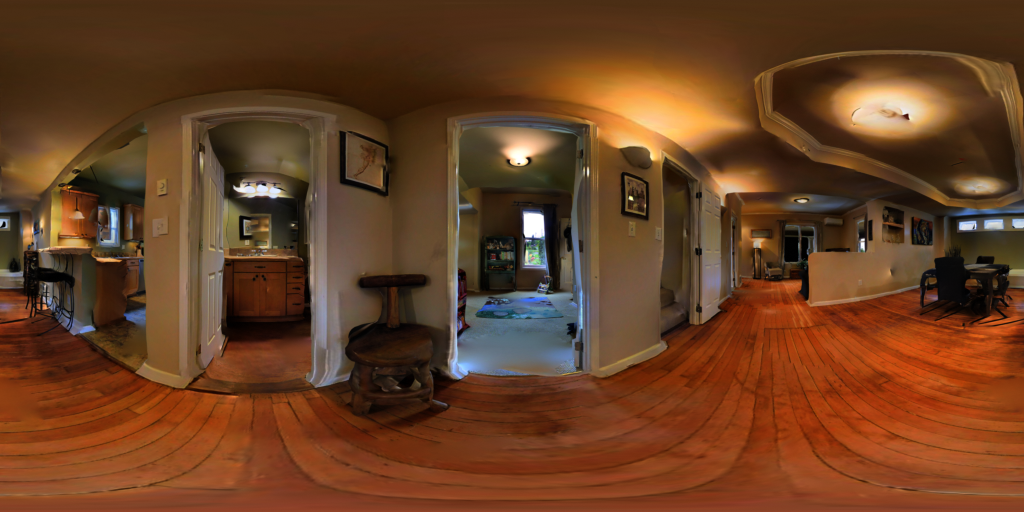} & 
\includegraphics[height=\datasetheight, clip,cfbox=gv2_fullplus 1.1pt 0pt]{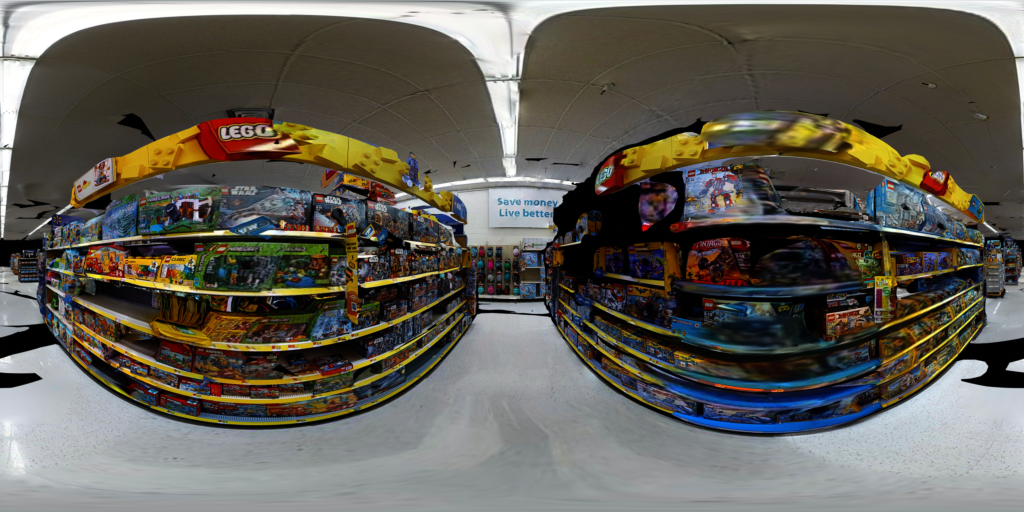} &
\includegraphics[height=\datasetheight,clip, cfbox=gv2_tiny 1.1pt 0pt]{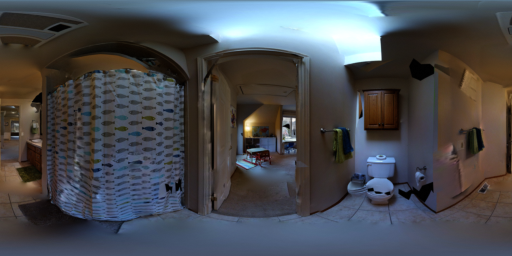} &
\includegraphics[height=\datasetheight, clip,cfbox=gv2_filmic 1.1pt 0pt]{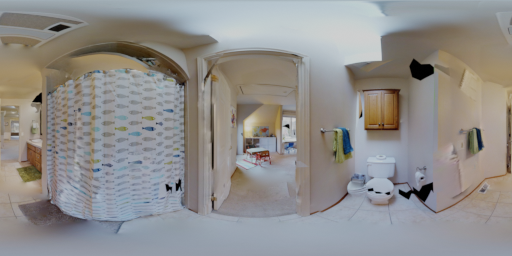}  \\
\includegraphics[height=\datasetheight,clip]{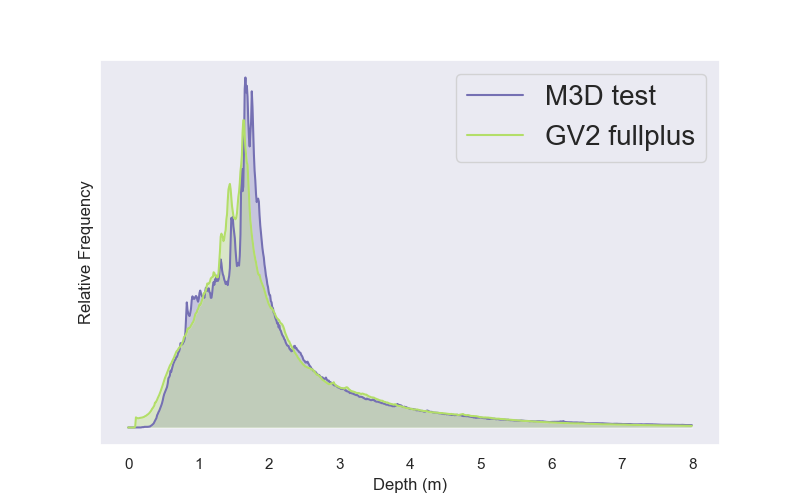} &
\includegraphics[height=\datasetheight,clip, cfbox=m3d_old 1.1pt 0pt]{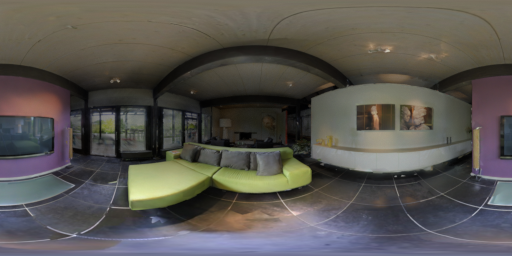} &
\includegraphics[height=\datasetheight,clip, cfbox=gv2_medium 1.1pt 0pt]{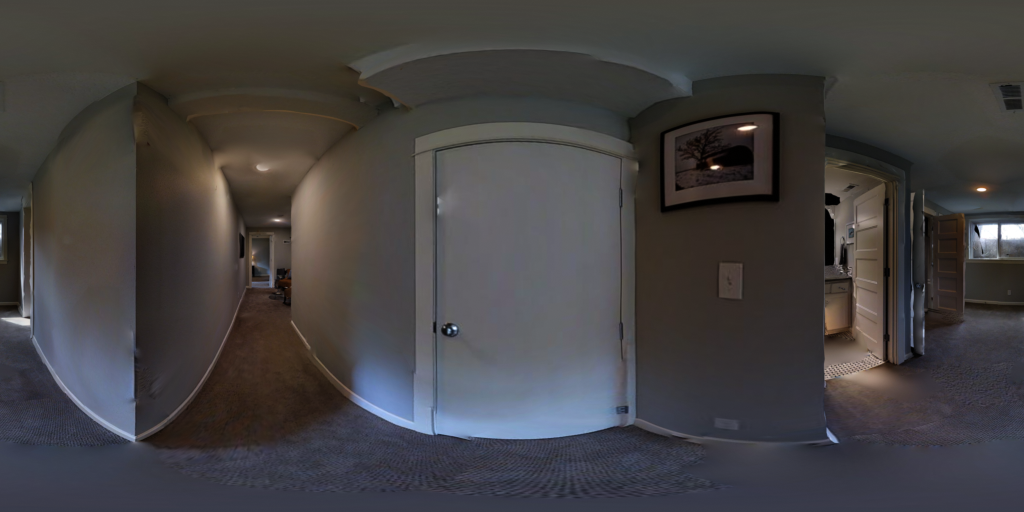} &
\includegraphics[height=\datasetheight,clip, cfbox=gv2_fullplus 1.1pt 0pt]{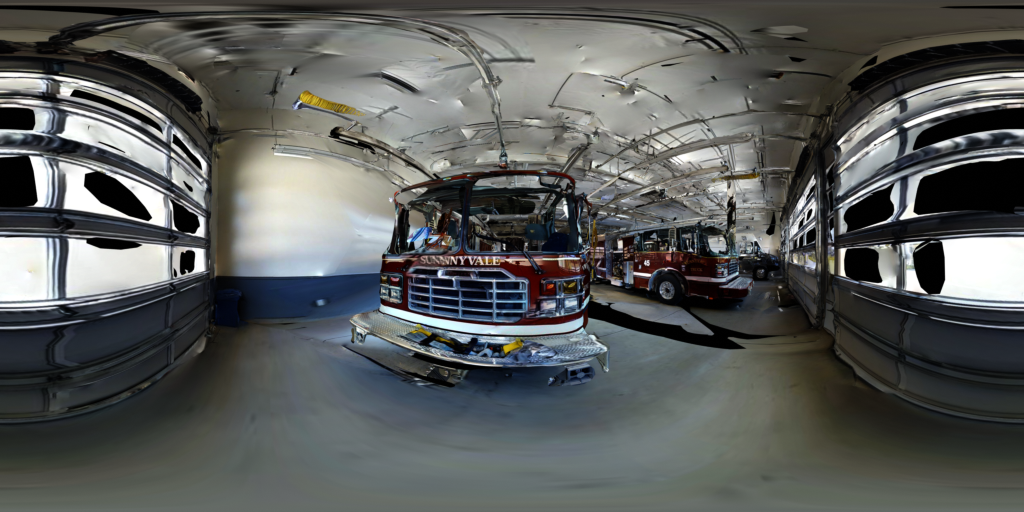} &
\includegraphics[height=\datasetheight,clip, cfbox=gv2_tiny 1.1pt 0pt]{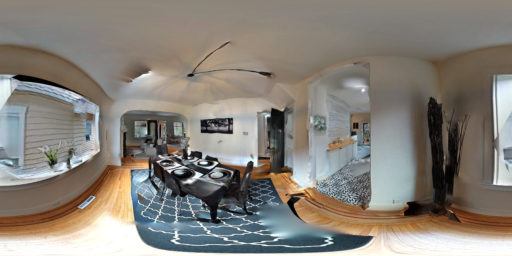} &
\includegraphics[height=\datasetheight,clip, cfbox=gv2_filmic 1.1pt 0pt]{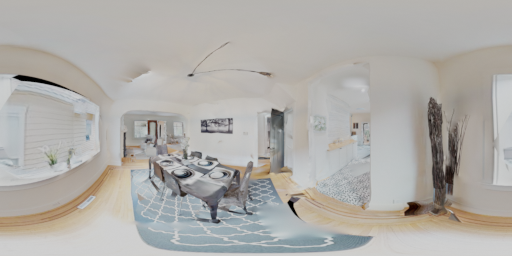} \\
\includegraphics[height=\datasetheight,clip]{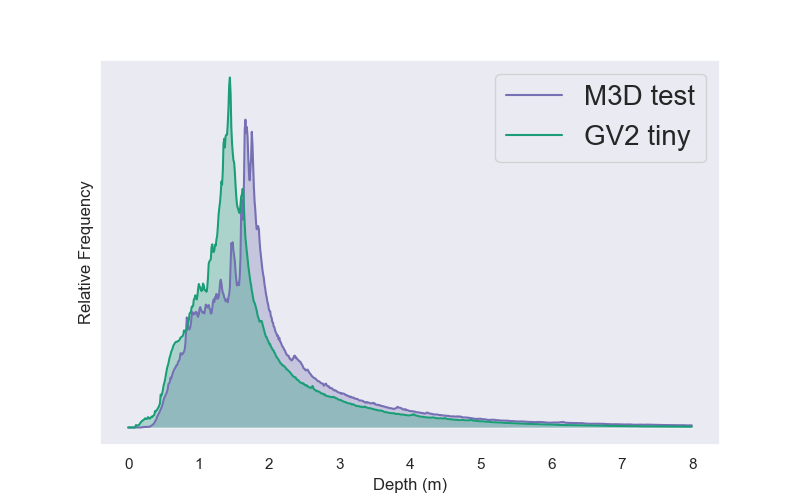} &
\includegraphics[height=\datasetheight,clip, cfbox=pano3d 1.1pt 0pt]{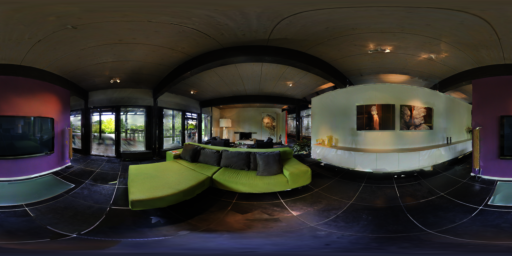} &
\includegraphics[height=\datasetheight,clip, cfbox=gv2_medium 1.1pt 0pt]{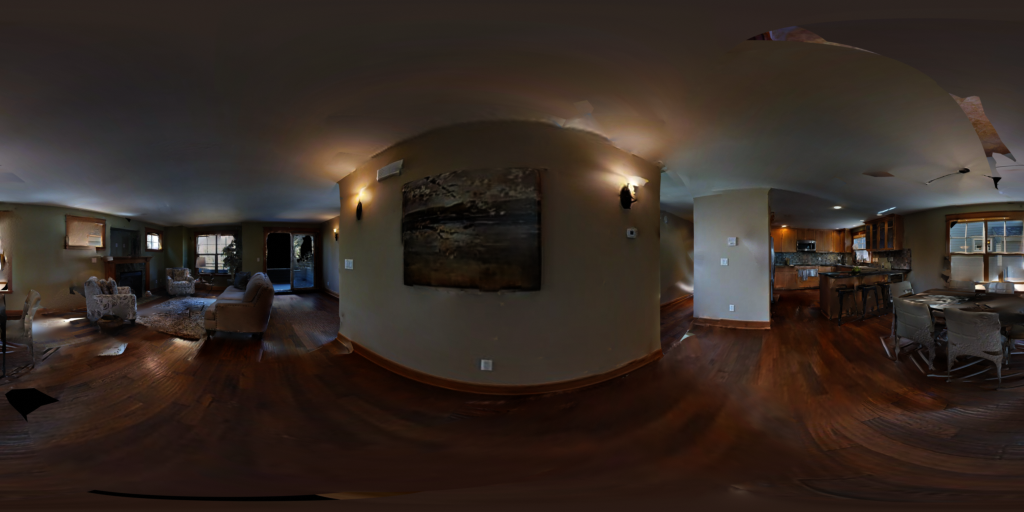} &
\includegraphics[height=\datasetheight,clip, cfbox=gv2_fullplus 1.1pt 0pt]{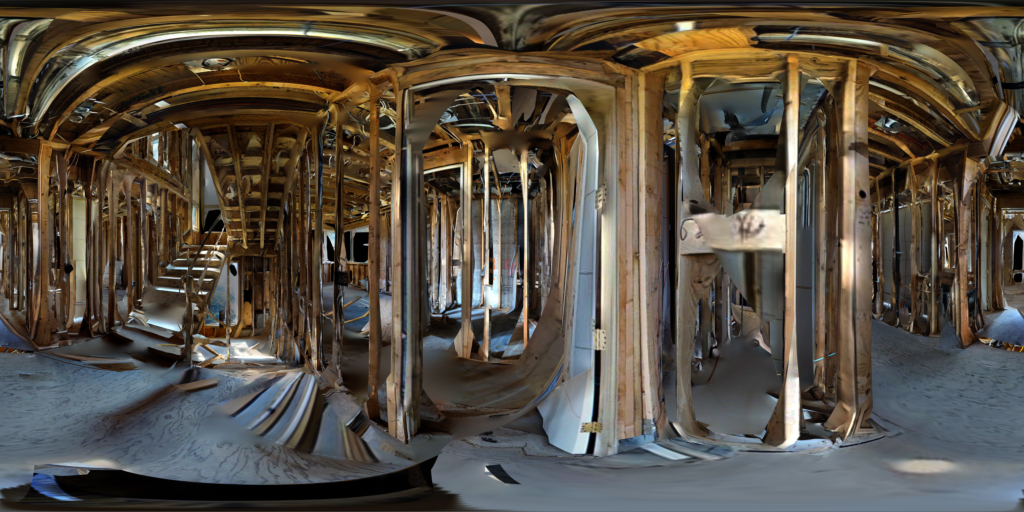} &
\includegraphics[height=\datasetheight,clip, cfbox=gv2_tiny 1.1pt 0pt]{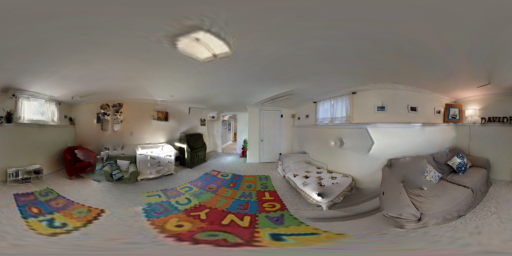} &
\includegraphics[height=\datasetheight,clip, cfbox=gv2_filmic 1.1pt 0pt]{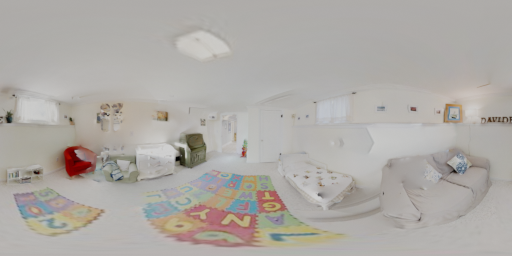}
\end{tabular}
\caption{Existing \360 depth benchmarks are either synthetic (Structured3D -- \emph{\color{s3d}\textbf{tomato red}}) or biased (3D60, with an extra light source --  \emph{\color{m3d_old}\textbf{orange}}).
In addition to a holistic evaluation scheme, our benchmark Pano3D fixes the lighting issues (Matterport3D data -- \emph{\color{pano3d}\textbf{violet}}), and additionally offers a variety of zero-shot cross dataset transfer splits (generated from GibsonV2), permitting the validation of close-to-real-world performance.
The \textit{tiny} (\emph{\color{gv2_tiny}\textbf{dark green}}) and \textit{medium} (\emph{\color{gv2_medium}\textbf{dark blue}}) splits contain residential building scenes but with a different depth distribution as presented on the left of the figure.
The \textit{fullplus} (\emph{\color{gv2_fullplus}\textbf{light green}}) split, albeit of similar distribution, contains different context scenes, like super-markers, garages and construction sites.
Finally, we additionally generate camera domain adapted splits like the \textit{tiny filmic} (\emph{\color{gv2_filmic}\textbf{purple}}) on the right, or the \textit{fullplus filmic} (not shown), effectively capturing all different generalization performance traits.
}
 \label{fig:dataset}
 %\vspace{-0.35cm}
\end{figure*}

%% file: PerspectiveDepth.tex
The pioneering work for data-driven monocular dense depth estimation \cite{eigen2014depth} employed a scale-invariant loss and established the set of metrics used to evaluate follow up works.
Naturally, the progress in monocular depth estimation for perspective images is larger, as traditional images find more widespread use.
While impressive gains have been presented using ordinal regression \cite{fu2018deep} or adaptive binning \cite{bhat2020adabins}, they have not been applied to \360 depth estimation, which exhibits more complex depth distributions than perspective depth maps due to its holistic FoV.

Results like the berHu loss presented in \cite{laina2016deeper} have found traction in omnidirectional models as they are more easily transferable.
On the contrary the more recently presented virtual normal loss \cite{yin2019enforcing} has not been applied to \360 depth, albeit its longer range depth relation modelling is highly aligned with the global reasoning required for the spherical task.
Recently, the balance between the multitude of losses required to balance smoothness, boundary preservation and depth accuracy were investigated in \cite{lee2020multi} to help models initially focus on easier to optimize losses (\textit{i.e.}~depth accuracy), and then on harder ones (\textit{i.e.}~smoothness, boundary).

Regarding depth discontinuity preservation performance, \cite{hu2019revisiting} showed that a combination of three different loss terms, a depth, a surface and a spatial derivative one, help increase performance at object  boundaries.
Similarly, a boundary consistency was introduced in \cite{huang2019indoor} to overcome blurriness and bleeding artifacts.
Another approach, based on learnable guided filtering \cite{wu2018fast}, exploits the color image as guidance.
Recently, displacement fields \cite{ramamonjisoa2020predicting} showed that predicting resampling offsets instead of residuals is more suitable to increase performance at sharp depth discontinuities, while preserving depth estimation accuracy.

%% file: Methodology.tex
Our goal is two-fold, first to deliver a new benchmark for \360 depth estimation, and second, to methodically analyze the task in light of recent developments, to identify a set of solid baselines, which future works will use as the starting points for assessing performance gains.
Section~\ref{sec:methodology_dataset} introduces the benchmark data, which set the ground for the subsequent analysis.
Section~\ref{sec:methodology_metrics} describes the benchmark's holistic approach in terms of evaluation, while Section~\ref{sec:methodology_experimental_setup} presents the experiment design rationale.

\subsection{Dataset}
\label{sec:methodology_dataset}
\input{Dataset.tex}

\input{Tables/Table1}

\subsection{Metrics}
\label{sec:methodology_metrics}
\input{Metrics.tex}

\subsection{Experimental Setup}
\label{sec:methodology_experimental_setup}
\input{ExperimentalSetup.tex}

%% file: Dataset.tex
Up to now \360 depth datasets either rendered purely synthetic scenes like Structured3D \cite{Structured3D} or 360D's SunCG and SceneNet parts \cite{zioulis2018omnidepth}, or relied on 3D scanned datasets like Matterport3D \cite{chang2018matterport3d} and Stanford2D3D \cite{armeni2017joint}.
The latter offer both panoramas and the 3D textured meshes, with some works using the original Matterport camera derived data and others the rendered panoramas from the 3D scanned meshes.
Both approaches come with certain drawbacks, the original data contain invalid (\textit{i.e.}~true black) regions towards the sphere's poles, while the 3D rendered data contain invalid regions where the 3D scans failed to reconstruct the surface.
At the same time, the original data present with stitching artifacts (mostly blurring), while the rendered data sometimes suffer from 3D reconstruction errors which manifest in color discontinuities.
We opt for the generation via rendering approach \cite{zioulis2018omnidepth} as it produces true spherical panoramas and higher quality depth maps at lower resolution compared to nearest neighbor sampling.
However, we fix a critical issue of the 360D \cite{zioulis2018omnidepth} and 3D60 \cite{zioulis2019spherical} datasets, namely, the introduction of a light source that alters the scene's photorealism.
Instead, we only sample the raw diffuse texture, preserving the original scene lighting, a crucial factor for unbiased learning and performance evaluation.

\textbf{Zero-shot Cross-Dataset Transfer.}
However, there is a need to move beyond traditional train/test split performance analysis to support model deployment in real-world conditions. 
Thus, assessing generalization performance is very important.
Towards that end, apart from re-rendering the Matterport3D scans for training, we introduce a new \360 color-depth pair generated dataset from the $572$ GibsonV2 (GV2) \cite{xia2018gibson} 3D scans.
Compared to Matterport3D's (M3D) $90$ buildings, it is a vastly larger dataset, whose scenes offer higher variety as well.
These renders can be used for assessing generalization performance across its different splits: \textit{tiny}, \textit{medium}, \textit{full}\footnote{The larger GV2 \textit{full} split is kept for future training purposes} %
and \textit{fullplus}.
After removing outlier scans and filtering samples (keeping those with $\leq 10\%$ invalid pixels), we are left with $7170$/$1527$ train/test M3D samples, and $2740$, $6999$, $3284$, $21203$ GV2 \textit{tiny}, \textit{medium}, \textit{fullplus}, and \textit{full} split samples respectively.

%% file: Tables/Table1.tex
\begin{table*}[!htbp]
\caption{Direct depth metrics performance across models and supervision schemes.
Best three performers are denoted with bold faced \colorbox{lightgreen}{\textbf{light green}} (1\textsuperscript{st}), \colorbox{lightblue}{light blue} (2\textsuperscript{nd}) and \colorbox{lightpurple}{light purple} (3\textsuperscript{rd}) respectively following the ranking order.
Same scheme applies to all tables.}
\label{tab:losses}
\footnotesize
\centering
\begin{tabular}{ll|cccc|ccccc}
\hline
\multicolumn{2}{c|}{\multirow{2}{*}{Model}}                                          & \multicolumn{4}{c|}{Depth Error $\downarrow$}                         & \multicolumn{5}{c}{Depth Accuracy $\uparrow$}                                                  \\
\multicolumn{2}{c|}{}                                                                & \textit{RMSE}   & \textit{RMSLE}  & \textit{AbsRel} & \textit{SqRel}  & $\delta_{1.05}$  & $\delta_{1.1}$   & $\delta_{1.25}$  & $\delta_{1.25^2}$ & $\delta_{1.25^3}$ \\ \hline
\multirow{5}{*}{\rotatebox{90}{Pnas}}                       & $\mathcal{L}_{1}$      & 0.4817          & 0.0780          & 0.1213          & 0.0933          & 34.59\%          & 59.98\%          & 87.25\%          & 96.30\%           & 98.50\%           \\
                                                            & $\mathcal{L}_{cosine}$ & 0.4825          & 0.0782          & 0.1216          & 0.1014          & 37.04\%          & 60.96\%          & 87.48\%          & 96.36\%           & 98.46\%           \\
                                                            & $\mathcal{L}_{grad}$   & \second{0.4616} & \third{0.0749}  & \second{0.1163} & \first{0.0889}  & \third{37.40\%}  & \third{62.46\%}  & \third{88.39\%}  & \third{96.63\%}   & \third{98.57\%}   \\
                                                            & $\mathcal{L}_{comb}$   & \first{0.4613}  & \first{0.0740}  & \first{0.1143}  & \second{0.0892} & \first{38.56\%}  & \first{63.31\%}  & \first{88.70\%}  & \first{96.68\%}   & \second{98.62\%}  \\
                                                            & $\mathcal{L}_{vnl}$    & \third{0.4640}  & \second{0.0743} & \third{0.1165}  & \third{0.0920}  & \second{37.67\%} & \second{62.60\%} & \second{88.47\%} & \second{96.64\%}  & \first{98.65\%}   \\ \hline
\multirow{5}{*}{\rotatebox{90}{UNet}}                       & $\mathcal{L}_{1}$      & 0.4215          & 0.2033          & \third{0.1138}  & 0.0744          & \second{37.54\%} & \third{60.47\%}  & 88.05\%          & 97.01\%           & 98.81\%           \\
                                                            & $\mathcal{L}_{cosine}$ & 0.4152          & \first{0.0841}  & 0.1170          & 0.0736          & 34.06\%          & 59.75\%          & 88.13\%          & 97.13\%           & \third{98.99\%}   \\
                                                            & $\mathcal{L}_{grad}$   & \third{0.4061}  & 0.4264          & \second{0.1135} & \second{0.0682} & \third{37.49\%}  & \second{60.93\%} & \second{88.50\%} & \second{97.17\%}  & 98.91\%           \\
                                                            & $\mathcal{L}_{comb}$   & \second{0.4041} & \third{0.1459}  & 0.1146          & \third{0.0692}  & 37.24\%          & 60.44\%          & \third{88.31\%}  & \third{97.15\%}   & \first{99.04\%}   \\
                                                            & $\mathcal{L}_{vnl}$    & \first{0.3967}  & \second{0.1182} & \first{0.1095}  & \first{0.0672}  & \first{38.62\%}  & \first{62.16\%}  & \first{89.08\%}  & \first{97.35\%}   & \second{99.03\%}  \\ \hline
\multirow{5}{*}{\rotatebox{90}{DenseNet}}                   & $\mathcal{L}_{1}$      & 0.4672          & 0.5580          & 0.1223          & 0.0896          & 37.53\%          & 60.52\%          & 86.72\%          & 96.27\%           & 98.37\%           \\
                                                            & $\mathcal{L}_{cosine}$ & 0.4603          & \first{0.0752}  & \third{0.1145}  & \third{0.0817}  & \third{37.57\%}  & \third{62.61\%}  & \third{88.03\%}  & \first{96.75\%}   & \first{98.64\%}   \\
                                                            & $\mathcal{L}_{grad}$   & \second{0.4488} & \third{0.3847}  & 0.1210          & 0.0827          & 33.25\%          & 59.71\%          & 87.39\%          & \second{96.73\%}  & \second{98.63\%}  \\
                                                            & $\mathcal{L}_{comb}$   & \third{0.4490}  & \second{0.2565} & \first{0.1129}  & \second{0.0806} & \second{38.30\%} & \second{63.02\%} & \first{88.56\%}  & \third{96.66\%}   & \third{98.54\%}   \\
                                                            & $\mathcal{L}_{vnl}$    & \first{0.4481}  & 0.6177          & \second{0.1142} & \first{0.0805}  & \first{39.28\%}  & \first{63.34\%}  & \second{88.49\%} & 96.66\%           & 98.43\%           \\ \hline
\multirow{5}{*}{\rotatebox{90}{ResNet}}                     & $\mathcal{L}_{1}$      & 0.4755          & \third{0.1639}  & 0.1310          & 0.0942          & 31.22\%          & 55.89\%          & \third{85.56\%}  & 96.27\%           & 98.57\%           \\
                                                            & $\mathcal{L}_{cosine}$ & \third{0.4700}  & \first{0.0804}  & 0.1279          & 0.0949          & \first{37.37\%}  & \third{57.92\%}  & 85.32\%          & \third{96.35\%}   & \second{98.62\%}  \\
                                                            & $\mathcal{L}_{grad}$   & 0.4734          & 0.2495          & \third{0.1278}  & \third{0.0916}  & \second{35.23\%} & 57.34\%          & 85.54\%          & 96.20\%           & 98.50\%           \\
                                                            & $\mathcal{L}_{comb}$   & \first{0.4573}  & \second{0.1200} & \second{0.1272} & \second{0.0894} & 34.53\%          & \second{57.97\%} & \first{86.26\%}  & \second{96.56\%}  & \first{98.71\%}   \\
                                                            & $\mathcal{L}_{vnl}$    & \second{0.4607} & 0.2938          & \first{0.1236}  & \first{0.0862}  & \third{34.75\%}  & \first{59.16\%}  & \second{86.11\%} & \first{96.60\%}   & \third{98.60\%}   \\ \hline
\multirow{5}{*}{\rotatebox{90}{ResNet\textsubscript{skip}}} & $\mathcal{L}_{1}$      & 0.4373          & 0.2430          & 0.1161          & 0.0783          & 37.07\%          & 60.60\%          & 87.68\%          & 96.86\%           & 98.75\%           \\
                                                            & $\mathcal{L}_{cosine}$ & 0.4347          & \third{0.1070}  & 0.1139          & 0.0772          & \second{39.80\%} & 61.31\%          & \third{88.27\%}  & 97.02\%           & 98.81\%           \\
                                                            & $\mathcal{L}_{grad}$   & \first{0.4107}  & 0.2710          & \first{0.1089}  & \first{0.0717}  & \third{38.93\%}  & \first{63.31\%}  & \first{89.51\%}  & \first{97.32\%}   & \first{98.92\%}   \\
                                                            & $\mathcal{L}_{comb}$   & \second{0.4165} & \first{0.0843}  & \second{0.1102} & \second{0.0722} & 36.71\%          & \third{61.92\%}  & \second{89.17\%} & \second{97.24\%}  & \second{98.90\%}  \\
                                                            & $\mathcal{L}_{vnl}$    & \third{0.4260}  & \second{0.0967} & \third{0.1125}  & \third{0.0756}  & \first{39.92\%}  & \second{62.53\%} & 88.22\%          & \third{97.04\%}   & \third{98.88\%}   \\ \hline
\end{tabular}
\normalsize
\vspace{-0.35cm}
\end{table*}

%% file: Metrics.tex
Since the introduction of the first set of metrics for data-driven depth estimation \cite{eigen2014depth}, namely root mean squared error (\textit{RMSE}), room mean squared logarithmic error (\textit{RMSLE}), absolute relative error (\textit{AbsRel}), squared relative error (\textit{SqRel}), and the relative threshold ($t$) based accuracies ($\delta_t$), these metrics are the standard approach for evaluation depth estimation performance.
More recent works have identified some shortcomings of these metrics.
Specifically in \cite{koch2018evaluation} an expanded analysis of depth estimation quality measures was conducted, and focused on two important traits, planarity and discontinuities.
The latter is very important for some downstream applications like view synthesis, and apart from the completeness (\textit{comp}) and accuracy (\textit{acc}) of the depth boundary errors (\textit{dbe}) proposed in \cite{koch2018evaluation}, another set of accuracy metrics were proposed in \cite{hu2019revisiting}.
The precision, recall and their harmonic mean (F1-score) were used after extracting different boundary layers via Sobel edge thresholding.
Planarity is also very important for various downstream applications, especially for indoor 3D reconstruction.
Finally, to overcome resolution \cite{cadena2016measuring} and focal length variations \cite{chen2020oasis}, recent perspective depth estimation works resort to nearest-neighbor 3D metrics.

\textbf{Direct Depth Metrics.}
We build upon these developments and design our benchmark to provide a holistic evaluation of depth estimation models.
Given the progress of recent data-driven models we expand the $\delta$ accuracies with two lower thresholds, a \textit{strict} at $\delta_{1.1}$, and a \textit{precise}, $\delta_{1.05}$, similar to \cite{huang2019indoor}.
However, these metrics, when applied directly on equirectangular images, are biased by its distortion towards the poles.
To remove this bias we take the spherically weighted mean (denoting with a $w$ prefix), which is standard practise for \360 image/video quality assessment \cite{xu2020state} and was also used in \cite{zioulis2019spherical}.
For the $\delta_t$ accuracies though, we turn to uniform sampling on the sphere using the projected vertices of a subdivided icosahedron, denoted as $\delta_t^{ico^K}$, $K$ being the icosahedron's order.

\textbf{Depth Discontinuity Metrics.}
We complement the direct depth performance metrics with a set of secondary metrics measuring performance at preserving the depth discontinuities, usually manifesting at object boundaries.
While \cite{koch2018evaluation} used manual annotation and structured edge detection \cite{dollar2014fast}, we follow the approach of \cite{ramamonjisoa2020predicting} that relies on automatic Canny edge detection \cite{canny1986computational}.
In addition, we complement the depth boundary errors (\textit{dbe}\textsuperscript{acc} and \textit{dbe}\textsuperscript{comp}) with the accuracy metrics of \cite{hu2019revisiting} (\textit{prec}\textsubscript{t}, \textit{rec}\textsubscript{t}) using the same thresholds, $t \in \{0.25, 0.5, 1.0\}$, for both set of metrics.

\textbf{Depth Smoothness Metrics.}
While the planarity metric of \cite{koch2018evaluation} required the manual annotation of samples, its goal is to measure the smoothness of the inferred depth with respect to dominant structures.
A straightforward adaptation that alleviates annotations is the use of surface orientation metrics, which is a property directly derived from the depth measurements.
Using spherical-to-Cartesian coordinates conversion the depth/radius measurements are lifted to 3D points, with the surface orientation extracted by exploiting the structured nature of images.
Similar to how surface estimation methods measure performance, we use the angular RMSE (\textit{RMSE\textsuperscript{o}}), and a set of accuracies $\alpha_{d^o}$ with pre-defined angle thresholds $d$, using those from \cite{wang2020vplnet} ($d \in \{11.25^o, 22.5^o, 30^o\}$).

\textbf{Geometric Metrics.}
Depth estimations are typically used in downstream applications for metric-scale 3D perception.
Therefore, 3D performance metrics are reasonable to assess suitability for downstream tasks.
We use two different metrics that aggregate the performance of the aforementioned different depth traits, \textit{i.e.}~accuracy and precision, boundary preservation and smoothness.
The first geometric metric is computed on the point cloud level (\textit{c2c}), using a point-to-plane distance between each point and its closest correspondence with the ground truth point cloud.
The point-to-plane distance jointly encodes depth correctness and smoothness, while the closest point query will penalize boundary errors.
The second geometric metric is computed on the mesh level, having each point cloud (predicted and ground truth) 3D reconstructed using the Screened Poisson Surface Reconstruction \cite{kazhdan2013screened}.
We then calculate the Hausdorff distance \cite{cignoni1998metro} between the two meshes.
Similarly, Poisson reconstruction leverages both position and surface information when generating the scene's mesh.
Through this metric we assess the capacity to represent the entire scene's geometry with the estimated depth, an important trait for some downstream applications.

%% file: ExperimentalSetup.tex
We design our experiments and search for a solid baseline taking recent developments into account.

\textbf{Supervision.}
As shown in \cite{carvalho2018regression} the L1 ($\mathcal{L}_1$) loss exhibits the best convergence for monocular depth estimation irrespective of the model size and  architecture complexity, indicating that models behaving like median estimators are more appropriate.
Most recent works for \360 depth estimation \cite{wang2020bifuse,jiang2021unifuse,eder2019mapped} use the berHu loss \cite{laina2016deeper}, with the exception to this rule being \cite{sun2020hohonet} that uses the L1 loss.

We additionally observe that these works rely solely on a single direct depth loss, while recent works on perspective depth estimation also include additional losses.
MiDaS \cite{ranftl2020towards}, MegaDepth \cite{li2018megadepth} as well as \cite{xian2020structure} and \cite{hu2019revisiting} use a multi-scale ($K=4$ scales) gradient matching term ($\mathcal{L}_{grad}$) that enforces consistent depth discontinuities.
While their terms are scale-invariant and operate in the log-space, \360 depth does not suffer from disparity/baseline or focal length variations, and since we do not use the L1 loss in log-space (as its performance is inferior to pure L1 \cite{carvalho2018regression}), we use a non-scale invariant version of this loss.
Apart from boundary preservation, the piece-wise smooth nature of depth, necessitates the use of a suitable prior for the predictions.
This was acknowledged in \cite{hu2019revisiting}, where a surface orientation consistency loss was used ($\mathcal{L}_{cosine}$).
Prior works employed smoothness priors on the predictions, and, to overcome cross boundary smoothing, relied on image gradient weighting \cite{godard2017unsupervised}.
Yet image gradients do not necessarily align with depth discontinuities, making the normal loss a better candidate.

Finally, the newly introduced virtual normal loss \cite{yin2019enforcing} ($\mathcal{L}_{vnl}$) is a long-range relationship oriented objective, which given the global context of spherical panoramas is well aligned with the task.
In our experiments we follow a progressive loss ablation starting with a $\mathcal{L}_{1}$ objective, examining the effect of $\mathcal{L}_{grad}$ and $\mathcal{L}_{cosine}$ on the $\mathcal{L}_{1}$ baseline, as well as their combined effect $\mathcal{L}_{comb}$, and finally further extend the combined objective with $\mathcal{L}_{vnl}$, with the latter experiment including all losses.

\textbf{Model Architecture.}
The importance of high-capacity encoders, pre-training and multi-scale predictions is acknowledged in the literature \cite{ranftl2020towards}.
Building on the first, we preserve a consistent convolution decoder and use a DenseNet-161 \cite{huang2017densely} ({$55$}M parameters) and ResNet-152 \cite{he2016identity} ({$110$}M parameters)  encoder as baselines.
Inspired by recent work \cite{lee2020multi} we also include their Pnas model ({$99$}M parameters) whose encoder is a product of neural architecture search \cite{liu2018progressive}.
In addition, taking into account the boundary preservation performance of skip connections, we also use the -- largely unpopular for depth estimation -- UNet model \cite{ronneberger2015u} ({$27$}M parameters).
Since it is a purely convolutional model, we additionally modify the ResNet-152 model with skip connections starting from the first residual block ({$112$}M parameters), in contrast to UNet's very early layer encoder-to-decoder skip.
Since pre-training weights are not available for UNet, we experiment with cold-started models, and also simplify training using a single-scale predictions as the multi-scale effect should be horizontal across all models with the same convolution decoder structure.

\textbf{Periodic Displacement Fields Refinement.}
We additionally consider the refinement of the predicted depth, using a shallow hourglass module \cite{newell2016stacked}.
It is adapted for the task at hand, with two branches, one for the input color image and the other for the predicted depth map.
Across each stage, we account for the varying nature of each branch's feature statistics using Adaptive Instance Normalization \cite{huang2017arbitrary}.
We follow the recent approach of \cite{ramamonjisoa2020predicting} that shows how predicting displacement fields instead of residuals produces higher quality depth refinement.
However, the spherical domain is continuous, and thus, we need to account for the horizontal discontinuity of the equirectangular projection.
To achieve this in a locally differentiable manner, we resort to a periodic reconstruction of the sampling coordinates.
Considering the final sampling coordinates $(\phi, \theta)$ after adding the displacement field, we wrap them around to $(\tau, \theta)$, with $\tau = \atantwo(-\sin\phi, -\cos\phi) + \pi$.

\textbf{Training and Evaluation.}
We train all models solely on the official \textit{train} split of M3D, and evaluate them on its official \textit{test} split as well.
Evaluation is conducted across all the aforementioned axes of depth performance.
Apart from this holistic performance analysis, we additionally take an orthogonal direction and assess the models' generalization performance on zero-shot cross-dataset transfer using the GV2 \textit{tiny}, \textit{medium} and \textit{fullplus} splits.
Given that both GV2 and M3D scenes were scanned with same type of camera (\textit{i.e.}~Matterport), we render another version of \textit{tiny} which is tone mapped to a film-like dynamic range, dubbed \textit{tiny-filmic}, changing the camera-related data domain.
Our experiments are conducted on two different resolutions $512 \times 256$ and $1024 \times 512$ (we render all datasets to both) to assess cross-resolution performance.

\input{Tables/Table2}

\input{Tables/Table3}

\input{Tables/Table4updt}

\input{Tables/Table5}

%% file: Tables/Table2.tex
\begin{table*}[!htbp]
\centering
\caption{Direct depth performance using spherical metrics. 
A UNet model with spherical padding is also presented (\colorbox{lightorg}{light pink}), as well as the two better performing models trained and tested on the 3D60 (\colorbox{m3d_old_table}{light orange}) and Structured3D (\colorbox{s3d_table}{light red}) datasets.}
\label{tab:spherical}
\footnotesize
\begin{tabular}{l|cccc|ccccc}
\hline
\multicolumn{1}{c|}{\multirow{2}{*}{Model}}                                            & \multicolumn{4}{c|}{Depth Error $\downarrow$}                                 & \multicolumn{5}{c}{Depth Accuracy $\uparrow$}                                                                                      \\
\multicolumn{1}{c|}{}                                                                  & \textit{$w$RMSE} & \textit{$w$RMSLE} & \textit{$w$AbsRel} & \textit{$w$SqRel} & $\delta_{1.05}^{ico^6}$ & $\delta_{1.1}^{ico^6}$ & $\delta_{1.25}^{ico^6}$ & $\delta_{1.25^2}^{ico^6}$ & $\delta_{1.25^3}^{ico^6}$ \\ \hline
Pnas\textsuperscript{$comb$}                                                           & 0.5367           & \first{0.0811}    & 0.1259             & 0.1153            & \second{36.44\%}        & \third{60.52\%}        & 86.80\%                 & 95.83\%                   & 98.11\%                   \\
UNet\textsuperscript{$vnl$}                                                            & \first{0.4520}   & \third{0.1300}    & \first{0.1147}     & \first{0.0811}    & \first{36.68\%}         & \second{60.59\%}       & \first{88.31\%}         & \first{96.96\%}           & \first{98.73\%}           \\
DenseNet\textsuperscript{$comb$}                                                       & \third{0.5209}   & 0.1982            & \third{0.1209}     & \third{0.1013}    & 35.97\%                 & 60.41\%                & \third{87.02\%}         & \third{95.96\%}           & 98.09\%                   \\
ResNet\textsuperscript{$comb$}                                                         & 0.5294           & 0.1365            & 0.1374             & 0.1127            & 32.03\%                 & 55.31\%                & 84.74\%                 & 95.81\%                   & \third{98.21\%}           \\
ResNet\rlap{\textsuperscript{$comb$}}\textsubscript{\textit{skip}}                         & \second{0.4788}  & \second{0.0927}   & \second{0.1166}    & \second{0.0893}   & \third{36.20\%}         & \first{60.64\%}        & \second{87.99\%}        & \second{96.62\%}          & \second{98.49\%}          \\ \hline
\cellcolor{lightorg}UNet\rlap{\textsuperscript{$vnl$}}\textsubscript{\textit{circ}}       & 0.4399           & 0.0685            & 0.1132             & 0.0769            & 36.85\%                 & 61.38\%                & 88.84\%                 & 97.25\%                   & 98.89\%                   \\ \hline
\cellcolor{m3d_old_table}UNet\textsuperscript{$vnl$} \quad @ 3D60                          & 0.3140           & 0.0455            & 0.0741             & 0.0316            & 49.99\%                 & 75.16\%                & 95.49\%                 & 99.11\%                   & 99.60\%                   \\
\cellcolor{m3d_old_table}ResNet \rlap{\textsuperscript{$comb$}}\textsubscript{\textit{skip}} \quad @ 3D60 & 0.3758           & 0.6100            & 0.0883             & 0.0481            & 46.03\%                 & 70.29\%                & 93.12\%                 & 98.41\%                   & 99.34\%                   \\ \hline
\cellcolor{s3d_table}UNet\textsuperscript{$vnl$} \quad @ S3D                           & 0.1815           & 0.0546            & 0.0919             & 0.0398            & 50.61\%                 & 75.98\%                & 92.23\%                 & 96.56\%                   & 97.53\%                   \\
\cellcolor{s3d_table}ResNet\rlap{\textsuperscript{$comb$}}\textsubscript{\textit{skip}} \quad @ S3D    & 0.2450           & 0.1335            & 0.1349             & 0.1249            & 40.48\%                 & 67.29\%                & 88.67\%                 & 95.01\%                   & 96.68\%                   \\ \hline
\end{tabular}%
\normalsize
\end{table*}

%% file: Tables/Table3.tex
\begin{table*}[!htbp]
\centering
\caption{\textit{Top half}: Depth discontinuity/boundary preservation and depth smoothness performance metrics.
\textit{Bottom half}: Same metrics after refining all models (subscript \textit{ref}) with a periodic displacement field hourglass module.}
\label{tab:ob_smooth}
\resizebox{\textwidth}{!}{%
\begin{tabular}{l|cccccccc|cccc}
\hline
\multicolumn{1}{c|}{\multirow{3}{*}{Model}}                           & \multicolumn{8}{c|}{Depth Discontinuity}                                                                                                                                                                                                                                                                             & \multicolumn{4}{c}{Depth Smoothness}                                                                              \\
\multicolumn{1}{c|}{}                                                 & \multicolumn{2}{c}{Error $\downarrow$}                                                      & \multicolumn{6}{c|}{Accuracy $\uparrow$}                                                                                                                                                                               & Error $\downarrow$                                    & \multicolumn{3}{c}{Accuracy $\uparrow$}                   \\
\multicolumn{1}{c|}{}                                                 & \textit{dbe}\textsuperscript{acc} & \multicolumn{1}{c|}{\textit{dbe}\textsuperscript{comp}} & \textit{prec}\textsubscript{$0.25$} & \textit{prec}\textsubscript{$0.5$} & \textit{prec}\textsubscript{$1$} & \textit{rec}\textsubscript{$0.25$} & \textit{rec}\textsubscript{$0.5$} & \textit{rec}\textsubscript{$1$} & \multicolumn{1}{c|}{\textit{RMSE\textsuperscript{o}}} & $\alpha_{11.25^o}$ & $\alpha_{22.5^o}$ & $\alpha_{30^o}$  \\ \hline
Pnas\textsuperscript{$comb$}                                          & 2.5119                            & \multicolumn{1}{c|}{5.3501}                             & 39.83\%                             & 31.59\%                            & 27.01\%                          & 23.53\%                            & 14.42\%                           & 10.98\%                         & \multicolumn{1}{c|}{\first{15.26}}                    & \first{67.73\%}    & \first{77.99\%}   & \third{81.67\%}  \\
UNet\textsuperscript{$vnl$}                                           & \first{1.2699}                    & \multicolumn{1}{c|}{\first{3.8876}}                     & \first{58.97\%}                     & \first{57.54\%}                    & \first{51.85\%}                  & \first{43.96\%}                    & \first{36.69\%}                   & \first{28.59\%}                 & \multicolumn{1}{c|}{16.02}                            & 61.80\%            & 76.58\%           & \second{81.70\%} \\
DenseNet\textsuperscript{$comb$}                                      & \third{2.0628}                    & \multicolumn{1}{c|}{\third{5.0977}}                     & \third{47.16\%}                     & \third{40.77\%}                    & \third{35.20\%}                  & \third{26.09\%}                    & \third{16.87\%}                   & \third{12.21\%}                 & \multicolumn{1}{c|}{\third{15.98}}                    & \second{64.58\%}   & \third{76.86\%}   & 81.20\%          \\
ResNet\textsuperscript{$comb$}                                        & 2.2393                            & \multicolumn{1}{c|}{5.3796}                             & 44.10\%                             & 36.70\%                            & 27.44\%                          & 22.91\%                            & 12.23\%                           & 7.20\%                          & \multicolumn{1}{c|}{16.63}                            & 63.09\%            & 75.70\%           & 80.20\%          \\
ResNet\rlap{\textsuperscript{$comb$}}\textsubscript{\textit{skip}}        & \second{1.4883}                   & \multicolumn{1}{c|}{\second{4.5346}}                    & \second{57.34\%}                    & \second{54.11\%}                   & \second{47.57\%}                 & \second{33.99\%}                   & \second{24.30\%}                  & \second{16.37\%}                & \multicolumn{1}{c|}{\second{15.27}}                   & \third{64.18\%}    & \second{77.57\%}  & \first{82.27\%}  \\ \hline
Pnas\rlap{\textsuperscript{$comb$}}\textsubscript{\textit{ref}}         & 2.2861                            & \multicolumn{1}{c|}{5.0435}                             & 46.66\%                             & 44.74\%                            & 37.96\%                          & \third{30.66\%}                    & 26.00\%                           & \third{22.49\%}                 & \multicolumn{1}{c|}{17.83}                            & \third{63.33\%}    & 74.01\%           & 78.15\%          \\
UNet\rlap{\textsuperscript{$vnl$}}\textsubscript{\textit{ref}}            & \first{1.4241}                    & \multicolumn{1}{c|}{\first{3.8505}}                     & \second{53.46\%}                    & \second{51.38\%}                   & \second{44.36\%}                 & \first{43.09\%}                    & \first{41.54\%}                   & \first{37.50\%}                 & \multicolumn{1}{c|}{16.86}                            & 61.50\%            & 75.70\%           & \second{80.64\%} \\
DenseNet\rlap{\textsuperscript{$comb$}}\textsubscript{\textit{ref}}      & \third{1.9769}                    & \multicolumn{1}{c|}{\third{4.9026}}                     & \third{51.94\%}                     & \third{48.43\%}                    & \third{40.63\%}                  & 30.61\%                            & \third{26.14\%}                   & 22.46\%                         & \multicolumn{1}{c|}{\second{16.49}}                   & \second{63.80\%}   & \second{76.17\%}  & \third{80.58\%}  \\
ResNet\rlap{\textsuperscript{$comb$}}\textsubscript{\textit{ref}}         & 2.1078                            & \multicolumn{1}{c|}{5.0783}                             & 50.77\%                             & 46.52\%                            & 36.58\%                          & 28.31\%                            & 23.33\%                           & 19.37\%                         & \multicolumn{1}{c|}{\third{16.68}}                    & 63.08\%            & \third{75.82\%}   & 80.36\%          \\
ResNet\rlap{\textsuperscript{$comb$}}\textsubscript{\textit{skip} \& \textit{ref}} & \second{1.4291}                   & \multicolumn{1}{c|}{\second{4.3115}}                    & \first{60.78\%}                     & \first{58.09\%}                    & \first{51.49\%}                  & \second{37.79\%}                   & \second{32.55\%}                  & \second{27.23\%}                & \multicolumn{1}{c|}{\first{15.05}}                    & \first{65.16\%}    & \first{78.26\%}   & \first{82.77\%}  \\ \hline
\end{tabular}%
}
\end{table*}

%% file: Tables/Table4updt.tex
\begin{table*}[!htbp]
\centering
\caption{Consolidated performance on the GibsonV2 splits, across all depth traits, using a subset of the available metrics, for models trained on the Matterport3D data.
The best performing model (UNet) trained with photometric augmentation is also presented (\colorbox{lightorg}{light pink}).}
\label{tab:gv2}
\resizebox{\textwidth}{!}{%

\begin{tabular}{ll|ccccc|ccccc|cc}
\hline
\multicolumn{1}{c}{\multirow{3}{*}{GV2}} &
  \multicolumn{1}{c|}{\multirow{3}{*}{Model}} &
  \multicolumn{5}{c|}{Direct Depth} &
  \multicolumn{5}{c|}{Depth Discontinuity} &
  \multicolumn{2}{c}{Depth Smoothness} \\
\multicolumn{1}{c}{} &
  \multicolumn{1}{c|}{} &
  \multicolumn{3}{c}{Error $\downarrow$} &
  \multicolumn{2}{c|}{Accuracy $\uparrow$} &
  \multicolumn{2}{c}{Error $\downarrow$} &
  \multicolumn{3}{c|}{Accuracy $\uparrow$} &
  Error $\downarrow$ &
  Accuracy $\uparrow$ \\
\multicolumn{1}{c}{} &
  \multicolumn{1}{c|}{} &
  \textit{$w$RMSE} &
  \textit{$w$RMSLE} &
  \multicolumn{1}{c|}{\textit{$w$AbsRel}} &
  $\delta_{1.1}^{ico^6}$ &
  $\delta_{1.25}^{ico^6}$ &
  \textit{dbe}\textsuperscript{acc} &
  \multicolumn{1}{c|}{\textit{dbe}\textsuperscript{comp}} &
  \textit{prec}\textsubscript{$0.25$} &
  \textit{prec}\textsubscript{$0.5$} &
  \textit{prec}\textsubscript{$1$} &
  \multicolumn{1}{c|}{\textit{RMSE\textsuperscript{o}}} &
  $\alpha_{30^o}$ \\ \hline
\multirow{6}{*}{\rotatebox{90}{\textit{tiny}}} &
  Pnas\textsuperscript{$comb$} &
  0.5574 &
  \second{0.0970} &
  0.1945 &
  36.01\% &
  76.76\% &
  2.6616 &
  5.6187 &
  34.90\% &
  30.67\% &
  25.07\% &
  \first{15.91} &
  \second{81.68\%} \\
 &
  UNet\textsuperscript{$vnl$} &
  \first{0.4723} &
  0.2060 &
  \first{0.1733} &
  \first{41.67\%} &
  \first{81.49\%} &
  \first{1.4726} &
  \first{4.3377} &
  \first{61.43\%} &
  \first{64.51\%} &
  \first{60.21\%} &
  17.35 &
  80.71\% \\
 &
  DenseNet\textsuperscript{$comb$} &
  \third{0.5131} &
  \third{0.1368} &
  \second{0.1738} &
  \third{38.62\%} &
  \second{79.99\%} &
  \third{2.2068} &
  \third{5.2911} &
  \third{43.19\%} &
  \third{40.05\%} &
  \third{35.32\%} &
  \second{16.24} &
  \third{81.66\%} \\
 &
  ResNet\textsuperscript{$comb$} &
  0.5426 &
  0.1427 &
  0.2113 &
  31.87\% &
  72.80\% &
  2.3665 &
  5.5963 &
  40.64\% &
  37.11\% &
  30.21\% &
  16.97 &
  80.64\% \\
 &
  ResNet\rlap{\textsuperscript{$comb$}}\textsubscript{\textit{skip}} &
  \second{0.4932} &
  \first{0.0900} &
  \third{0.1747} &
  \second{39.26\%} &
  \third{79.86\%} &
  \second{1.6406} &
  \second{4.7710} &
  \second{55.44\%} &
  \second{56.69\%} &
  \second{52.48\%} &
  \third{16.24} &
  \first{81.93\%} \\ \cline{2-14} 
 &
  \cellcolor{lightorg} UNet\rlap{\textsuperscript{$vnl$}}\textsubscript{\textit{skip} \& \textit{aug}} &
  0.4580 &
  0.0840 &
  0.1701 &
  39.73\% &
  81.19\% &
  1.4480 &
  4.2681 &
  62.69\% &
  66.19\% &
  62.27\% &
  16.30 &
  82.16\% \\ \hline
\multirow{6}{*}{\rotatebox{90}{\textit{medium}}} &
  Pnas\textsuperscript{$comb$} &
  0.5053 &
  \second{0.0926} &
  0.1866 &
  34.85\% &
  78.58\% &
  2.6420 &
  5.5068 &
  36.54\% &
  31.80\% &
  27.25\% &
  \first{14.31} &
  \first{84.06\%} \\
 &
  UNet\textsuperscript{$vnl$} &
  \first{0.4416} &
  0.1876 &
  \first{0.1665} &
  \first{42.49\%} &
  \first{82.50\%} &
  \first{1.5245} &
  \first{4.3178} &
  \first{62.75\%} &
  \first{65.68\%} &
  \first{60.22\%} &
  16.39 &
  82.43\% \\
 &
  DenseNet\textsuperscript{$comb$} &
  \third{0.4661} &
  0.1670 &
  \second{0.1669} &
  \third{39.30\%} &
  \second{81.72\%} &
  \third{2.2311} &
  \third{5.2215} &
  \third{44.53\%} &
  \third{41.16\%} &
  \third{36.07\%} &
  \second{15.15} &
  \second{83.50\%} \\
 &
  ResNet\textsuperscript{$comb$} &
  0.5023 &
  \third{0.1317} &
  0.2058 &
  32.12\% &
  73.67\% &
  2.3915 &
  5.4622 &
  41.86\% &
  37.73\% &
  30.38\% &
  15.86 &
  82.48\% \\
 &
  ResNet\rlap{\textsuperscript{$comb$}}\textsubscript{\textit{skip}} &
  \second{0.4563} &
  \first{0.0884} &
  \third{0.1677} &
  \second{39.98\%} &
  \third{81.34\%} &
  \second{1.6930} &
  \second{4.7230} &
  \second{56.33\%} &
  \second{57.24\%} &
  \second{51.81\%} &
  \third{15.44} &
  \third{83.30\%} \\ \cline{2-14} 
 &
  \cellcolor{lightorg} UNet\rlap{\textsuperscript{$vnl$}}\textsubscript{\textit{skip} \& \textit{aug}} &
  0.4321 &
  0.0823 &
  0.1673 &
  39.70\% &
  81.90\% &
  1.5045 &
  4.2659 &
  63.94\% &
  67.27\% &
  61.69\% &
  15.43 &
  83.70\% \\ \hline
\multirow{6}{*}{\rotatebox{90}{\textit{fullplus}}} &
  Pnas\textsuperscript{$comb$} &
  0.6759 &
  \first{0.1139} &
  0.1991 &
  38.60\% &
  73.75\% &
  2.8383 &
  6.1612 &
  32.61\% &
  26.83\% &
  21.81\% &
  \first{19.83} &
  \first{75.93\%} \\
 &
  UNet\textsuperscript{$vnl$} &
  \first{0.6167} &
  0.2657 &
  \third{0.1844} &
  \first{42.42\%} &
  \first{76.21\%} &
  \first{1.7228} &
  \first{5.0369} &
  \first{54.45\%} &
  \first{56.37\%} &
  \first{52.31\%} &
  22.05 &
  73.41\% \\
 &
  DenseNet\textsuperscript{$comb$} &
  \third{0.6684} &
  0.1649 &
  \second{0.1835} &
  \second{40.79\%} &
  \second{74.87\%} &
  \third{2.4985} &
  \third{6.0993} &
  \third{39.33\%} &
  \third{34.44\%} &
  \third{27.63\%} &
  \second{20.57} &
  \third{75.18\%} \\
 &
  ResNet\textsuperscript{$comb$} &
  0.6690 &
  \third{0.1504} &
  0.2095 &
  37.35\% &
  71.42\% &
  2.6259 &
  6.2642 &
  37.82\% &
  32.27\% &
  23.59\% &
  21.00 &
  74.54\% \\
 &
  ResNet\rlap{\textsuperscript{$comb$}}\textsubscript{\textit{skip}} &
  \second{0.6370} &
  \second{0.1183} &
  \first{0.1828} &
  \second{41.28\%} &
  \second{75.45\%} &
  \second{1.9257} &
  \second{5.5758} &
  \second{50.05\%} &
  \second{48.96\%} &
  \second{41.74\%} &
  \third{20.61} &
  \second{75.18\%} \\ \cline{2-14} 
 &
  \cellcolor{lightorg} UNet\rlap{\textsuperscript{$vnl$}}\textsubscript{\textit{skip} \& \textit{aug}} &
  0.6014 &
  0.1033 &
  0.1758 &
  42.70\% &
  76.97\% &
  1.7040 &
  5.0063 &
  56.24\% &
  58.18\% &
  53.33\% &
  20.87 &
  75.09\% \\ \hline
\multirow{6}{*}{\rotatebox{90}{\textit{tiny filmic}}} &
  Pnas\textsuperscript{$comb$} &
  \third{0.6268} &
  \first{0.1088} &
  \third{0.1939} &
  \third{37.03\%} &
  \second{75.66\%} &
  2.9347 &
  \third{6.1523} &
  32.01\% &
  27.16\% &
  21.20\% &
  \first{17.34} &
  \first{79.73\%} \\
 &
  UNet\textsuperscript{$vnl$} &
  \first{0.5448} &
  0.2315 &
  \first{0.1848} &
  \first{42.82\%} &
  \first{79.43\%} &
  \first{1.6943} &
  \first{4.8443} &
  \first{57.63\%} &
  \first{59.49\%} &
  \first{53.19\%} &
  \third{19.21} &
  \third{78.00\%} \\
 &
  DenseNet\textsuperscript{$comb$} &
  0.6903 &
  0.1896 &
  0.1968 &
  35.34\% &
  73.48\% &
  2.8225 &
  6.3933 &
  37.14\% &
  31.85\% &
  \third{24.24\%} &
  19.29 &
  77.37\% \\
 &
  ResNet\textsuperscript{$comb$} &
  \second{0.6107} &
  \third{0.1479} &
  0.2036 &
  35.08\% &
  73.29\% &
  \third{2.7016} &
  6.1781 &
  \third{37.34\%} &
  \third{32.57\%} &
  22.22\% &
  \second{18.30} &
  \second{78.64\%} \\
 &
  ResNet\rlap{\textsuperscript{$comb$}}\textsubscript{\textit{skip}} &
  0.6445 &
  \second{0.1195} &
  \second{0.1863} &
  \second{39.00\%} &
  \third{75.19\%} &
  \second{2.1093} &
  \second{5.7670} &
  \second{50.19\%} &
  \second{48.58\%} &
  \second{37.50\%} &
  19.26 &
  77.35\% \\ \cline{2-14} 
 &
 \cellcolor{lightorg} UNet\rlap{\textsuperscript{$vnl$}}\textsubscript{\textit{skip \& \textit{aug}}} &
  0.4750 &
  0.0871 &
  0.1743 &
  39.51\% &
  80.39\% &
  1.5326 &
  4.3923 &
  60.69\% &
  63.32\% &
  59.43\% &
  16.66 &
  81.62\% \\ \hline
\multirow{6}{*}{\rotatebox{90}{\textit{fullplus filmic}}} &
  Pnas\textsuperscript{$comb$} &
  \third{0.7866} &
  \first{0.1344} &
  \first{0.2129} &
  \third{35.97\%} &
  \second{68.86\%} &
  3.1505 &
  \third{6.7791} &
  29.06\% &
  22.40\% &
  16.16\% &
  \first{21.55} &
  \first{73.61\%} \\
 &
  UNet\textsuperscript{$vnl$} &
  \first{0.7368} &
  0.2975 &
  \third{0.2199} &
  \first{38.47\%} &
  \first{70.20\%} &
  \first{1.9476} &
  \first{5.5601} &
  \first{50.65\%} &
  \first{50.90\%} &
  \first{44.46\%} &
  \third{23.89} &
  \third{70.69\%} \\
 &
  DenseNet\textsuperscript{$comb$} &
  0.9258 &
  0.2207 &
  0.2292 &
  33.54\% &
  63.80\% &
  3.1514 &
  7.1900 &
  32.52\% &
  25.41\% &
  \third{17.75\%} &
  24.04 &
  70.14\% \\
 &
  ResNet\textsuperscript{$comb$} &
  \second{0.7786} &
  \third{0.1727} &
  \second{0.2154} &
  \second{35.99\%} &
  \third{68.24\%} &
  \third{2.9668} &
  6.8743 &
  \third{34.37\%} &
  \third{27.06\%} &
  16.76\% &
  \second{22.50} &
  \second{72.35\%} \\
 &
  ResNet\rlap{\textsuperscript{$comb$}}\textsubscript{\textit{skip}} &
  0.8705 &
  \second{0.1632} &
  0.2217 &
  34.94\% &
  65.30\% &
  \second{2.4696} &
  \second{6.6184} &
  \second{43.73\%} &
  \second{39.23\%} &
  \second{27.65\%} &
  23.91 &
  70.13\% \\ \cline{2-14} 
 &
  \cellcolor{lightorg} UNet\rlap{\textsuperscript{$vnl$}}\textsubscript{\textit{skip} \& \textit{aug}} &
  0.6237 &
  0.1084 &
  0.1829 &
  41.57\% &
  75.56\% &
  1.7688 &
  5.1482 &
  54.64\% &
  55.56\% &
  50.02\% &
  21.23 &
  74.58\% \\ \hline
\end{tabular}%

}
\end{table*}

%% file: Tables/Table5.tex
\begin{table*}[!htbp]
\centering
\caption{Consolidated depth performance across all traits using a subset of the available metrics.
Results are presented for the two best performing models on the GibsonV2 splits which have been trained on the Matterport3D train split.}
\label{tab:resolution}
\resizebox{\textwidth}{!}{%
\begin{tabular}{ll|ccccc|ccccc|cc}
\hline
\multicolumn{1}{c}{\multirow{3}{*}{GV2}}               & \multicolumn{1}{c|}{\multirow{3}{*}{Model}}                    & \multicolumn{5}{c|}{Direct Depth}                                                                                                 & \multicolumn{5}{c|}{Depth Discontinuity}                                                                                                                                                                  & \multicolumn{2}{c}{Depth Smoothness}                                        \\
\multicolumn{1}{c}{}                                   & \multicolumn{1}{c|}{}                                          & \multicolumn{3}{c}{Error $\downarrow$}                                         & \multicolumn{2}{c|}{Accuracy $\uparrow$}         & \multicolumn{2}{c}{Error $\downarrow$}                                                      & \multicolumn{3}{c|}{Accuracy $\uparrow$}                                                                    & Error $\downarrow$                                    & Accuracy $\uparrow$ \\
\multicolumn{1}{c}{}                                   & \multicolumn{1}{c|}{}                                          & \textit{$w$RMSE} & \textit{$w$RMSLE} & \multicolumn{1}{c|}{\textit{$w$AbsRel}} & $\delta_{1.1}^{ico^6}$ & $\delta_{1.25}^{ico^6}$ & \textit{dbe}\textsuperscript{acc} & \multicolumn{1}{c|}{\textit{dbe}\textsuperscript{comp}} & \textit{prec}\textsubscript{$0.25$} & \textit{prec}\textsubscript{$0.5$} & \textit{prec}\textsubscript{$1$} & \multicolumn{1}{c|}{\textit{RMSE\textsuperscript{o}}} & $\alpha_{30^o}$     \\ \hline
\multirow{2}{*}{\textit{tiny\textsuperscript{HR}}}     & UNet\textsuperscript{$vnl$}                                    & 0.5794           & \first{0.1247}    & 0.2151                                  & 31.98\%                & 62.05\%                 & \first{1.4330}                    & \first{5.1737}                                          & \first{44.84\%}                     & \first{46.13\%}                    & \first{41.57\%}                  & 22.36                                                 & 74.12\%             \\
                                                       & ResNet\rlap{\textsuperscript{$comb$}}\textsubscript{\textit{skip}} & \first{0.4993}   & 0.1273            & \first{0.1758}                          & \first{40.78\%}        & \first{80.31\%}         & 1.9271                            & 5.9666                                                  & 36.24\%                             & 37.68\%                            & 30.77\%                          & \first{15.65}                                         & \first{82.78\%}     \\ \hline
\multirow{2}{*}{\textit{medium\textsuperscript{HR}}}   & UNet\textsuperscript{$vnl$}                                    & 0.5901           & \first{0.1291}    & 0.2269                                  & 31.21\%                & 61.02\%                 & \first{1.6221}                    & \first{5.5436}                                          & \first{43.98\%}                     & \first{44.21\%}                    & \first{38.46\%}                  & 22.13                                                 & 74.73\%             \\
                                                       & ResNet\rlap{\textsuperscript{$comb$}}\textsubscript{\textit{skip}} & \first{0.4528}   & 0.1618            & \first{0.1664}                          & \first{42.03\%}        & \first{81.91\%}         & 2.0356                            & 5.8467                                                  & 34.27\%                             & 34.60\%                            & 27.81\%                          & \first{14.71}                                         & \first{84.46\%}     \\ \hline
\multirow{2}{*}{\textit{fullplus\textsuperscript{HR}}} & UNet\textsuperscript{$vnl$}                                    & 0.8772           & \first{0.1769}    & 0.2730                                  & 22.46\%                & 46.09\%                 & \first{1.7532}                    & \first{6.4628}                                          & \first{36.46\%}                     & \first{35.80\%}                    & \first{28.67\%}                  & 27.43                                                 & 65.07\%             \\
                                                       & ResNet\rlap{\textsuperscript{$comb$}}\textsubscript{\textit{skip}} & \first{0.6607}   & 0.2308            & \first{0.1836}                          & \first{41.18\%}        & \first{74.77\%}         & 2.3775                            & 6.9102                                                  & 28.70\%                             & 28.15\%                            & 20.71\%                          & \first{19.88}                                         & \first{76.30\%}     \\ \hline
\end{tabular}%
}
\vspace{-0.35cm}
\end{table*}

%% file: Results.tex
\textbf{Implementation Details.}
We implement all experiments with \href{https://github.com/ai-in-motion/moai}{\textit{moai}} \cite{moai}, using the same seed across all experiments.
For data generation we use \href{https://www.blender.org/}{Blender} and the \href{https://www.cycles-renderer.org/}{Cycles} path tracer using $256$ samples.
Our ResNets are built with pre-activated bottleneck blocks \cite{he2016identity} and all our models' weights are initialized with \cite{he2015delving}.
We optimize all models for $60$ epochs on a NVidia 2080 Ti, using Adam \cite{kingma2014adam} with a learning rate of $0.0002$ and default momentum parameters, and a consistent batch size of $4$.
All losses are unity (\textit{i.e.}~equally) weighted across all experiments.
We use \href{https://www.danielgm.net/cc/}{CloudCompare} to calculate the \textit{c2c} distance \cite{girardeaumontaut:pastel-00001745}, and \href{https://www.meshlab.net/}{MeshLab} to calculate the \textit{m2m} distance \cite{cignoni1998metro}.
During evaluation, we consider the raw values predicted by the models and clip the valid depth range to $10$m.

\input{Discussions/Losses}

\input{Discussions/Architecture}

\input{Discussions/Secondary}

\input{Discussions/Refinement}

\input{Discussions/Other}

\input{Discussions/Generalization}

\input{Discussions/Resolution}

\input{Discussions/3D}

%% file: Discussions/Losses.tex
\textbf{\textit{Which loss combination offers better performance?}}
Contrary to their focused nature both $\mathcal{L}_{cosine}$ and $\mathcal{L}_{grad}$ increase depth estimation performance across all models when complementing the direct $\mathcal{L}_1$ objective, as evident in Table~\ref{tab:losses}.
In addition, they provide the expected boost in smoothness/discontinuity preservation across all models as presented in our supplementary material which is appended after the references.
When viewed purely from a depth estimation perspective, it is observed that their combination, $\mathcal{L}_{comb}$, benefits performance.
But, when examining the specific depth traits that they seek to enforce, their conflicting nature is also apparent.
Overall, we observe almost all models achieve highest overall performance when both losses are present, with, our without the virtuan normal loss (VNL) which is added in the $\mathcal{L}_{vnl}$ case.
The latter greatly boosts the UNet model, which is reasonable as the localised nature of skip connections is aided by the global depth constraints that VNL introduces.

%% file: Discussions/Architecture.tex
\textbf{\textit{Which architecture is better performing?}}
We compare architectures after selecting the best performing models for each, which for UNet is $\mathcal{L}_{vnl}$, and for the rest the $\mathcal{L}_{comb}$.
The rationale for choosing $\mathcal{L}_{comb}$ for ResNet\textsubscript{\textit{skip}} is that while $\mathcal{L}_{grad}$ behaves better on direct depth metrics (except closer distances, as indicated by the RMSLE), there is a large performance gap in the discontinuity and smoothness metrics\footnotemark[2], compared to the performance discrepancy on depth estimation.
Table~\ref{tab:spherical} presents the results using the spherical metrics that account for the distortion.
These are unbiased metrics, which is evident given the deteriorated performance across all metrics compared to those estimated on the image level on each equirectangular panorama.
A more straightforward comparison is available in our supplementary material appended after the references. %
Interestingly, we observe that models employing encoder-decoder skip connections exhibit better performance both in direct depth metrics (Table~\ref{tab:spherical}).
Curiously, contrary to the expectation as set by the literature \cite{ranftl2020towards} that high-capacity encoders are required, the UNet architecture showcases the best performance.
Regarding domain oriented techniques, we train the better performing model with circular padding \cite{sun2019horizonnet,zioulis2021single} that connects features across the horizontal equirectangular boundary, denoted as \textit{circ}.
Evidently, this simple scheme increases the performance across all metrics, allowing the model to exploit its spherical nature.

%% file: Discussions/Secondary.tex
\textbf{\textit{Is this performance consistent when considering secondary traits?}}
Regarding the discontinuity and smoothness traits as presented in Table~\ref{tab:ob_smooth} it is evident that skip connections result in higher performance, but especially for the dominating UNet, at the expense of the smoothness trait.
This is reasonable as early layer skip connections result into texture transfer, and further evidenced by the improved performance of ResNet\textsubscript{\textit{skip}}, which lacks early layer skips, on both discontinuity and smoothness metrics.
Overall, UNet achieves the best performance on depth and discontinuity metrics at the expense of the smoothness trait and closer range performance as indicated by its inferior RMSLE.
On the other side, the different metrics indicate that the PNAS model produces oversmoothed results that are more metrically accurate and precise at closer distances.
Nonetheless, ResNet\textsubscript{skip} achieves a better balance without significant sacrifices across the secondary traits.

%% file: Discussions/Refinement.tex
\textbf{\textit{How helpful is depth refinement?}}
We also examine the effect of a shallow depth refinement module on these models, with the results after training for $10$ epochs presented in Table~\ref{tab:ob_smooth}.
All models, apart from UNet, improve their performance at boundary preservation while also preserving depth estimation performance, but at the expense of smoothness, with the exception in this case being ResNet\textsubscript{\textit{skip}}.
For UNet specifically, texture transfer leads to noise, which prevents an interpolation-based warping technique to improve results, as it was designed to improve smooth depth predictions.
However, ResNet\textsubscript{\textit{skip}} closes the performance gap and even improves smoothness performance, further solidifying its well-balanced nature.

%% file: Discussions/Other.tex
\textbf{\textit{Why this benchmark?}}
Table~\ref{tab:spherical} shows the performance of the two higher performing models when trained and tested on other recently introduced \360 depth datasets, namely 3D60 \cite{zioulis2019spherical} which is an extension of \cite{zioulis2018omnidepth} and Structured3D \cite{Structured3D} with $512 \times 256$ and $1024 \times 512$ resolutions respectively.
All metrics are significantly higher which evidences their insuitability to be used for benchmarking progress.
This is largely because of their inherent bias which is the result of lighting for 3D60, which includes an extra light source at the center, as also explained in \cite{jiang2021unifuse}, an unfortunate bias that models learn to exploit as farther depths are darker; and the omission of the noisy camera-based image formation process and lack of real-world scene complexity exhibited by the purely synthetic Structure3D dataset,

%% file: Discussions/Generalization.tex
\textbf{\textit{What is their generalization capacity?}}
We test these models in a zero-shot cross-dataset transfer setting using the GV2 splits using a subset of all metrics with the results presented in Table~\ref{tab:gv2}.
We observe reduced performance for all models across all splits which is the result applying these models in different contexts/scenes and to out-of-distribution depths (\textit{tiny}/\textit{medium}).
Yet, the ranking between models is not severely disrupted, indicating that architecture changes do not significantly affect generalization. 
The \textit{fullplus} split is noticeably harder than the others, as all metrics are considerably worse, showcasing that pure context shifts (similar depth distribution) are detrimental to performance.
However, camera domain shifts are another generalization barrier that is significant, as shown by the models' results on the \textit{filmic} splits, where a different color transfer function was applied during rendering.
The latter also received the bigger gains when training with photometric augmentation (UNet\textsubscript{\textit{aug}}), specifically random gamma, contrast, brightness and saturation shifts, which also boosted performance horizontally across all splits.
Still, augmentation alone did not raise performance to levels similar to the M3D test set, indicating that other techniques are required.

\input{Tables/Table3dmetrics}

%% file: Tables/Table3dmetrics.tex
\begin{table}[!htbp]
  \centering
  \caption{Performance of all models using the 3D metrics.
  For the \textit{m2m} metric inside the parentheses, we also report the percentage (\%) of the error w.r.t the bounding box diagonal, while for the \textit{c2c} metric we also report the error standard deviation.}
  \label{tab:3d}
  \footnotesize
    \begin{tabular}{l|c|c}
    \toprule
    \multicolumn{1}{c|}{\multirow{1}[1]{*}{model}} & \multicolumn{1}{c|}{\textit{m2m}} & \multicolumn{1}{c}{\textit{c2c}} \\
    \midrule
    Pnas\textsuperscript{$comb$} & 0.2502 (7.02\%) & 0.1439 (0.1881) \\
    UNet\textsuperscript{$vnl$} & \first{0.2397 (6.52\%)} & \second{0.1305 (0.1663)} \\
    DenseNet\textsuperscript{$comb$} & \third{0.2475 (6.98\%)} & 0.1425 (0.1852) \\
    ResNet\textsuperscript{$comb$} & 0.2573 (7.01\%) & \third{0.1405 (0.1907)} \\
    ResNet\rlap{\textsuperscript{$comb$}}\textsubscript{\textit{\textit{skip}}} & \second{0.2424 (6.83\%)} & \first{0.1300 (0.1770)} \\
    \bottomrule
    \end{tabular}%
    \normalsize
  \vspace{-0.15cm}
\end{table}

%% file: Discussions/Resolution.tex
\textbf{\textit{How does performance vary with resolution?}}
Given their \360 FoV, spherical panoramas require higher resolutions to be able to more robustly estimate detailed depth.
Table~\ref{tab:resolution} presents the results of the two better performing models, trained on M3D's $1024 \times 512$ resolution data, and tested on the GV2 splits with the same resolution.
We observe a change in performance between the UNet and the ResNet with skip connections.
The latter's expanded receptive field and higher capacity encoder offer significantly higher performance in the direct depth and smoothness metrics,  albeit the UNet still localizes boundaries better.

%% file: Discussions/3D.tex
\textbf{\textit{How about downstream application suitability?}}
We also assess each model's performance using the 3D metrics that aggregate performance across all axes.
Table~\ref{tab:3d} presents the results using the cloud and mesh distances are presented in Section~\ref{sec:methodology_metrics}.
Overall the performance ranking is preserved, with UNet's noisy predictions being moderated by the reconstruction process in the mesh distance metric, while the point cloud distance's nearest-neighbor nature is more sensitive to it.
Thus, downstream applications like view synthesis should investigate model results using \textit{c2c} metrics, while applications relying on 3D reconstruction should resort to the \textit{m2m} metric.
Again, as shown by these metrics, the skip connections based ResNet is a reasonably balanced choice, that follows UNet's top performance.

%% file: Discussion.tex
Spherical depth estimation is a task that comes with certain advantages (holistic view) and disadvantages (resolution requirements) compared to traditional -- perspective -- depth estimation.
Preserving boundaries is challenging because of the distortion frequently squeezing objects towards the equator, and thus, smaller spatial areas; and due to the discontinuities that the different projections introduce.
Imposing a smoothness prior is also not straightforward as for perspective depth.
The presented Pano3D benchmark can stimulate future progress in \360 depth estimation that will take all these aspects into account.
From our extensive analysis -- which nonetheless does not cover all cases -- we identify the effectiveness of skip connections in terms of boundary preservation, as a means to overcome the weakness of spatial downscaling, which in turn, is necessary to exploit the panoramas' global context.
While the UNet architecture achieves top performance in lower resolutions, a ResNet with skip connections is a more balanced architectural choice that scales better across resolutions. 

Finally, Pano3D relies on zero-shot cross-dataset transfer to move beyond a simple train/test split performance comparison.
By decomposing generalization into three distinct performance reducing barriers, our goal is better facilitate the assessment towards real-world applicability of data-driven models for \360 geometric inference.

\section*{Supplementary Material.}
We provide extra quantitative and qualitative comparisons in the supplementary material following the references.
Supplementing experiments also reproduce prior work used as a basis for designing our methodology.
Finally, a live web demo of our baseline models can be found at \href{https://share.streamlit.io/tzole1155/ThreeDit}{share.streamlit.io/tzole1155/ThreeDit}.

%% file: Tables/Table1_supp.tex
\begin{table*}[!htbp]
\centering
\caption{Three axis depth metrics performance across models and supervision schemes. Best three performers are denoted with bold faced \colorbox{lightgreen}{\textbf{light green}} (1\textsuperscript{st}), \colorbox{lightblue}{light blue} (2\textsuperscript{nd}) and \colorbox{lightpurple}{light purple} (3\textsuperscript{rd}) respectively following the ranking order.
Same scheme applies to all tables.}
\label{tab:Allloses}
\resizebox{\textwidth}{!}{%
\begin{tabular}{ll|ccccccccc|cccccccc|cccc}
\multicolumn{2}{c|}{\multirow{3}{*}{Model}}                                          & \multicolumn{9}{c|}{Direct Depth}                                                                                                                                                                                                       & \multicolumn{8}{c|}{Depth Discontinuity}                                                                                                                                                                                                                                                                             & \multicolumn{4}{c}{Depth Smoothness}                                                                              \\
\multicolumn{2}{c|}{}                                                                & \multicolumn{4}{c}{Error $\downarrow$}                                                             & \multicolumn{5}{c|}{Accuracy $\uparrow$}                                                                                           & \multicolumn{2}{c}{Error $\downarrow$}                                                      & \multicolumn{6}{c|}{Accuracy $\uparrow$}                                                                                                                                                                               & Error $\downarrow$                                    & \multicolumn{3}{c}{Accuracy $\uparrow$}                   \\
\multicolumn{2}{c|}{}                                                                & \textit{$w$RMSE} & \textit{$w$RMSLE} & \textit{$w$AbsRel} & \multicolumn{1}{c|}{\textit{$w$SqRel}} & $\delta_{1.05}^{ico^6}$ & $\delta_{1.1}^{ico^6}$ & $\delta_{1.25}^{ico^6}$ & $\delta_{1.25^2}^{ico^6}$ & $\delta_{1.25^3}^{ico^6}$ & \textit{dbe}\textsuperscript{acc} & \multicolumn{1}{c|}{\textit{dbe}\textsuperscript{comp}} & \textit{prec}\textsubscript{$0.25$} & \textit{prec}\textsubscript{$0.5$} & \textit{prec}\textsubscript{$1$} & \textit{rec}\textsubscript{$0.25$} & \textit{rec}\textsubscript{$0.5$} & \textit{rec}\textsubscript{$1$} & \multicolumn{1}{c|}{\textit{RMSE\textsuperscript{o}}} & $\alpha_{11.25^o}$ & $\alpha_{22.5^o}$ & $\alpha_{30^o}$  \\ \hline
\multirow{5}{*}{\rotatebox{90}{Pnas}}                       & $\mathcal{L}_{1}$      & 0.5606           & 0.0854            & 0.1328             & 0.1196                                 & 32.69\%                 & 56.94\%                & 85.12\%                 & 95.38\%                   & 97.95\%                   & 2.6542                            & 5.7303                                                  & 38.73\%                             & 30.26\%                            & 23.58\%                          & 18.74\%                            & 10.48\%                           & 8.48\%                          & 20.12                                                 & 53.88\%            & 69.81\%           & 75.65\%          \\
                                                            & $\mathcal{L}_{cosine}$ & 0.5622           & 0.0858            & 0.1338             & 0.1317                                 & 34.49\%                 & 58.06\%                & 85.52\%                 & 95.44\%                   & 97.88\%                   & 2.7194                            & 5.4964                                                  & 36.16\%                             & 27.76\%                            & 22.22\%                          & 21.48\%                            & \second{13.55\%}                  & \second{10.02\%}                & \third{15.70}                                         & \second{67.14\%}   & \second{77.37\%}  & \third{81.05\%}  \\
                                                            & $\mathcal{L}_{grad}$   & \second{0.5374}  & \third{0.0822}    & \second{0.1276}    & \first{0.1146}                         & \second{35.68\%}        & \third{59.54\%}        & \third{86.41\%}         & \third{95.72\%}           & \third{98.04\%}           & \first{2.5008}                    & \third{5.4548}                                          & \first{40.91\%}                     & \first{32.05\%}                    & \second{25.27\%}                 & \second{22.49\%}                   & 12.17\%                           & \third{9.07\%}                  & 18.12                                                 & 59.56\%            & 73.24\%           & 78.30\%          \\
                                                            & $\mathcal{L}_{comb}$   & \first{0.5367}   & \first{0.0811}    & \first{0.1259}     & \second{0.1153}                        & \first{36.44\%}         & \first{60.52\%}        & \first{86.80\%}         & \first{95.83\%}           & \first{98.11\%}           & \second{2.5119}                   & \first{5.3501}                                          & \third{39.83\%}                     & \third{31.59\%}                    & \first{27.01\%}                  & \first{23.53\%}                    & \first{14.42\%}                   & \first{10.98\%}                 & \first{15.26}                                         & \first{67.73\%}    & \first{77.99\%}   & \first{81.67\%}  \\
                                                            & $\mathcal{L}_{vnl}$    & \third{0.5403}   & \second{0.0815}   & \third{0.1280}     & \third{0.1183}                         & \third{35.43\%}         & \second{59.72\%}       & \second{86.58\%}        & \second{95.79\%}          & \second{98.11\%}          & \third{2.5141}                    & \second{5.3893}                                         & \second{40.14\%}                    & \second{31.77\%}                   & \third{24.47\%}                  & \third{22.14\%}                    & \third{12.69\%}                   & 8.74\%                          & \second{15.57}                                        & \third{66.61\%}    & \third{77.34\%}   & \second{81.23\%} \\ \hline
\multirow{5}{*}{\rotatebox{90}{UNet}}                       & $\mathcal{L}_{1}$      & 0.4834           & 0.2361            & \third{0.1211}     & 0.0913                                 & \third{35.18\%}         & \third{58.24\%}        & 86.80\%                 & 96.45\%                   & 98.43\%                   & 1.4011                            & 4.3152                                                  & 57.59\%                             & \third{58.00\%}                    & \third{53.85\%}                  & 38.74\%                            & \third{31.57\%}                   & \second{24.31\%}                & 24.66                                                 & 36.80\%            & 60.60\%           & 69.73\%          \\
                                                            & $\mathcal{L}_{cosine}$ & 0.4736           & \first{0.0906}    & 0.1217             & 0.0891                                 & 32.65\%                 & 58.04\%                & \second{87.40\%}        & \second{96.68\%}          & \third{98.61\%}           & 1.4513                            & 5.0455                                                  & 55.35\%                             & 52.16\%                            & 46.01\%                          & 39.36\%                            & 30.01\%                           & 21.69\%                         & \first{15.80}                                         & \first{63.10\%}    & \first{77.60\%}   & \first{82.38\%}  \\
                                                            & $\mathcal{L}_{grad}$   & \third{0.4659}   & 0.5186            & \second{0.1209}    & \second{0.0833}                        & \second{35.25\%}        & \second{58.79\%}       & \third{87.33\%}         & 96.56\%                   & 98.45\%                   & \third{1.3305}                    & \second{4.0582}                                         & \second{63.27\%}                    & \first{63.13\%}                    & \first{56.54\%}                  & \second{40.39\%}                   & \second{32.47\%}                  & \third{23.37\%}                 & 19.52                                                 & 52.23\%            & 70.40\%           & 76.91\%          \\
                                                            & $\mathcal{L}_{comb}$   & \second{0.4630}  & \third{0.1690}    & 0.1222             & \third{0.0847}                         & 34.79\%                 & 58.21\%                & 87.08\%                 & \third{96.63\%}           & \second{98.66\%}          & \second{1.3077}                   & \third{4.2080}                                          & \first{63.31\%}                     & \second{61.74\%}                   & \second{54.96\%}                 & \third{39.38\%}                    & 30.27\%                           & 22.00\%                         & \third{16.19}                                         & \third{61.01\%}    & \third{76.18\%}   & \third{81.45\%}  \\
                                                            & $\mathcal{L}_{vnl}$    & \first{0.4520}   & \second{0.1300}   & \first{0.1147}     & \first{0.0811}                         & \first{36.68\%}         & \first{60.59\%}        & \first{88.31\%}         & \first{96.96\%}           & \first{98.73\%}           & \first{1.2699}                    & \first{3.8876}                                          & \third{58.97\%}                     & 57.54\%                            & 51.85\%                          & \first{43.96\%}                    & \first{36.69\%}                   & \first{28.59\%}                 & \second{16.02}                                        & \second{61.80\%}   & \second{76.58\%}  & \second{81.70\%} \\ \hline
\multirow{5}{*}{\rotatebox{90}{DenseNet}}                   & $\mathcal{L}_{1}$      & 0.5441           & 0.6872            & 0.1348             & 0.1144                                 & 34.34\%                 & 57.10\%                & 84.73\%                 & 95.28\%                   & 97.69\%                   & 2.3690                            & 5.5135                                                  & 40.40\%                             & 36.07\%                            & 28.78\%                          & 20.45\%                            & 11.54\%                           & 8.05\%                          & 21.08                                                 & 49.98\%            & 68.29\%           & 74.78\%          \\
                                                            & $\mathcal{L}_{cosine}$ & 0.5361           & \first{0.0822}    & \second{0.1239}    & \third{0.1034}                         & \third{34.98\%}         & \third{59.34\%}        & \third{86.36\%}         & \second{95.94\%}          & \first{98.13\%}           & 2.3486                            & 5.3702                                                  & 41.01\%                             & 35.45\%                            & 29.10\%                          & 22.80\%                            & \third{14.19\%}                   & \third{9.39\%}                  & \first{15.97}                                         & \first{64.92\%}    & \first{76.91\%}   & \second{81.15\%} \\
                                                            & $\mathcal{L}_{grad}$   & \first{0.5202}   & \third{0.4655}    & 0.1304             & 0.1045                                 & 32.68\%                 & 57.59\%                & 85.69\%                 & \third{95.85\%}           & \third{98.06\%}           & \third{2.0789}                    & \third{5.2159}                                          & \second{47.01\%}                    & \second{40.61\%}                   & \second{33.32\%}                 & \third{23.68\%}                    & 13.71\%                           & 9.35\%                          & 18.90                                                 & 56.86\%            & 71.79\%           & 77.23\%          \\
                                                            & $\mathcal{L}_{comb}$   & \second{0.5209}  & \second{0.1982}   & \first{0.1209}     & \first{0.1013}                         & \second{35.97\%}        & \first{60.41\%}        & \first{87.02\%}         & \first{95.96\%}           & \second{98.09\%}          & \second{2.0628}                   & \second{5.0977}                                         & \first{47.16\%}                     & \first{40.77\%}                    & \first{35.20\%}                  & \first{26.09\%}                    & \first{16.87\%}                   & \first{12.21\%}                 & \second{15.98}                                        & \second{64.58\%}   & \second{76.86\%}  & \first{81.20\%}  \\
                                                            & $\mathcal{L}_{vnl}$    & \third{0.5232}   & 0.7560            & \third{0.1258}     & \second{0.1030}                        & \first{36.28\%}         & \second{60.04\%}       & \second{86.61\%}        & 95.66\%                   & 97.74\%                   & \first{2.0525}                    & \first{5.0931}                                          & \third{44.81\%}                     & \third{40.14\%}                    & \third{32.30\%}                  & \second{25.21\%}                   & \second{15.71\%}                  & \second{10.33\%}                & \third{16.51}                                         & \third{63.43\%}    & \third{76.02\%}   & \third{80.53\%}  \\ \hline
\multirow{5}{*}{\rotatebox{90}{ResNet}}                     & $\mathcal{L}_{1}$      & 0.5500           & \third{0.1922}    & 0.1394             & 0.1186                                 & 30.59\%                 & 54.17\%                & 84.07\%                 & 95.47\%                   & 98.03\%                   & 2.4386                            & \third{5.7688}                                          & 39.10\%                             & 31.69\%                            & 23.28\%                          & 20.92\%                            & 10.24\%                           & 6.32\%                          & 22.83                                                 & 44.68\%            & 64.51\%           & 72.02\%          \\
                                                            & $\mathcal{L}_{cosine}$ & \third{0.5435}   & \first{0.0864}    & \second{0.1364}    & 0.1194                                 & \first{34.77\%}         & \second{56.32\%}       & \third{84.29\%}         & \third{95.64\%}           & \second{98.11\%}          & 2.6918                            & 5.7928                                                  & 38.35\%                             & 32.13\%                            & \third{26.82\%}                  & \third{21.88\%}                    & \second{12.61\%}                  & \second{8.71\%}                 & \first{16.37}                                         & \first{64.24\%}    & \first{76.30\%}   & \first{80.63\%}  \\
                                                            & $\mathcal{L}_{grad}$   & 0.5475           & 0.2976            & 0.1387             & \third{0.1151}                         & \third{32.43\%}         & 54.46\%                & 83.76\%                 & 95.37\%                   & 97.97\%                   & \third{2.4112}                    & 5.7959                                                  & \third{41.87\%}                     & \third{33.23\%}                    & 21.60\%                          & 21.31\%                            & 9.27\%                            & 4.95\%                          & 20.50                                                 & 52.77\%            & 68.97\%           & 75.00\%          \\
                                                            & $\mathcal{L}_{comb}$   & \first{0.5294}   & \second{0.1365}   & \third{0.1374}     & \second{0.1127}                        & 32.03\%                 & \third{55.31\%}        & \second{84.74\%}        & \second{95.81\%}          & \first{98.21\%}           & \second{2.2393}                   & \second{5.3796}                                         & \second{44.10\%}                    & \second{36.70\%}                   & \second{27.44\%}                 & \second{22.91\%}                   & \third{12.23\%}                   & \third{7.20\%}                  & \second{16.63}                                        & \second{63.09\%}   & \second{75.70\%}  & \second{80.20\%} \\
                                                            & $\mathcal{L}_{vnl}$    & \second{0.5324}  & 0.3320            & \first{0.1301}     & \first{0.1070}                         & \second{33.60\%}        & \first{57.50\%}        & \first{85.20\%}         & \first{95.83\%}           & \third{98.07\%}           & \first{2.1335}                    & \first{5.1866}                                          & \first{45.00\%}                     & \first{38.70\%}                    & \first{30.85\%}                  & \first{24.88\%}                    & \first{14.43\%}                   & \first{9.28\%}                  & \third{17.07}                                         & \third{61.99\%}    & \third{75.22\%}   & \third{79.91\%}  \\ \hline
\multirow{5}{*}{\rotatebox{90}{ResNet\textsubscript{\textit{skip}}}} & $\mathcal{L}_{1}$      & 0.5041           & 0.2924            & 0.1259             & 0.0977                                 & 34.10\%                 & 57.64\%                & 86.05\%                 & 96.13\%                   & 98.30\%                   & 1.5462                            & 4.7640                                                  & 49.48\%                             & 47.23\%                            & 43.31\%                          & 32.86\%                            & 23.57\%                           & \third{16.63\%}                 & 22.30                                                 & 44.07\%            & 65.82\%           & 73.55\%          \\
                                                            & $\mathcal{L}_{cosine}$ & 0.5024           & \third{0.1207}    & 0.1208             & 0.0958                                 & \second{37.15\%}        & 59.61\%                & 87.03\%                 & 96.34\%                   & 98.35\%                   & 1.6012                            & 4.7078                                                  & 52.83\%                             & 49.23\%                            & 41.05\%                          & 32.03\%                            & \third{23.82\%}                   & \second{16.75\%}                & \third{15.76}                                         & \second{63.32\%}   & \second{77.05\%}  & \third{81.83\%}  \\
                                                            & $\mathcal{L}_{grad}$   & \first{0.4754}   & 0.3274            & \second{0.1183}    & \second{0.0905}                        & \third{36.23\%}         & \second{60.44\%}       & \second{87.96\%}        & \first{96.62\%}           & \third{98.45\%}           & \third{1.5013}                    & \second{4.4831}                                         & \second{56.27\%}                    & \first{54.26\%}                    & \first{47.88\%}                  & \third{33.96\%}                    & 23.52\%                           & 16.07\%                         & 18.72                                                 & 55.00\%            & 71.76\%           & 77.82\%          \\
                                                            & $\mathcal{L}_{comb}$   & \second{0.4788}  & \first{0.0927}    & \first{0.1166}     & \first{0.0893}                         & 36.20\%                 & \first{60.64\%}        & \first{87.99\%}         & \second{96.62\%}          & \first{98.49\%}           & \second{1.4883}                   & \third{4.5346}                                          & \first{57.34\%}                     & \second{54.11\%}                   & \second{47.57\%}                 & \second{33.99\%}                   & \second{24.30\%}                  & 16.37\%                         & \first{15.27}                                         & \first{64.18\%}    & \first{77.57\%}   & \first{82.27\%}  \\
                                                            & $\mathcal{L}_{vnl}$    & \third{0.4923}   & \second{0.1095}   & \third{0.1197}     & \third{0.0941}                         & \first{37.55\%}         & \third{60.43\%}        & \third{87.23\%}         & \third{96.42\%}           & \second{98.46\%}          & \first{1.4629}                    & \first{4.1408}                                          & \third{54.99\%}                     & \third{51.98\%}                    & \third{45.40\%}                  & \first{35.29\%}                    & \first{25.22\%}                   & \first{17.68\%}                 & \second{15.67}                                        & \third{63.28\%}    & \third{77.05\%}   & \second{81.94\%} \\ \hline
\end{tabular}%
}
\end{table*}

%% file: Tables/Table2_supp.tex
\begin{table*}[!htbp]
\centering
\caption{Direct depth performance using spherical and conventional metrics. Bottom part results are the same as those presented in Table 2 of the original document. Top part are the corresponding results from Table 1 of the original manuscript.}
\label{tab:table2_oldandsperical}
\small
\begin{tabular}{l|cccc|ccccc}
\hline
\multicolumn{1}{c|}{\multirow{2}{*}{Model}}                    & \multicolumn{4}{c|}{Depth Error $\downarrow$}                                 & \multicolumn{5}{c}{Depth Accuracy $\uparrow$}                                                                                      \\
\multicolumn{1}{c|}{}                                          & \textit{RMSE}    & \textit{RMSLE}    & \textit{AbsRel}    & \textit{SqRel}    & $\delta_{1.05}$         & $\delta_{1.1}$         & $\delta_{1.25}$         & $\delta_{1.25^2}$         & $\delta_{1.25^3}$         \\ \hline
Pnas\textsuperscript{$comb$}                                   & 0.4613           & \first{0.0740}    & 0.1143             & 0.0892            & \second{38.56\%}        & \first{63.31\%}        & \third{88.70\%}         & \third{96.68\%}           & 98.62\%                   \\
UNet\textsuperscript{$vnl$}                                    & \first{0.3967}   & \third{0.1182}    & \first{0.1095}     & \first{0.0672}    & \first{38.62\%}         & \third{62.16\%}        & \second{89.08\%}        & \first{97.35\%}           & \first{99.03\%}           \\
DenseNet\textsuperscript{$comb$}                               & \third{0.4490}   & 0.2565            & \third{0.1129}     & \third{0.0806}    & \third{38.30\%}         & \second{63.02\%}       & 88.56\%                 & 96.66\%                   & 98.54\%                   \\
ResNet\textsuperscript{$comb$}                                 & 0.4573           & 0.1200            & 0.1272             & 0.0894            & 34.53\%                 & 57.97\%                & 86.26\%                 & 96.56\%                   & \third{98.71\%}           \\
ResNet\rlap{\textsuperscript{$comb$}}\textsubscript{\textit{skip}} & \second{0.4165}  & \second{0.0843}   & \second{0.1102}    & \second{0.0722}   & 36.71\%                 & 61.92\%                & \first{89.17\%}         & \second{97.24\%}          & \second{98.90\%}          \\ \hline
                                                               & \textit{$w$RMSE} & \textit{$w$RMSLE} & \textit{$w$AbsRel} & \textit{$w$SqRel} & $\delta_{1.05}^{ico^6}$ & $\delta_{1.1}^{ico^6}$ & $\delta_{1.25}^{ico^6}$ & $\delta_{1.25^2}^{ico^6}$ & $\delta_{1.25^3}^{ico^6}$ \\ \hline
Pnas\textsuperscript{$comb$}                                   & 0.5367           & \first{0.0811}    & 0.1259             & 0.1153            & \second{36.44\%}        & \third{60.52\%}        & 86.80\%                 & 95.83\%                   & 98.11\%                   \\
UNet\textsuperscript{$vnl$}                                    & \first{0.4520}   & \third{0.1300}    & \first{0.1147}     & \first{0.0811}    & \first{36.68\%}         & \second{60.59\%}       & \first{88.31\%}         & \first{96.96\%}           & \first{98.73\%}           \\
DenseNet\textsuperscript{$comb$}                               & \third{0.5209}   & 0.1982            & \third{0.1209}     & \third{0.1013}    & 35.97\%                 & 60.41\%                & \third{87.02\%}         & \third{95.96\%}           & 98.09\%                   \\
ResNet\textsuperscript{$comb$}                                 & 0.5294           & 0.1365            & 0.1374             & 0.1127            & 32.03\%                 & 55.31\%                & 84.74\%                 & 95.81\%                   & \third{98.21\%}           \\
ResNet\rlap{\textsuperscript{$comb$}}\textsubscript{\textit{skip}} & \second{0.4788}  & \second{0.0927}   & \second{0.1166}    & \second{0.0893}   & \third{36.20\%}         & \first{60.64\%}        & \second{87.99\%}        & \second{96.62\%}          & \second{98.49\%}          \\ \hline
\end{tabular}%
\normalsize
\end{table*}

%% file: Figures/sup_fig_1.tex
\begin{figure*}[!htbp]

\centering

\subfloat{\includegraphics[width=0.33\linewidth]{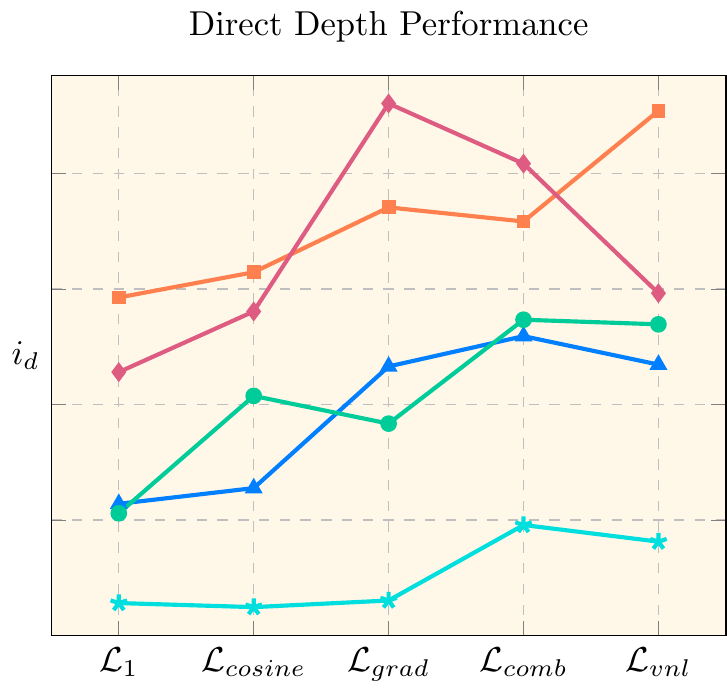}}
\subfloat{\includegraphics[width=0.33\linewidth]{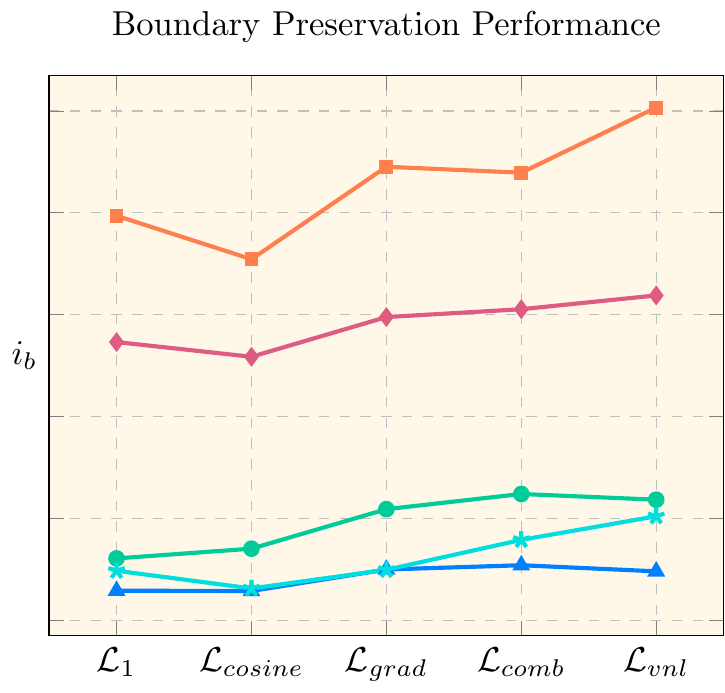}}
\subfloat{\includegraphics[width=0.33\linewidth]{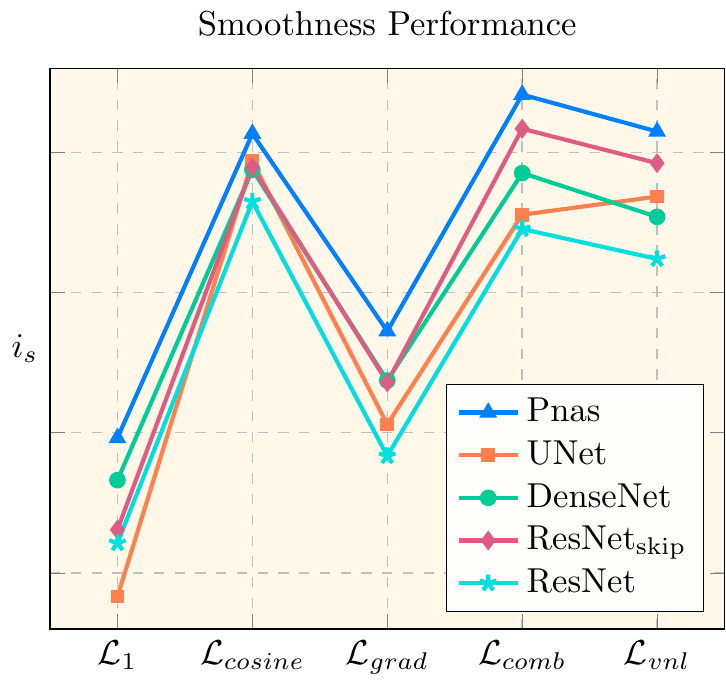}}

\caption{Performance indicators (higher is better) of different loss functions per model in three different axis. 
From left to right: depth indicator $i_d$, boundary indicator $i_b$ and smoothness indicator $i_s$.}

\label{fig:losses}

\end{figure*}

%% file: Tables/Table3_supp_updt.tex
\begin{table*}[htbp]
  \centering
\caption{Direct depth performance metrics across different variations of  DenseNet and Pnas.}
\label{tab:lossesablation}
    \begin{tabular}{c|c|c|cccc|c}
    \toprule
          &       &       & \multicolumn{4}{c|}{Depth Error $\downarrow$} & Depth Accuracy $\uparrow$ \\
    \midrule
    model & pretrained & $\mathcal{L}$ & \textit{RMSE} & \textit{RMSLE} & \textit{AbsRel} & \textit{SqRel} & $\delta_{1.25}$ \\
    \midrule
    \multirow{4}[4]{*}{\rotatebox{90}{DenseNet}} & \xmark & $\mathcal{L}_{1}$ & 0.4672 & 0.5580 & 0.1223 & 0.0896 & 86.72\% \\
\cmidrule{2-8}          & \cmark & $\mathcal{L}_{1}$ & \first{0.4072} & \first{0.3194} & \first{0.1140} & \first{0.0694} & \first{88.91\%} \\
          & \cmark & $\mathcal{L}_{log}$ & \third{0.5597} & \third{0.5720} & \third{0.1528} & \third{0.4475} & \third{80.48\%} \\
          & \cmark & $\mathcal{L}_{berHu}$ & \second{0.4532} & \second{0.3754} & \second{0.1228} & \second{0.0852} & \second{86.68\%} \\
    \midrule
    \multirow{4}[4]{*}{\rotatebox{90}{Pnas}} & \xmark & $\mathcal{L}_{1}$ & 0.4817 & 0.0780 & 0.1213 & 0.0933 & 87.25\% \\
\cmidrule{2-8}          & \cmark & $\mathcal{L}_{1}$ & \first{0.3998} & \first{0.0634} & \first{0.0975} & \first{0.0661} & \first{91.91\%} \\
          & \cmark & $\mathcal{L}_{log}$ & \third{0.4135} & \third{0.0656} & \third{0.0999} & \third{0.0697} & \third{91.09\%} \\
          & \cmark & $\mathcal{L}_{berHu}$ & \second{0.4059} & \second{0.0643} & \second{0.0992} & \second{0.0666} & \second{91.56\%} \\
    \bottomrule
    \end{tabular}%
\end{table*}%

%% file: Figures/adv_res_reskip.tex
\begin{figure*}[!htbp]

\centering

\subfloat{\includegraphics[width=0.2\linewidth]{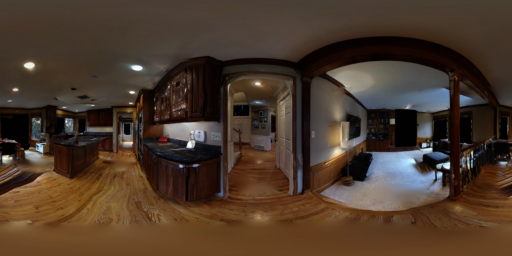}}
\subfloat{\includegraphics[width=0.2\linewidth]{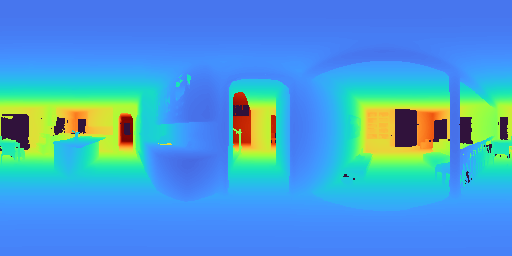}}
\subfloat{\includegraphics[width=0.2\linewidth]{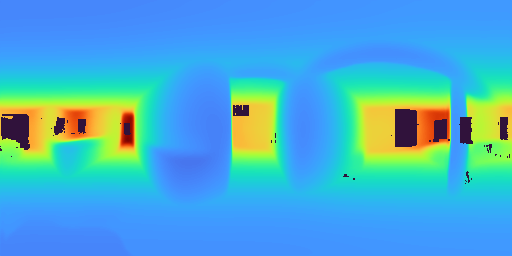}}
\subfloat{\includegraphics[width=0.2\linewidth]{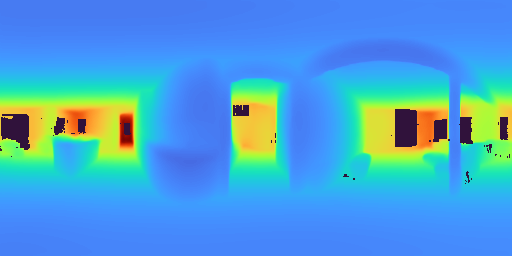}}
\subfloat{\includegraphics[width=0.2\linewidth]{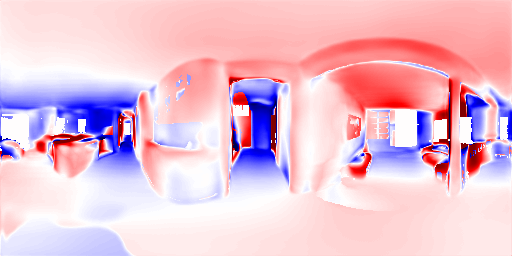}}
\\[-2.5ex]

\subfloat{\includegraphics[width=0.2\linewidth]{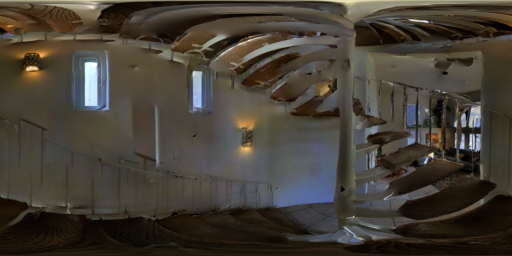}}
\subfloat{\includegraphics[width=0.2\linewidth]{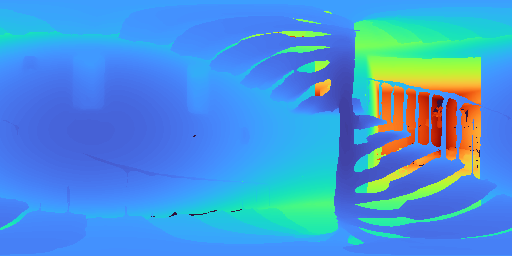}}
\subfloat{\includegraphics[width=0.2\linewidth]{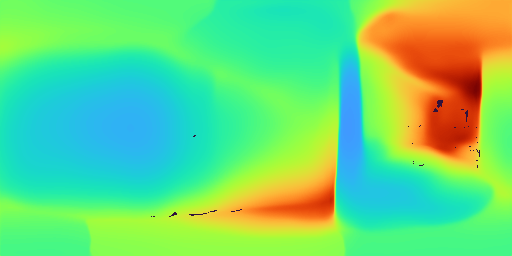}}
\subfloat{\includegraphics[width=0.2\linewidth]{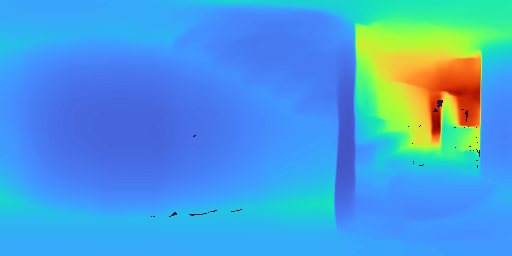}}
\subfloat{\includegraphics[width=0.2\linewidth]{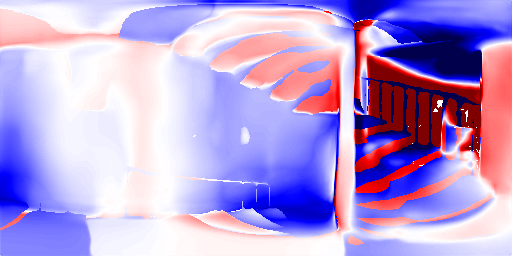}}
\\[-2.5ex]

\subfloat{\includegraphics[width=0.2\linewidth]{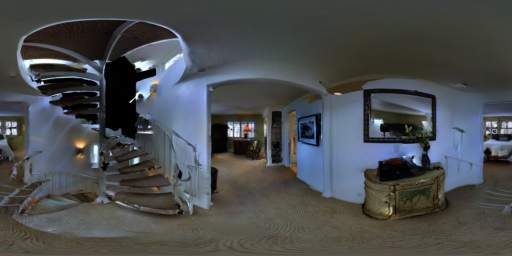}}
\subfloat{\includegraphics[width=0.2\linewidth]{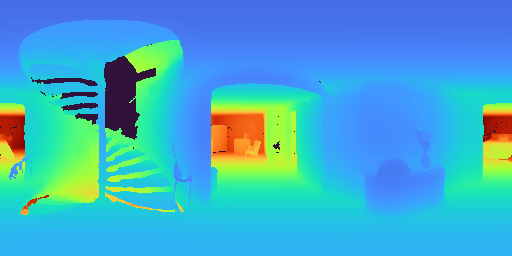}}
\subfloat{\includegraphics[width=0.2\linewidth]{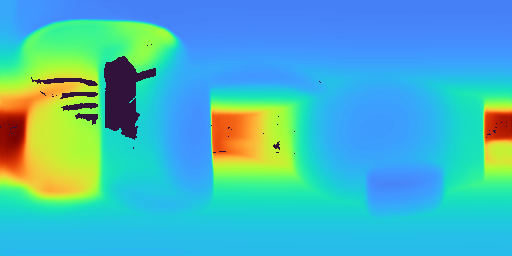}}
\subfloat{\includegraphics[width=0.2\linewidth]{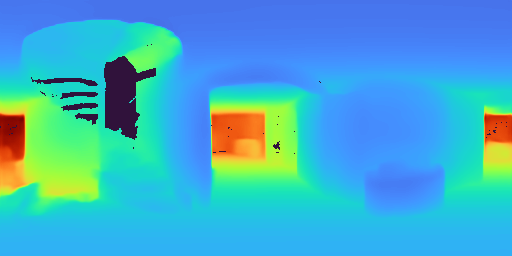}}
\subfloat{\includegraphics[width=0.2\linewidth]{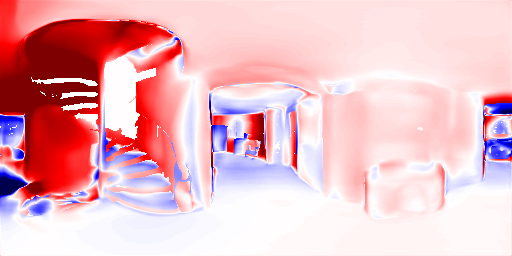}}
\\[-2.5ex]

\subfloat{\includegraphics[width=0.2\linewidth]{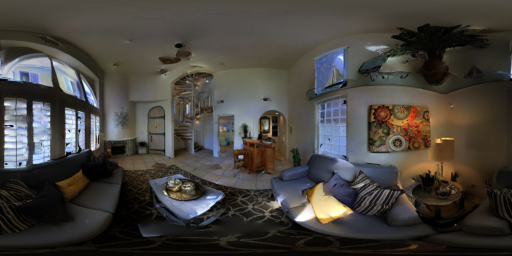}}
\subfloat{\includegraphics[width=0.2\linewidth]{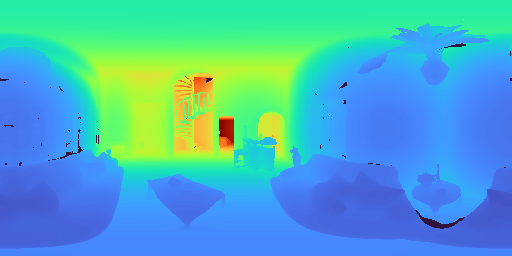}}
\subfloat{\includegraphics[width=0.2\linewidth]{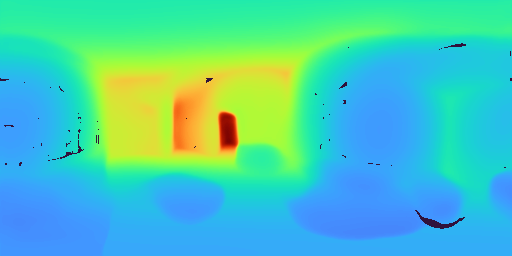}}
\subfloat{\includegraphics[width=0.2\linewidth]{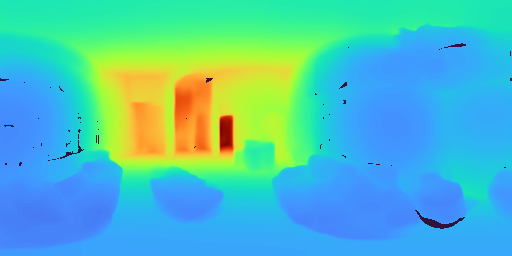}}
\subfloat{\includegraphics[width=0.2\linewidth]{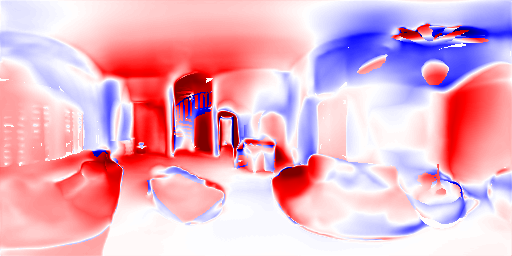}}
\\[-2.5ex]

\setcounter{subfigure}{0}%

\subfloat[Color]{\includegraphics[width=0.2\linewidth]{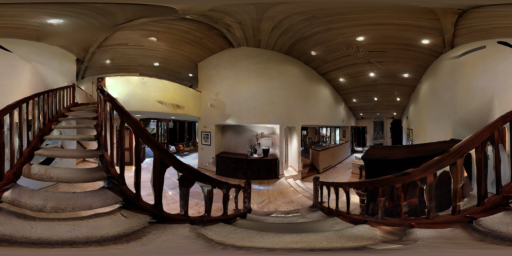}}
\subfloat[Ground Truth Depth]{\includegraphics[width=0.2\linewidth]{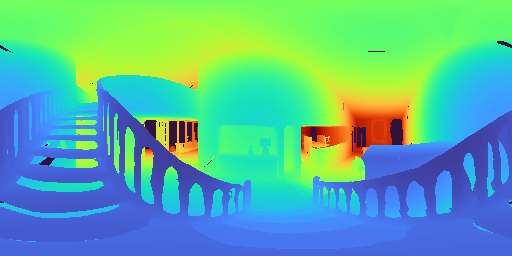}}
\subfloat[ResNet Depth]{\includegraphics[width=0.2\linewidth]{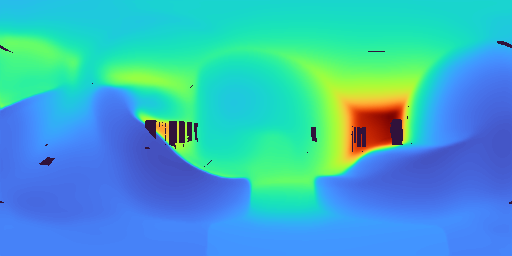}}
\subfloat[ResNet\textsubscript{skip} Depth]{\includegraphics[width=0.2\linewidth]{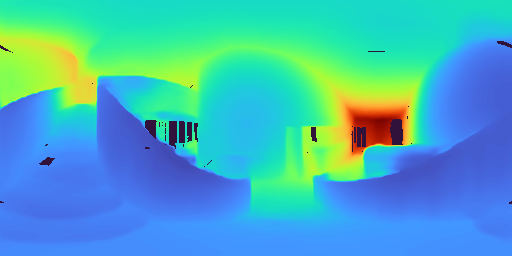}}
\subfloat[Depth Advantage]{\includegraphics[width=0.2\linewidth]{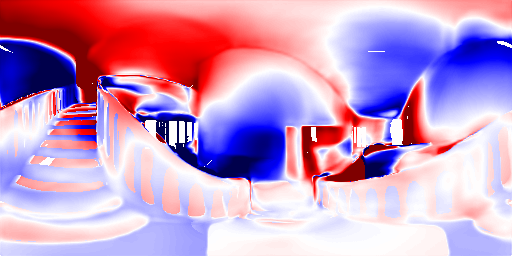}}

\caption{
Qualitative comparison between the ResNet and ResNet\textsubscript{skip} architectures.
On the right the advantage visualization shows with \textcolor{blue}{blue} color the areas where the former performs better, and with \textcolor{red}{red} color the areas where the latter performs better.
The color magnitude corresponds to the MAE difference between the two models, illustrating the performance deviation between the two models.
The addition of skip connections allows ResNet\textsubscript{skip} to capture finer-grained details.
}

\label{fig:adv_res_reskip}

\end{figure*}

%% file: Figures/adv_unet_pnas.tex
\begin{figure*}[!htbp]

\centering

\subfloat{\includegraphics[width=0.2\linewidth]{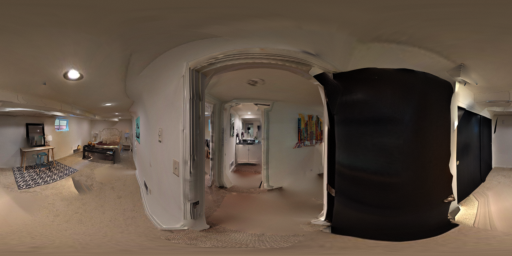}}
\subfloat{\includegraphics[width=0.2\linewidth]{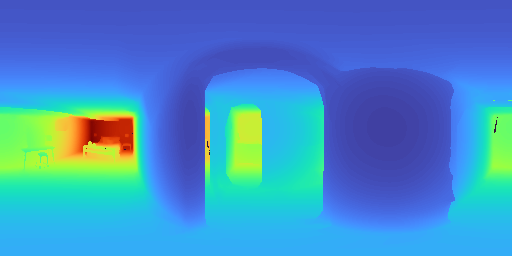}}
\subfloat{\includegraphics[width=0.2\linewidth]{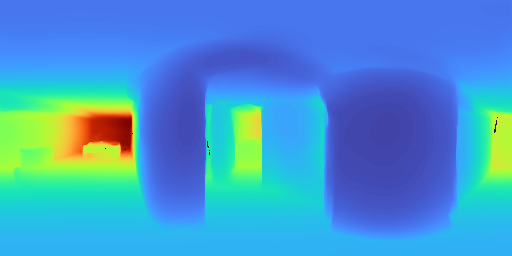}}
\subfloat{\includegraphics[width=0.2\linewidth]{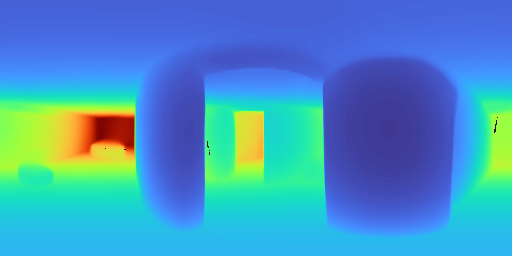}}
\subfloat{\includegraphics[width=0.2\linewidth]{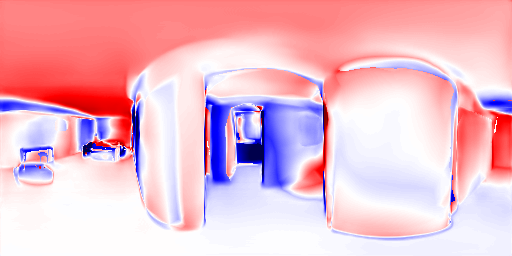}}
\\[-2.5ex]

\subfloat{\includegraphics[width=0.2\linewidth]{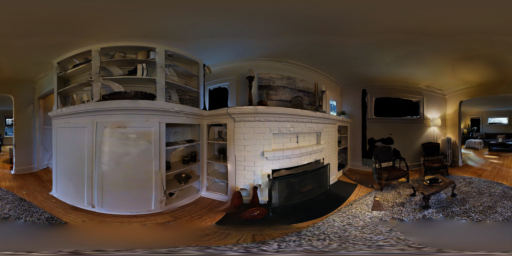}}
\subfloat{\includegraphics[width=0.2\linewidth]{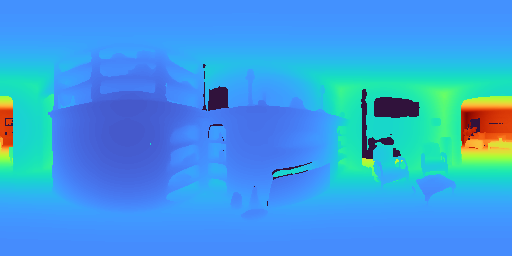}}
\subfloat{\includegraphics[width=0.2\linewidth]{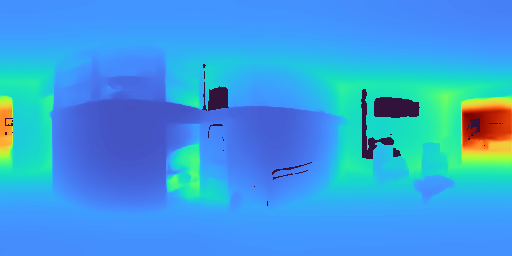}}
\subfloat{\includegraphics[width=0.2\linewidth]{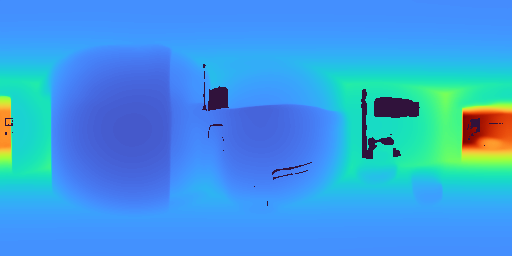}}
\subfloat{\includegraphics[width=0.2\linewidth]{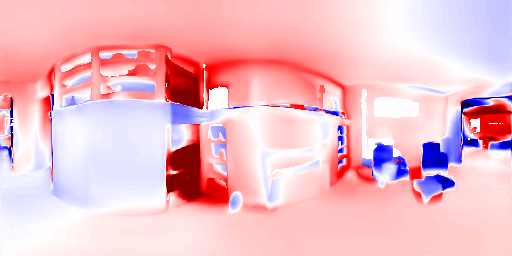}}
\\[-2.5ex]

\subfloat{\includegraphics[width=0.2\linewidth]{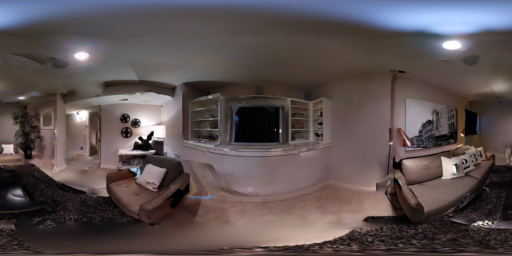}}
\subfloat{\includegraphics[width=0.2\linewidth]{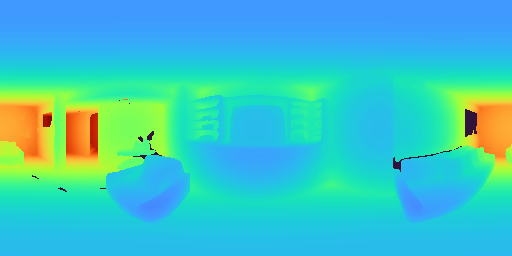}}
\subfloat{\includegraphics[width=0.2\linewidth]{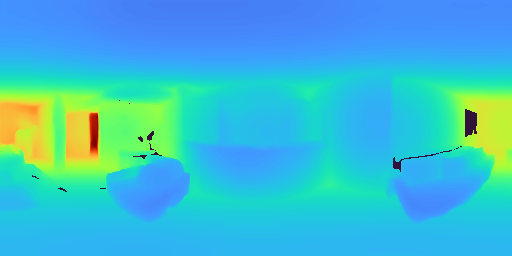}}
\subfloat{\includegraphics[width=0.2\linewidth]{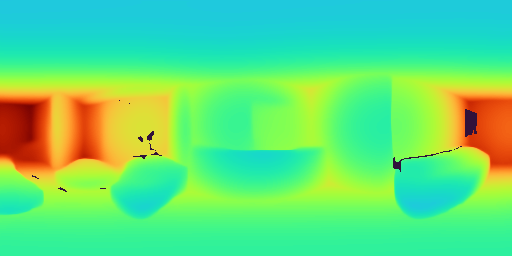}}
\subfloat{\includegraphics[width=0.2\linewidth]{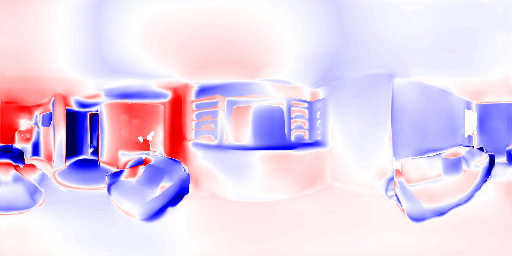}}
\\[-2.5ex]

\subfloat{\includegraphics[width=0.2\linewidth]{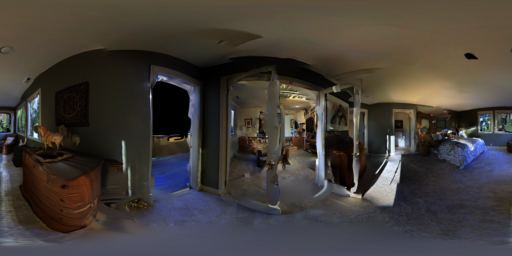}}
\subfloat{\includegraphics[width=0.2\linewidth]{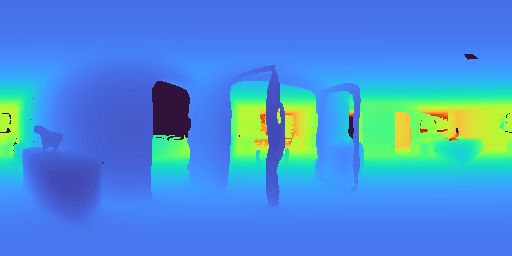}}
\subfloat{\includegraphics[width=0.2\linewidth]{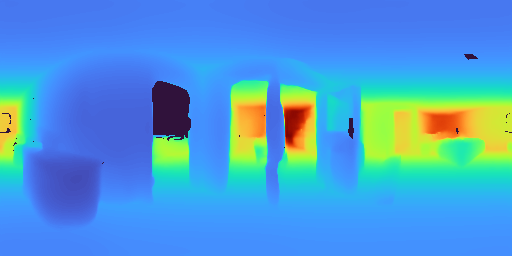}}
\subfloat{\includegraphics[width=0.2\linewidth]{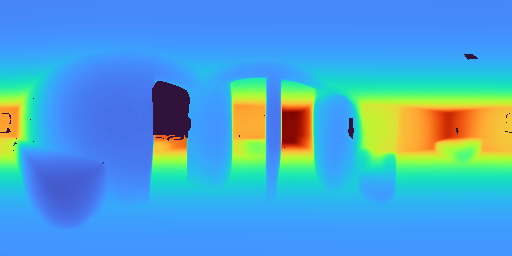}}
\subfloat{\includegraphics[width=0.2\linewidth]{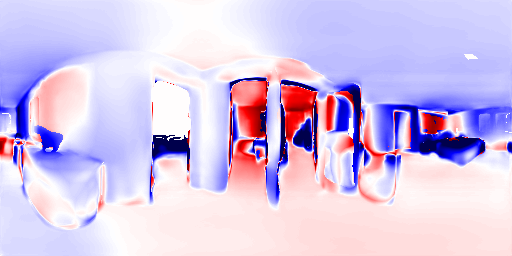}}
\\[-2.5ex]

\setcounter{subfigure}{0}%

\subfloat[Color]{\includegraphics[width=0.2\linewidth]{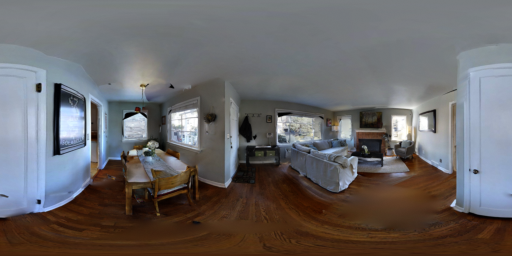}}
\subfloat[Ground Truth Depth]{\includegraphics[width=0.2\linewidth]{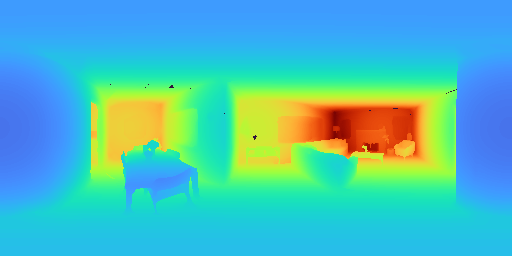}}
\subfloat[Unet Depth]{\includegraphics[width=0.2\linewidth]{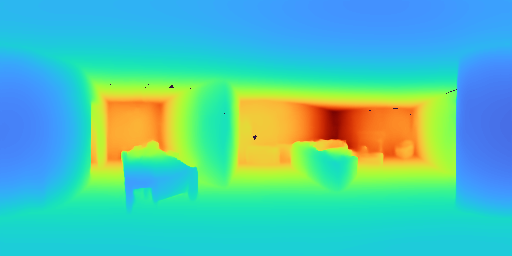}}
\subfloat[Pnas Depth]{\includegraphics[width=0.2\linewidth]{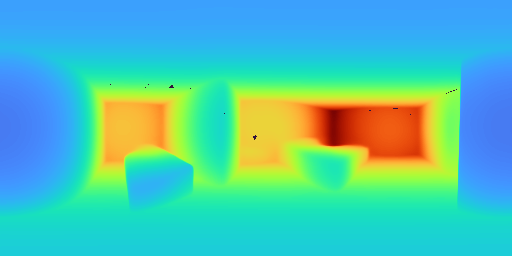}}
\subfloat[Depth Advantage]{\includegraphics[width=0.2\linewidth]{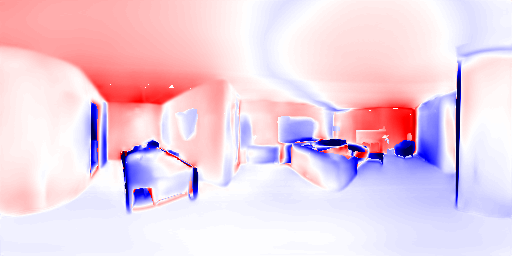}}

\caption{
Qualitative comparison between the UNet and Pnas architectures. 
On the right the advantage visualization shows with \textcolor{blue}{blue} color the areas where the former performs better, and with \textcolor{red}{red} color the areas where the latter performs better.
The color magnitude corresponds to the MAE difference between the two models, illustrating the performance deviation between the two models.
Pnas provides smoother results while it is clear that UNet is able to capture finer-grained details.
}

\label{fig:adv_unet_pnas}

\end{figure*}

%% file: Figures/adv_unet_reskip.tex
\begin{figure*}[!htbp]

\centering

\subfloat{\includegraphics[width=0.2\linewidth]{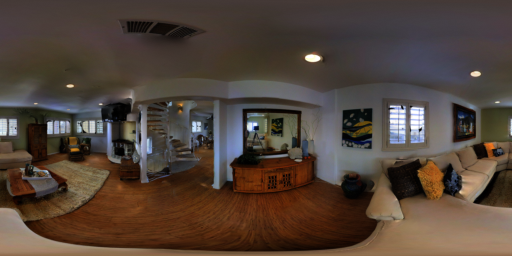}}
\subfloat{\includegraphics[width=0.2\linewidth]{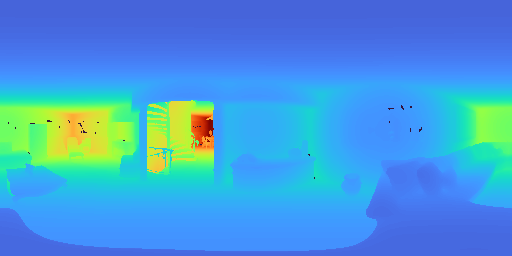}}
\subfloat{\includegraphics[width=0.2\linewidth]{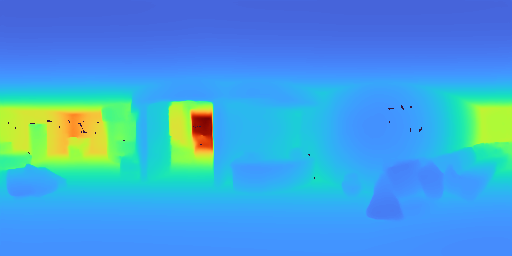}}
\subfloat{\includegraphics[width=0.2\linewidth]{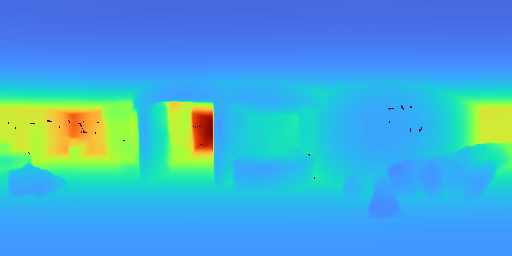}}
\subfloat{\includegraphics[width=0.2\linewidth]{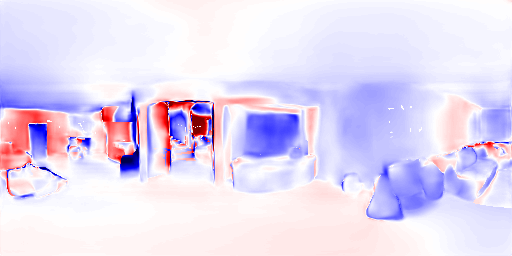}}
\\[-2.5ex]

\subfloat{\includegraphics[width=0.2\linewidth]{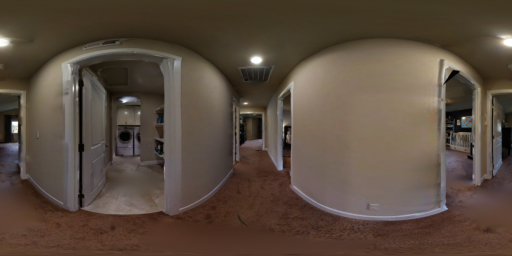}}
\subfloat{\includegraphics[width=0.2\linewidth]{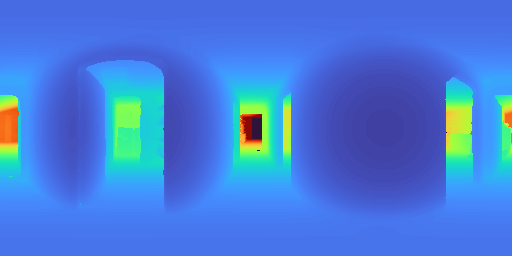}}
\subfloat{\includegraphics[width=0.2\linewidth]{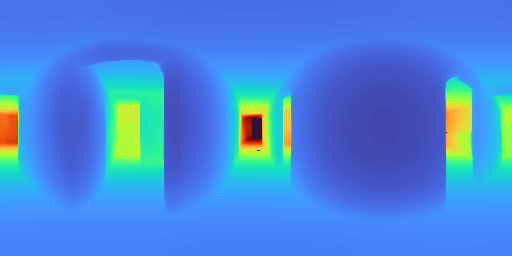}}
\subfloat{\includegraphics[width=0.2\linewidth]{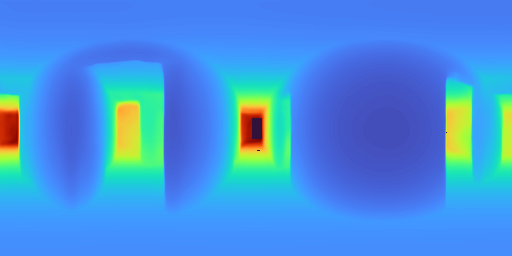}}
\subfloat{\includegraphics[width=0.2\linewidth]{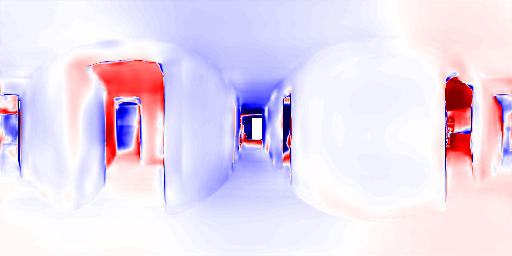}}
\\[-2.5ex]

\subfloat{\includegraphics[width=0.2\linewidth]{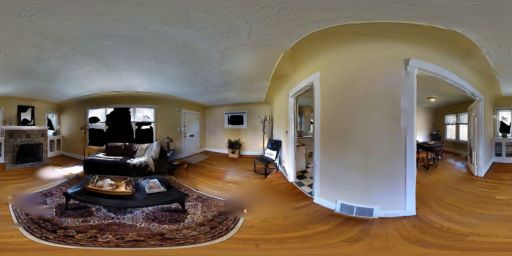}}
\subfloat{\includegraphics[width=0.2\linewidth]{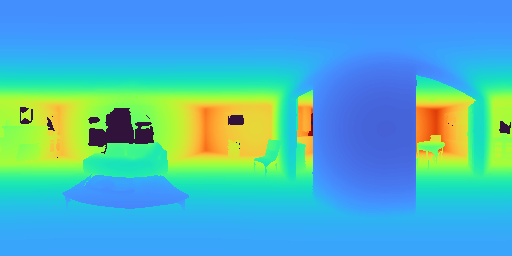}}
\subfloat{\includegraphics[width=0.2\linewidth]{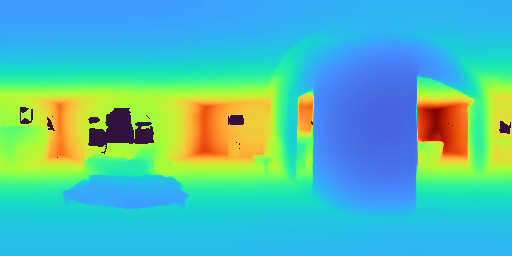}}
\subfloat{\includegraphics[width=0.2\linewidth]{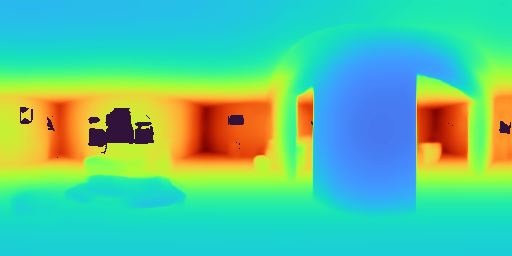}}
\subfloat{\includegraphics[width=0.2\linewidth]{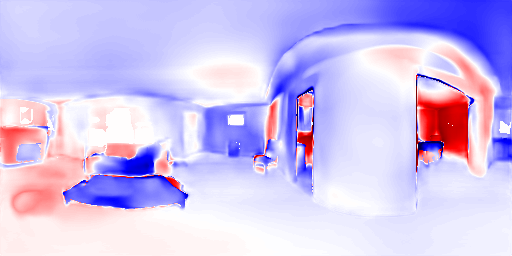}}
\\[-2.5ex]

\subfloat{\includegraphics[width=0.2\linewidth]{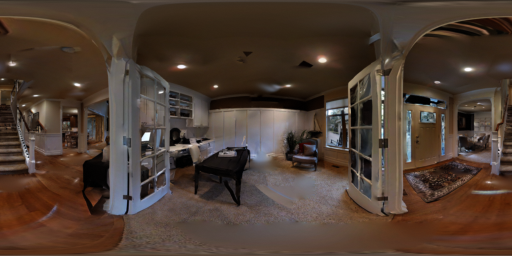}}
\subfloat{\includegraphics[width=0.2\linewidth]{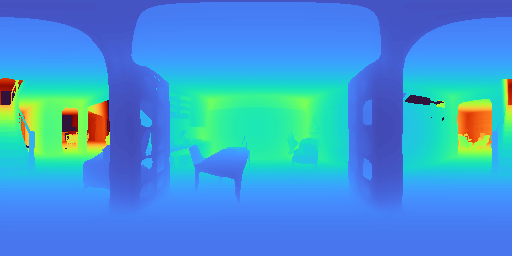}}
\subfloat{\includegraphics[width=0.2\linewidth]{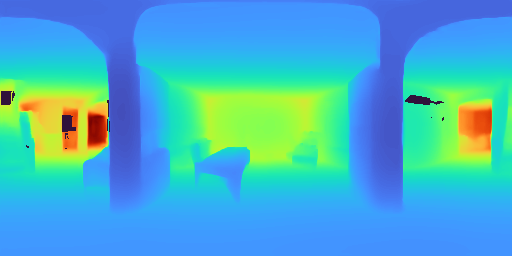}}
\subfloat{\includegraphics[width=0.2\linewidth]{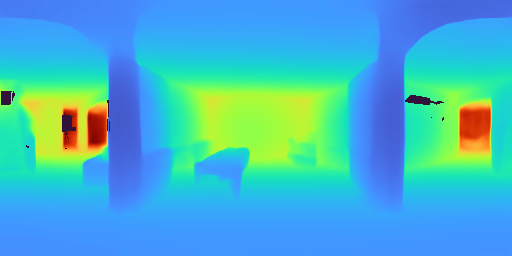}}
\subfloat{\includegraphics[width=0.2\linewidth]{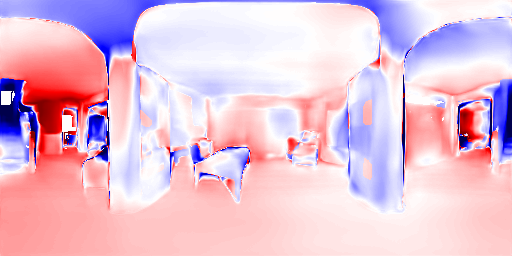}}
\\[-2.5ex]

\setcounter{subfigure}{0}%

\subfloat[Color]{\includegraphics[width=0.2\linewidth]{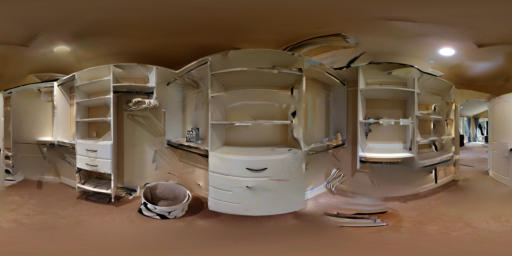}}
\subfloat[Ground Truth Depth]{\includegraphics[width=0.2\linewidth]{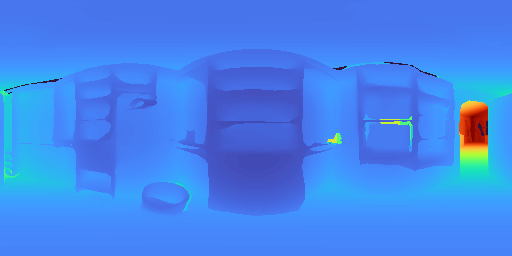}}
\subfloat[Unet Depth]{\includegraphics[width=0.2\linewidth]{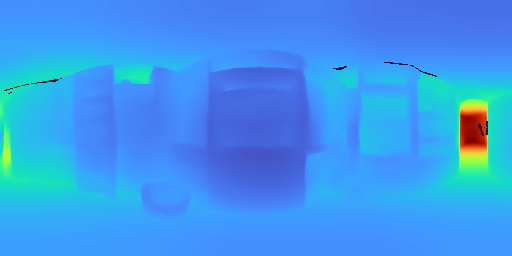}}
\subfloat[ResNet\textsubscript{skip} Depth]{\includegraphics[width=0.2\linewidth]{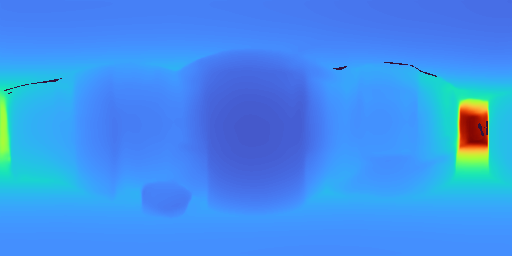}}
\subfloat[Depth Advantage]{\includegraphics[width=0.2\linewidth]{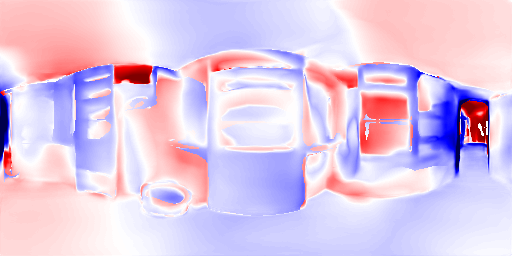}}

\caption{
Qualitative comparison between the UNet and ResNet\textsubscript{skip} architectures.
On the right the advantage visualization shows with \textcolor{blue}{blue} color the areas where the former performs better, and with \textcolor{red}{red} color the areas where the latter performs better.
The color magnitude corresponds to the MAE difference between the two models, illustrating the performance deviation between the two models.
}

\label{fig:adv_unet_reskip}

\end{figure*}

%% file: Figures/supp/edges/edgescomparisonupdt.tex
\begin{figure*}[!htbp]
  \def\mycolspace{0.5mm}
  \def\datasetheight{1.8cm}
  \centering
    \begin{tabular}{@{}c@{\hspace{\mycolspace}}c@{\hspace{\mycolspace}}c@{\hspace{\mycolspace}}c@{\hspace{\mycolspace}}c@{}}
    \includegraphics[height=\datasetheight]{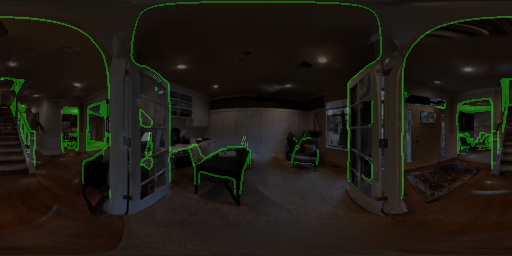} & 
    \includegraphics[height=\datasetheight]{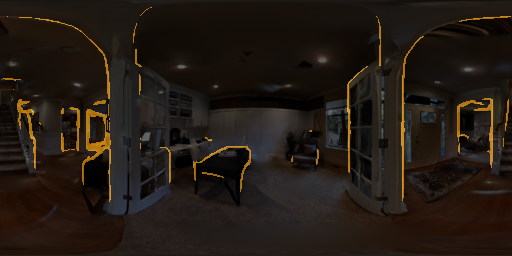} & 
    \includegraphics[height=\datasetheight]{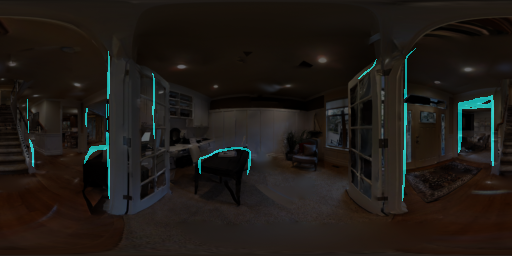} & 
    \includegraphics[height=\datasetheight]{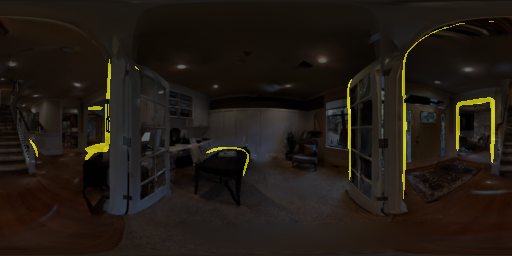}  & \includegraphics[height=\datasetheight]{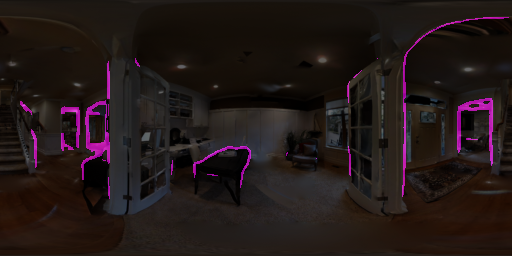}  
    \\
    \includegraphics[height=\datasetheight]{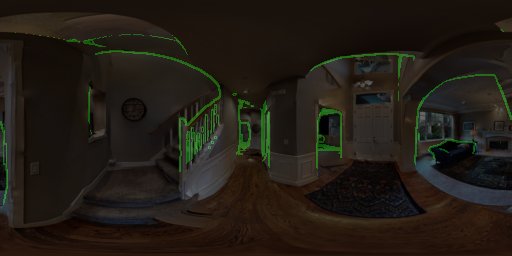} & 
    \includegraphics[height=\datasetheight]{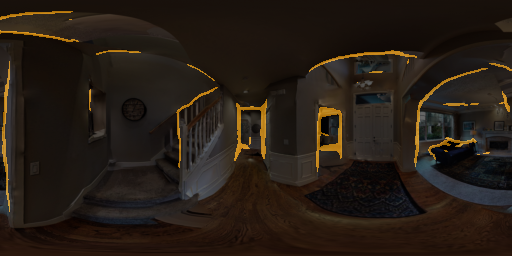} & 
    \includegraphics[height=\datasetheight]{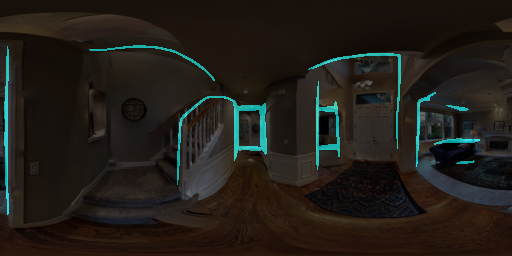} & 
    \includegraphics[height=\datasetheight]{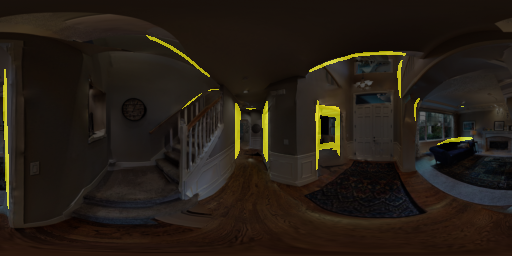}  & \includegraphics[height=\datasetheight]{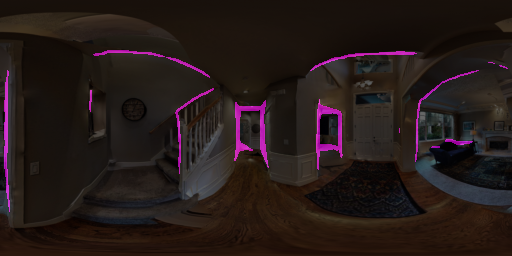}  
    \\
    \includegraphics[height=\datasetheight]{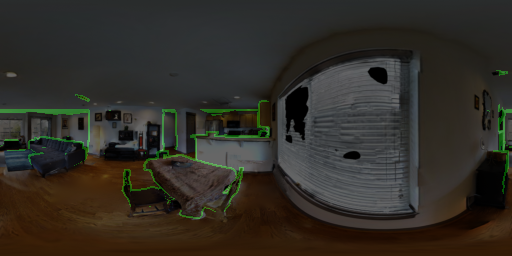} & 
    \includegraphics[height=\datasetheight]{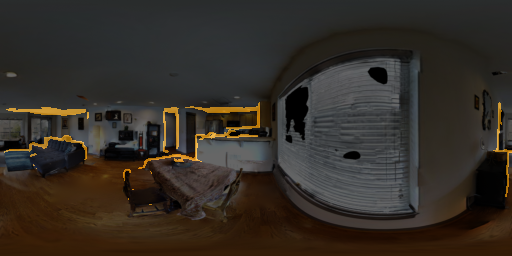} & 
    \includegraphics[height=\datasetheight]{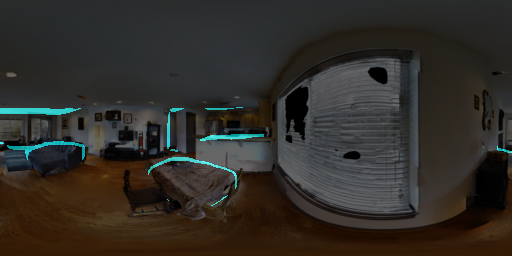} & 
    \includegraphics[height=\datasetheight]{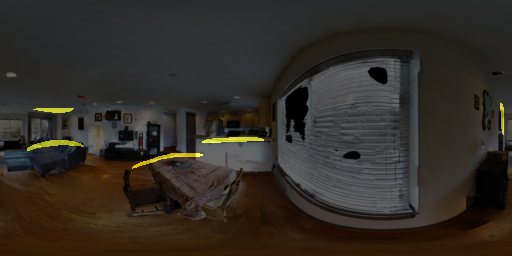}  & \includegraphics[height=\datasetheight]{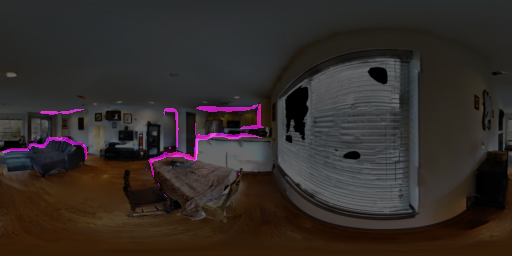}  
    \\
    \includegraphics[height=\datasetheight]{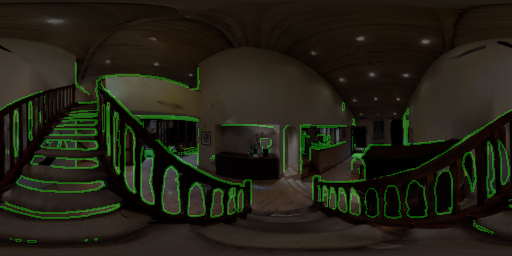} & 
    \includegraphics[height=\datasetheight]{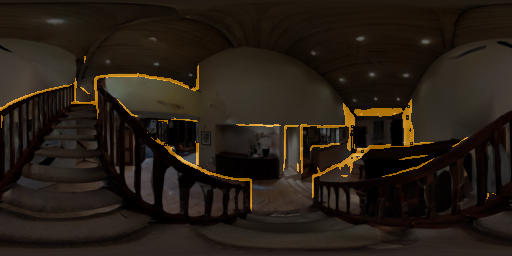} & 
    \includegraphics[height=\datasetheight]{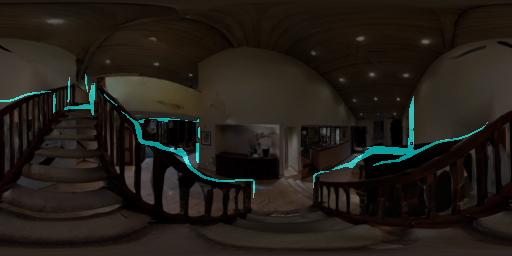} & 
    \includegraphics[height=\datasetheight]{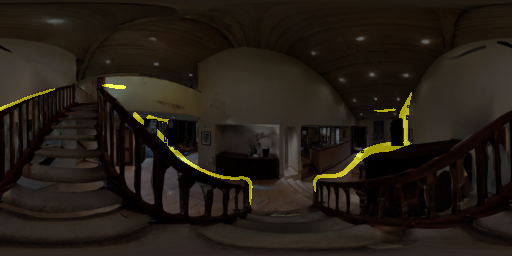}  & \includegraphics[height=\datasetheight]{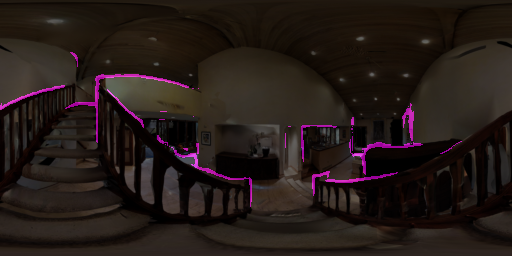}  
    \\
    \includegraphics[height=\datasetheight]{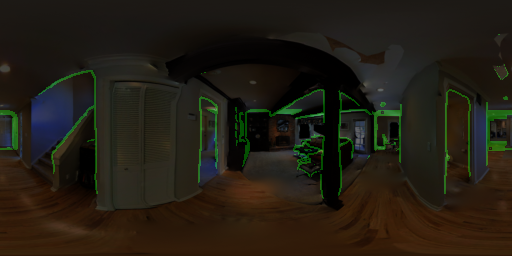} & 
    \includegraphics[height=\datasetheight]{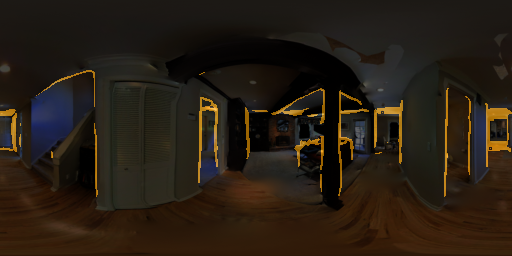} & 
    \includegraphics[height=\datasetheight]{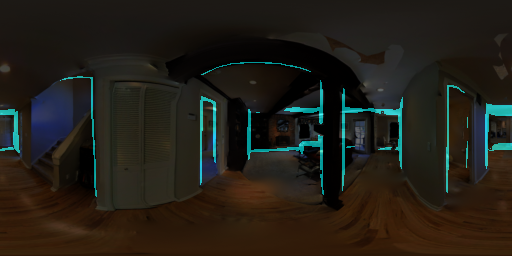} & 
    \includegraphics[height=\datasetheight]{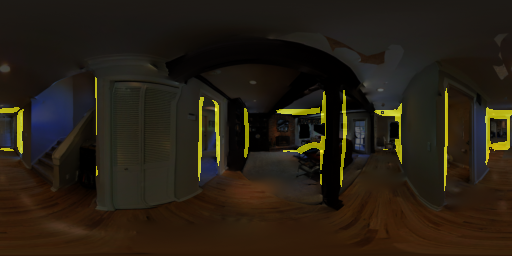}  & \includegraphics[height=\datasetheight]{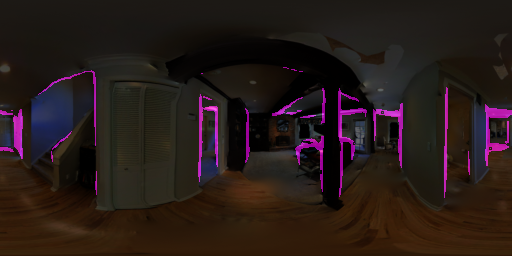}  
     \\
    \end{tabular}

\caption{
Boundary preservation qualitative comparison between the UNet, Pnas, ResNet and ResNet\textsubscript{skip} models.
Boundaries are extracted by applying a Canny edge detector \cite{canny1986computational} with predefined thresholds on normalized predicted depth maps, and then are blended with the original color panorama.
From left to right: \textbf{i)} GT depth (\textcolor{caribbeangreen}{green}), \textbf{ii)} UNet (\textcolor{chromeyellow}{orange}), \textbf{iii)} Pnas (\textcolor{cyan}{cyan}), \textbf{iv)} ResNet (\textcolor{citrine}{yellow}), and \textbf{v)} ResNet\textsubscript{skip} (\textcolor{magenta}{magenta}).
}

\label{fig:edges}

\end{figure*}

%% file: Figures/supp/unet_mesh.tex
\begin{figure*}[!htbp]

\centering

\subfloat{\includegraphics[width=0.24\linewidth]{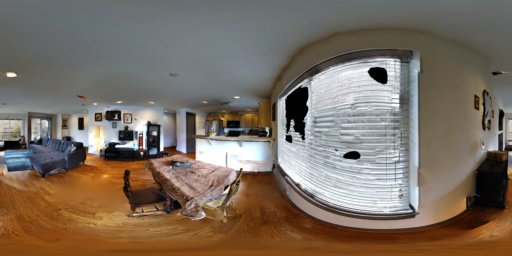}}
\subfloat{\includegraphics[width=0.24\linewidth]{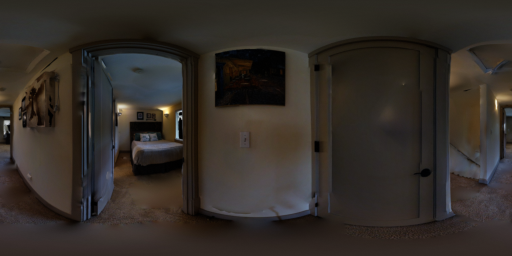}}
\subfloat{\includegraphics[width=0.24\linewidth]{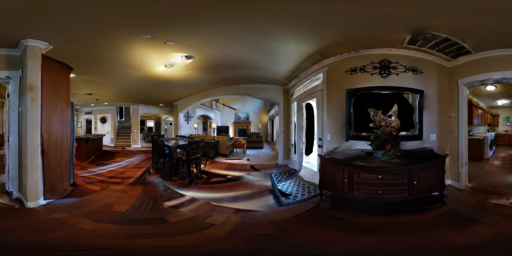}}
\subfloat{\includegraphics[width=0.24\linewidth]{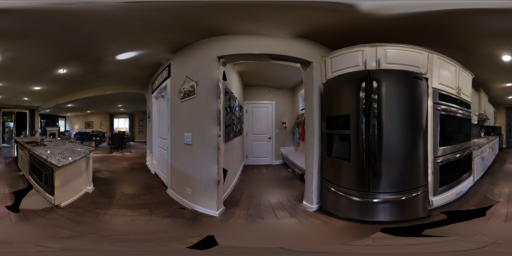}}

\subfloat{\includegraphics[width=0.24\linewidth]{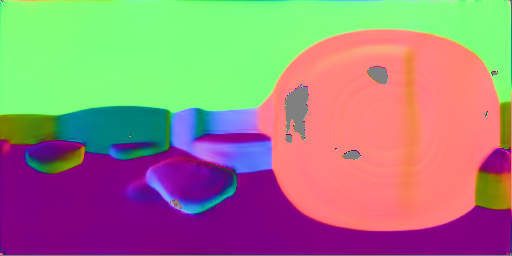}}
\subfloat{\includegraphics[width=0.24\linewidth]{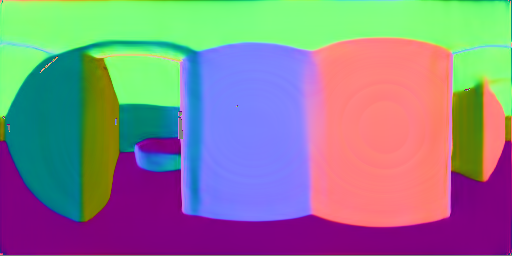}}
\subfloat{\includegraphics[width=0.24\linewidth]{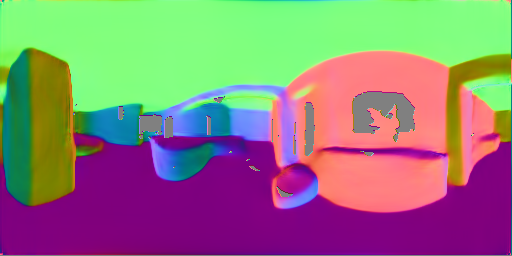}}
\subfloat{\includegraphics[width=0.24\linewidth]{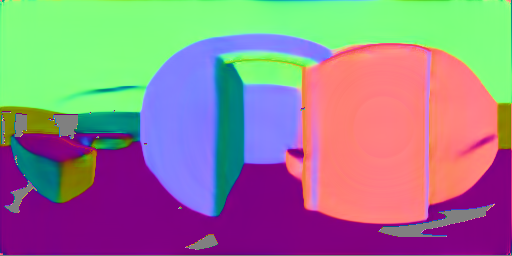}}

\subfloat{\includegraphics[width=0.24\linewidth]{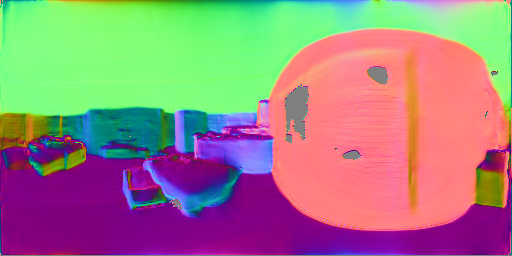}}
\subfloat{\includegraphics[width=0.24\linewidth]{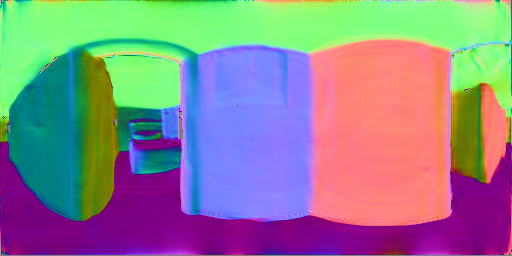}}
\subfloat{\includegraphics[width=0.24\linewidth]{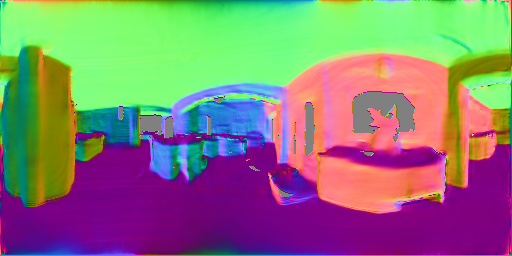}}
\subfloat{\includegraphics[width=0.24\linewidth]{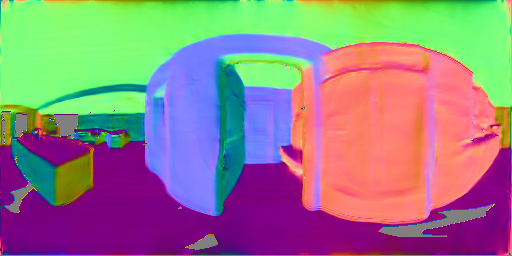}}

\subfloat{\includegraphics[width=0.24\linewidth]{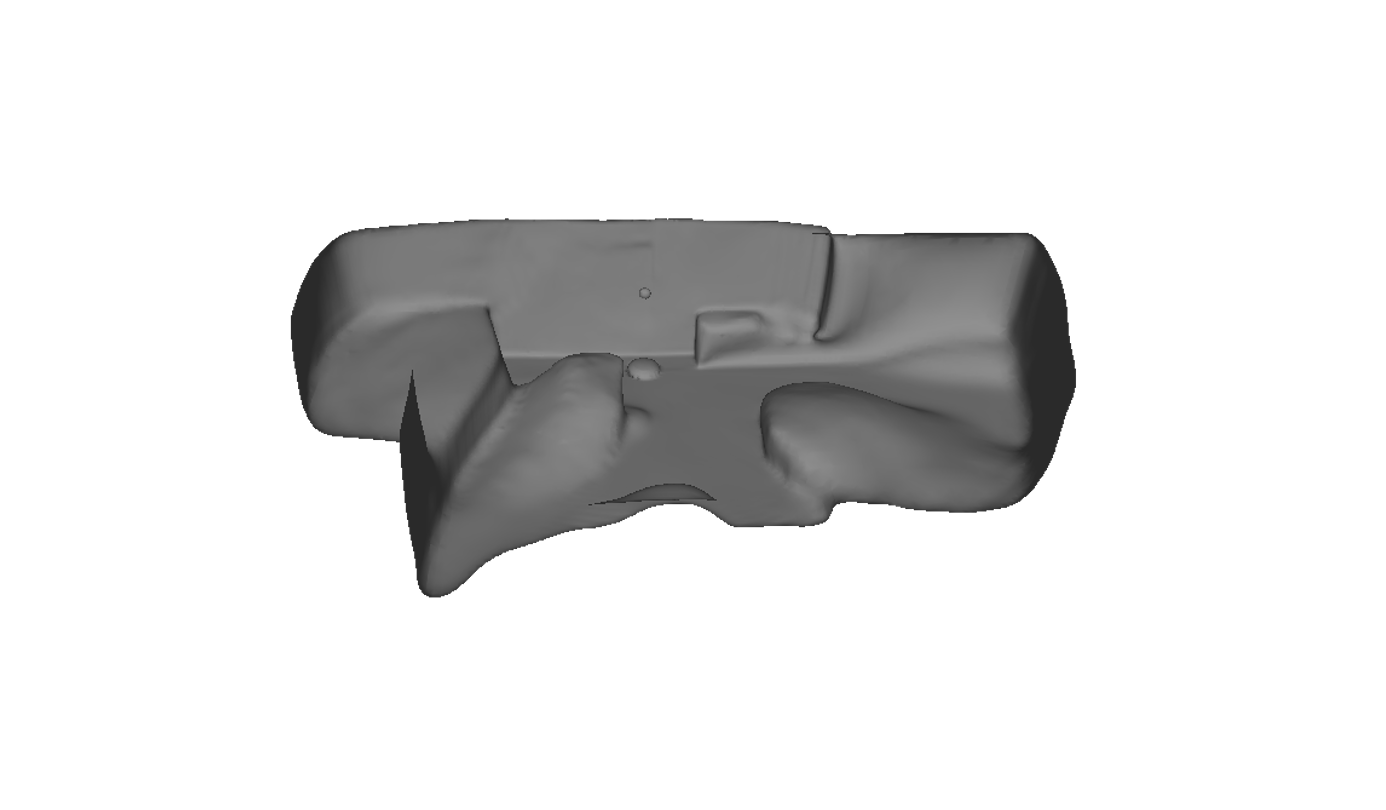}}
\subfloat{\includegraphics[width=0.24\linewidth]{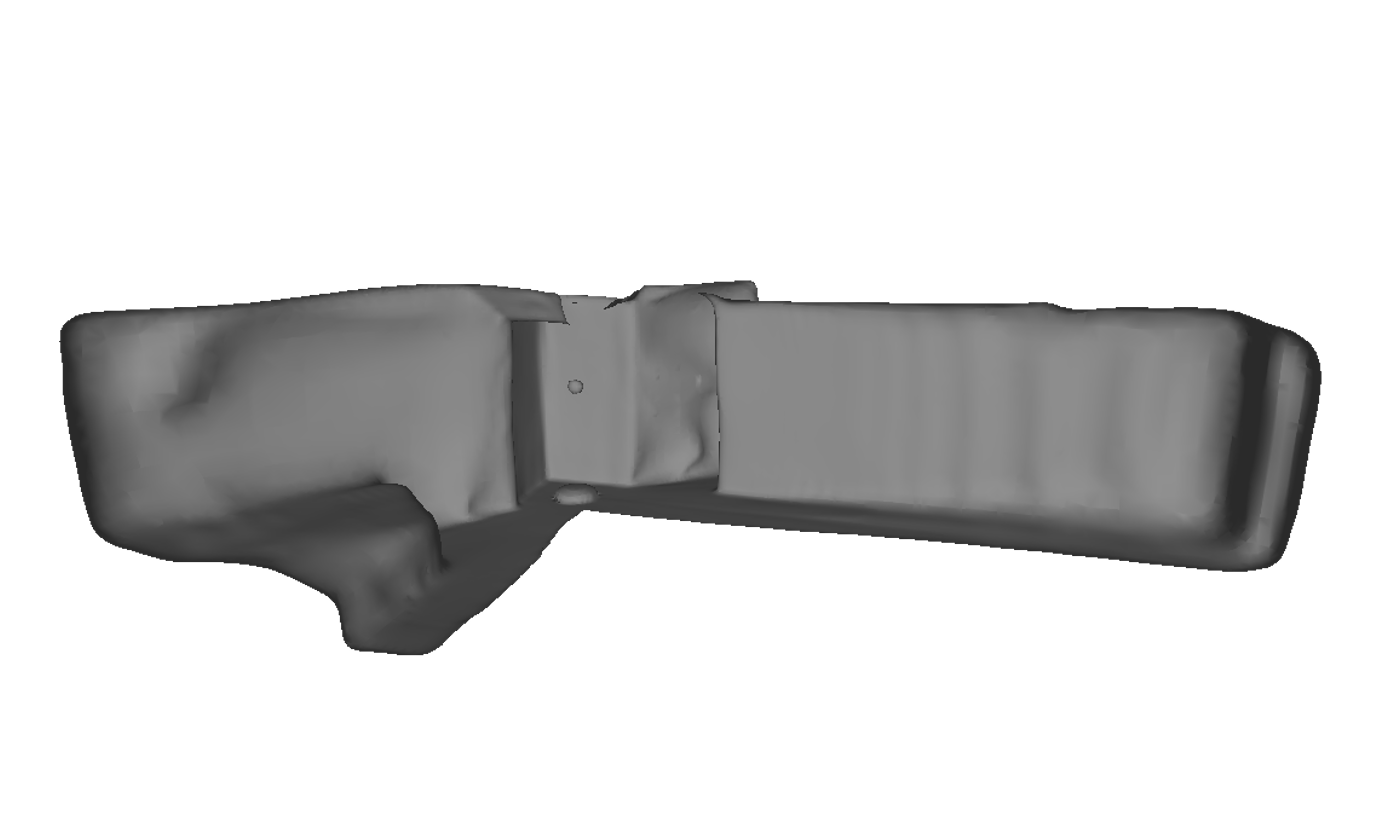}}
\subfloat{\includegraphics[width=0.24\linewidth]{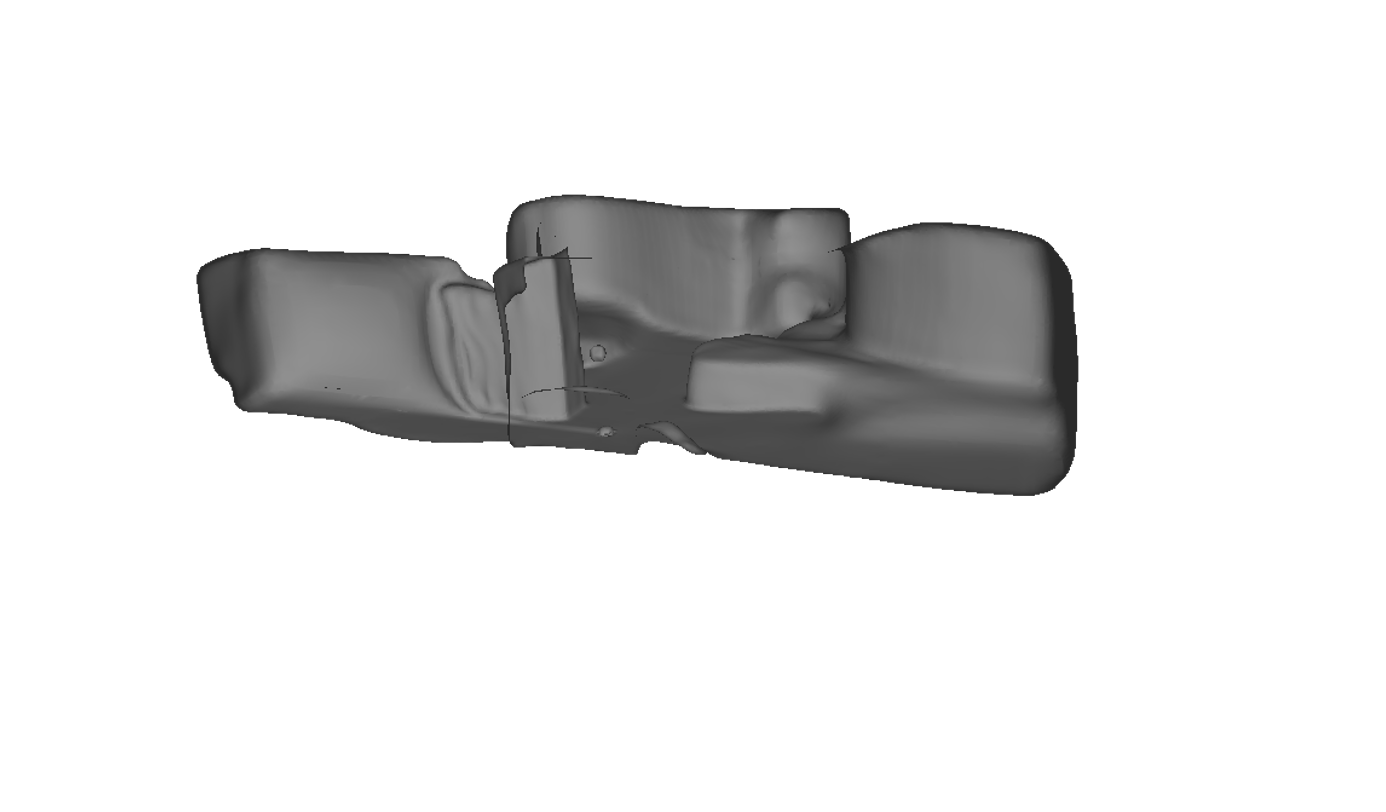}}
\subfloat{\includegraphics[width=0.24\linewidth]{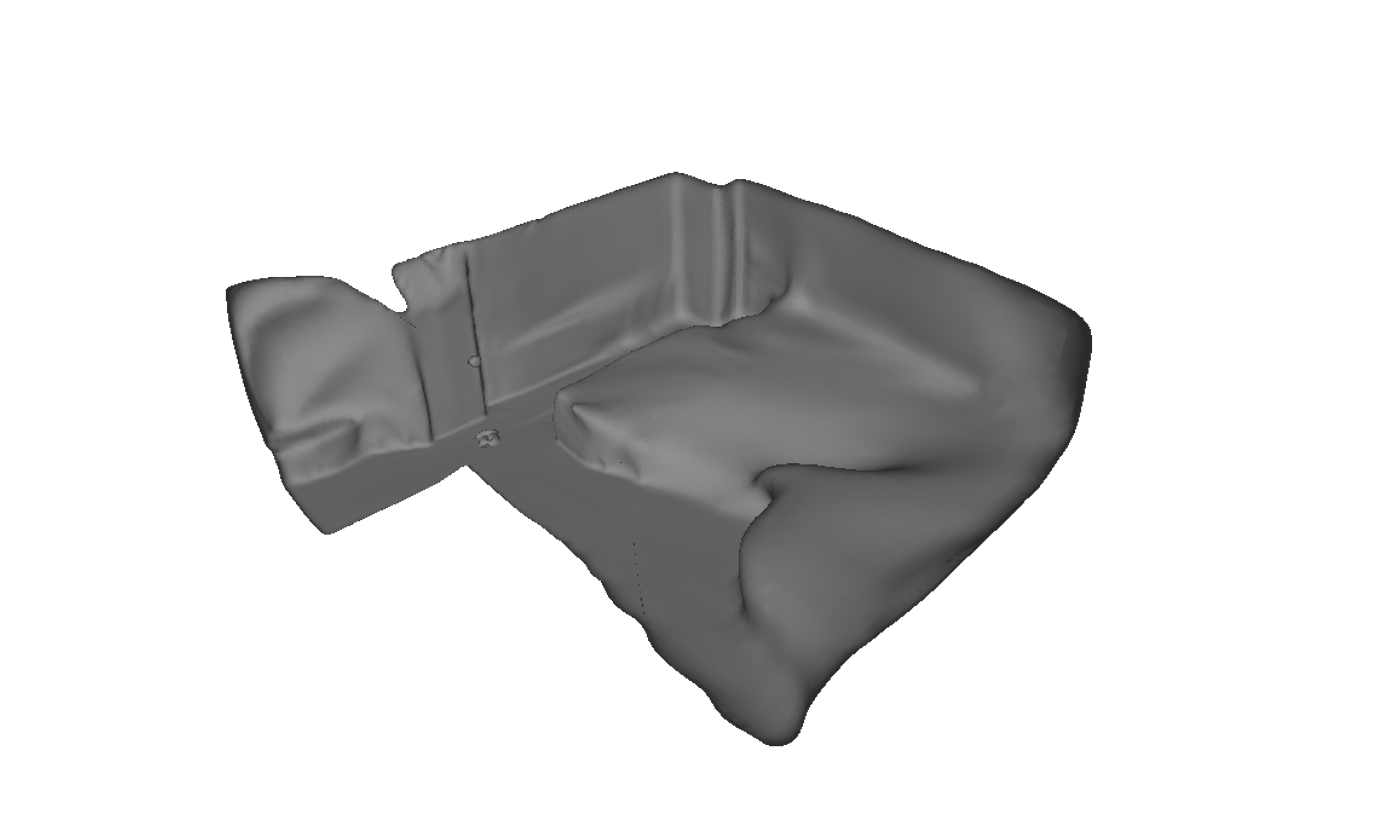}}

\subfloat{\includegraphics[width=0.24\linewidth]{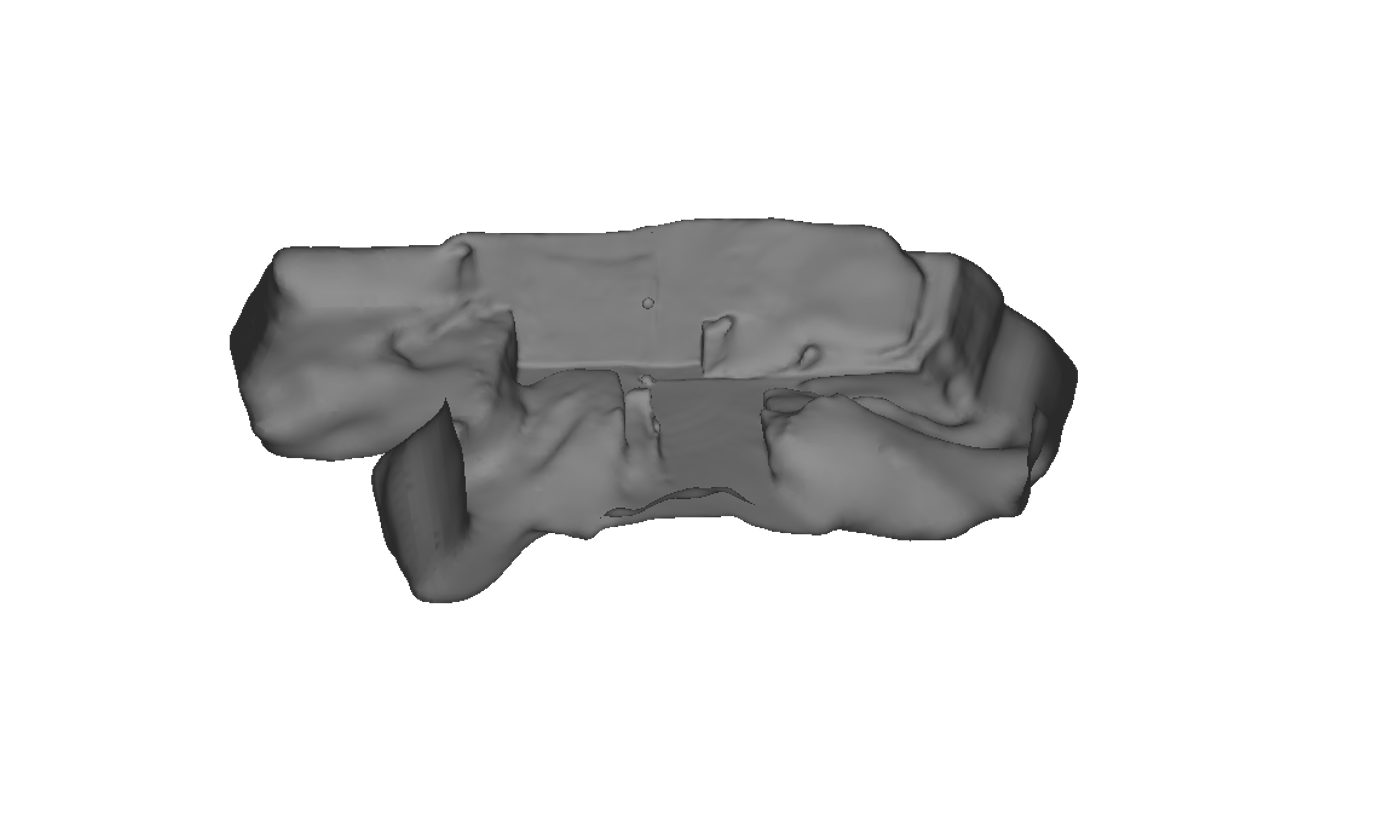}}
\subfloat{\includegraphics[width=0.24\linewidth]{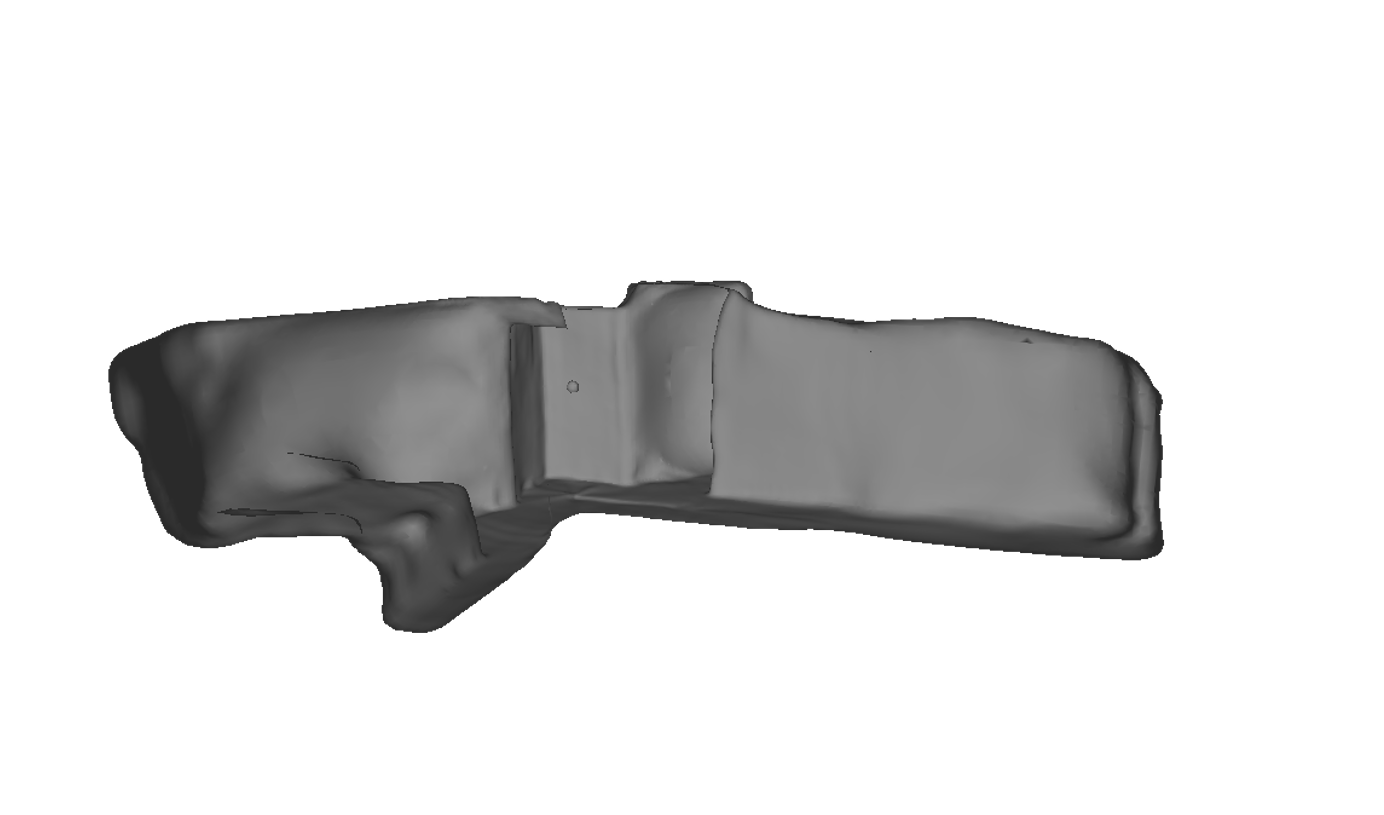}}
\subfloat{\includegraphics[width=0.24\linewidth]{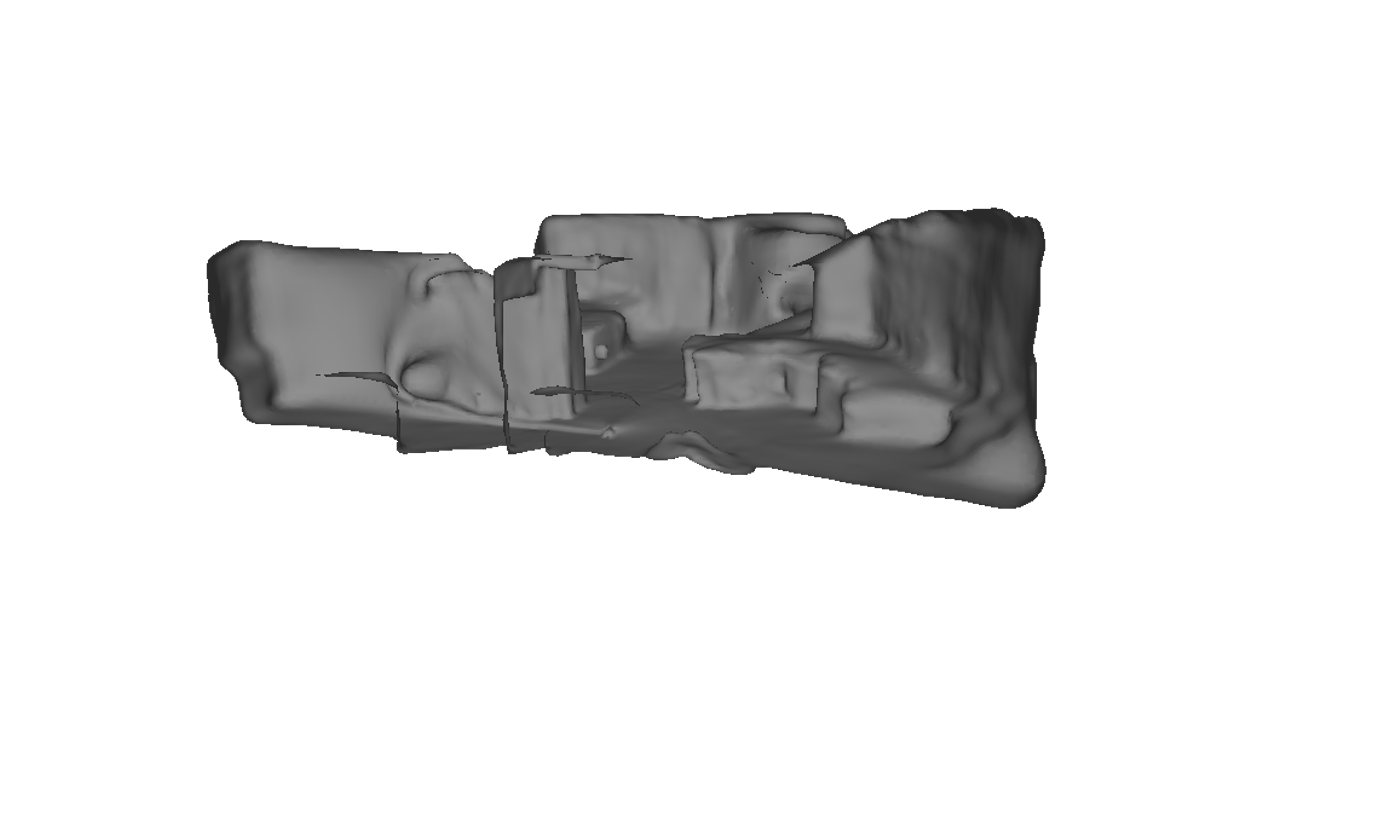}}
\subfloat{\includegraphics[width=0.24\linewidth]{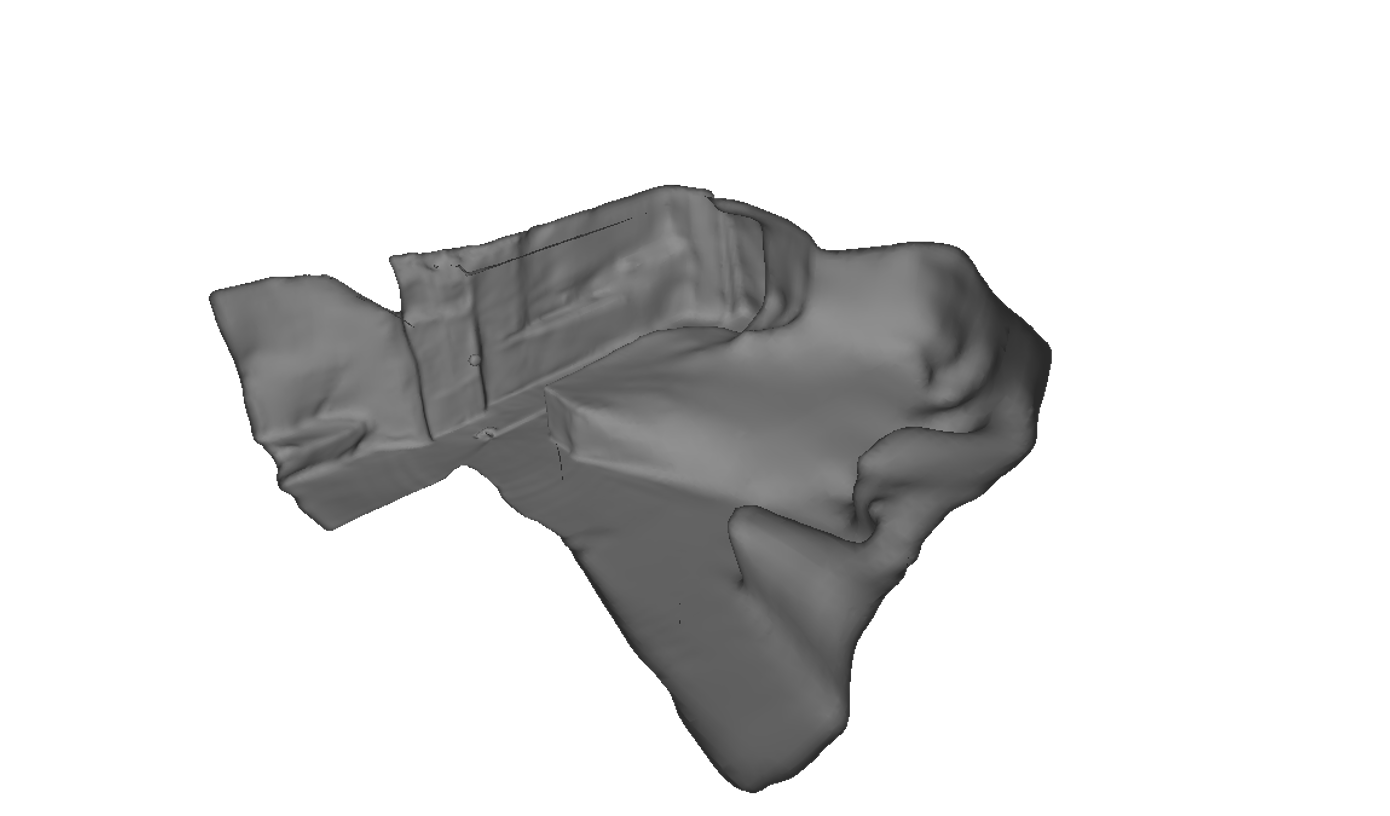}}

\subfloat{\includegraphics[width=0.24\linewidth]{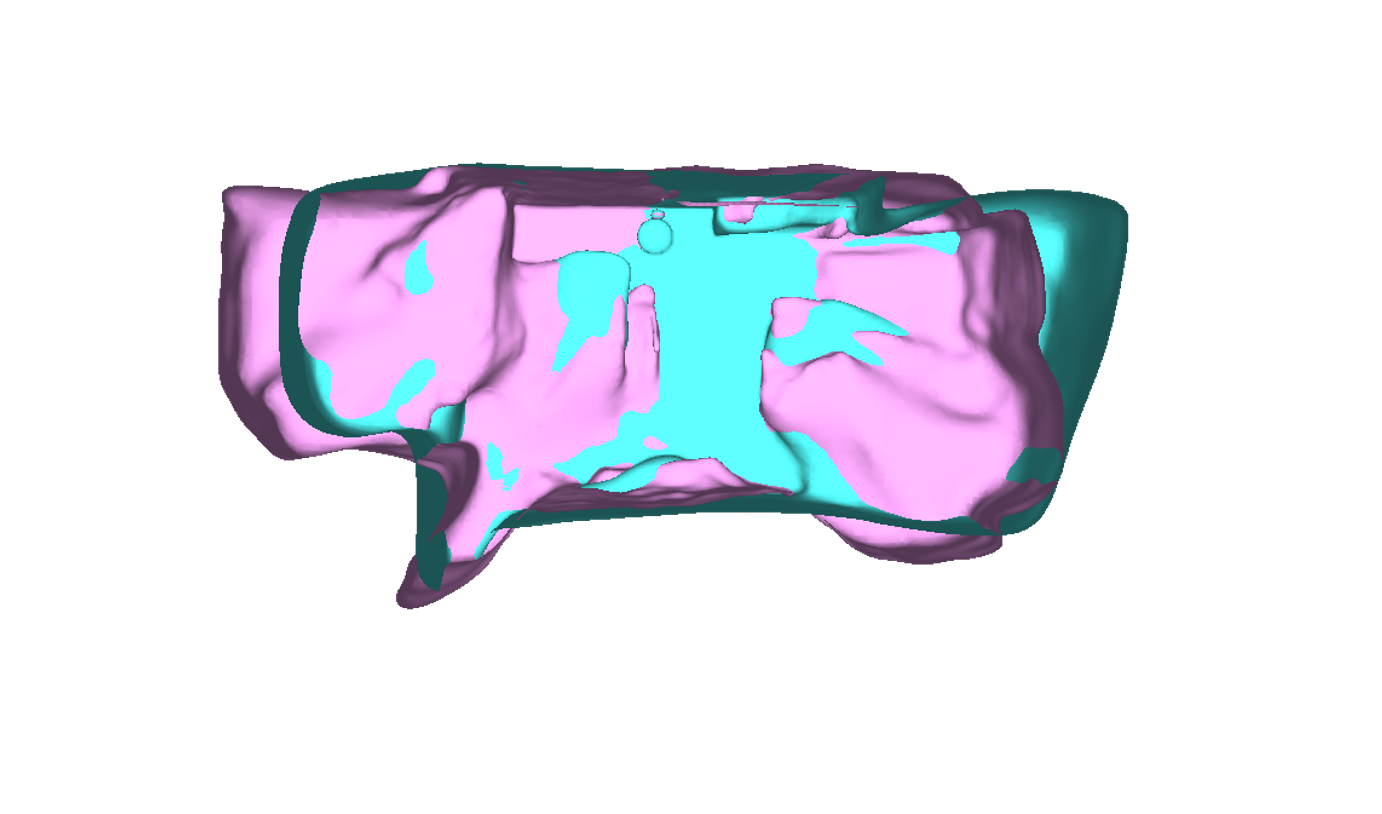}}
\subfloat{\includegraphics[width=0.24\linewidth]{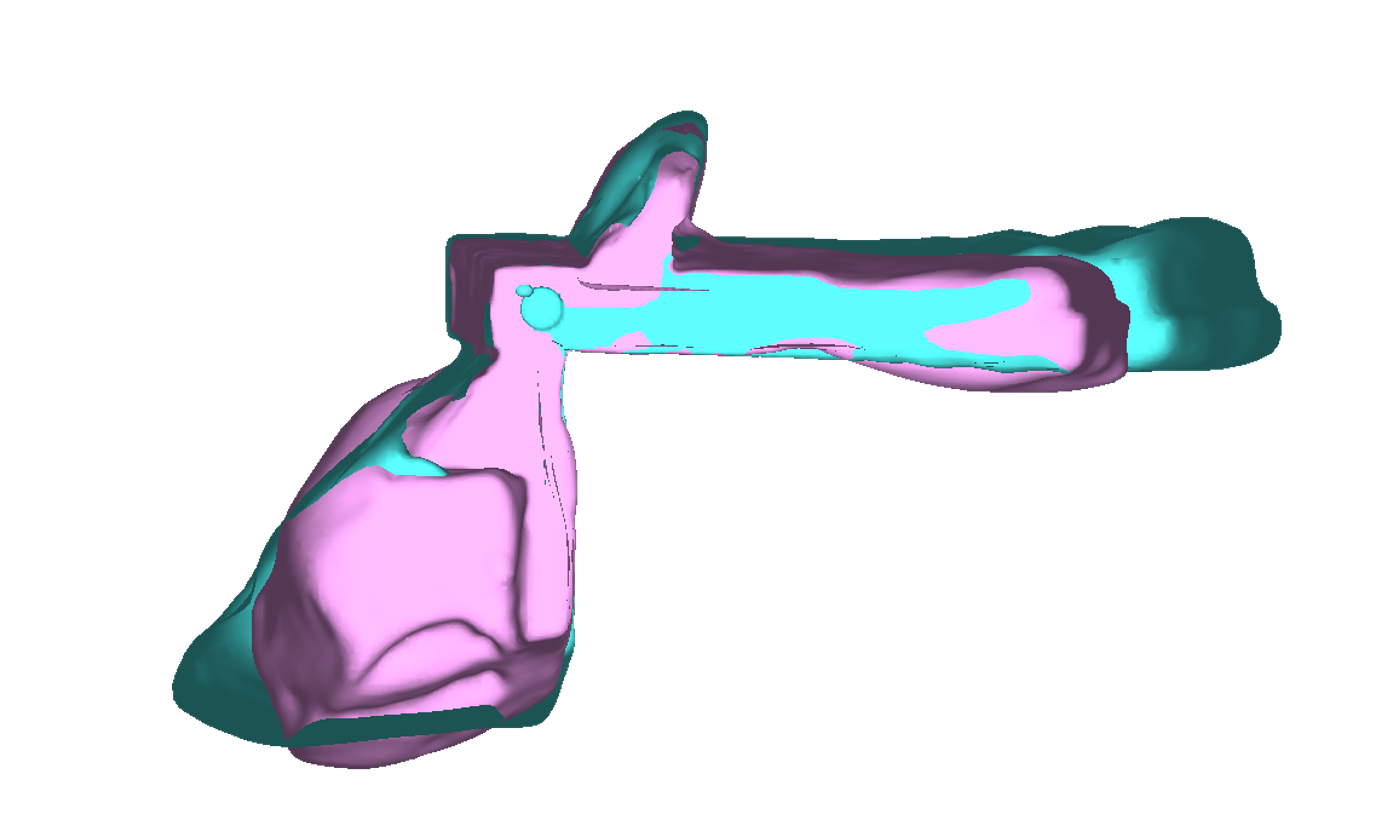}}
\subfloat{\includegraphics[width=0.24\linewidth]{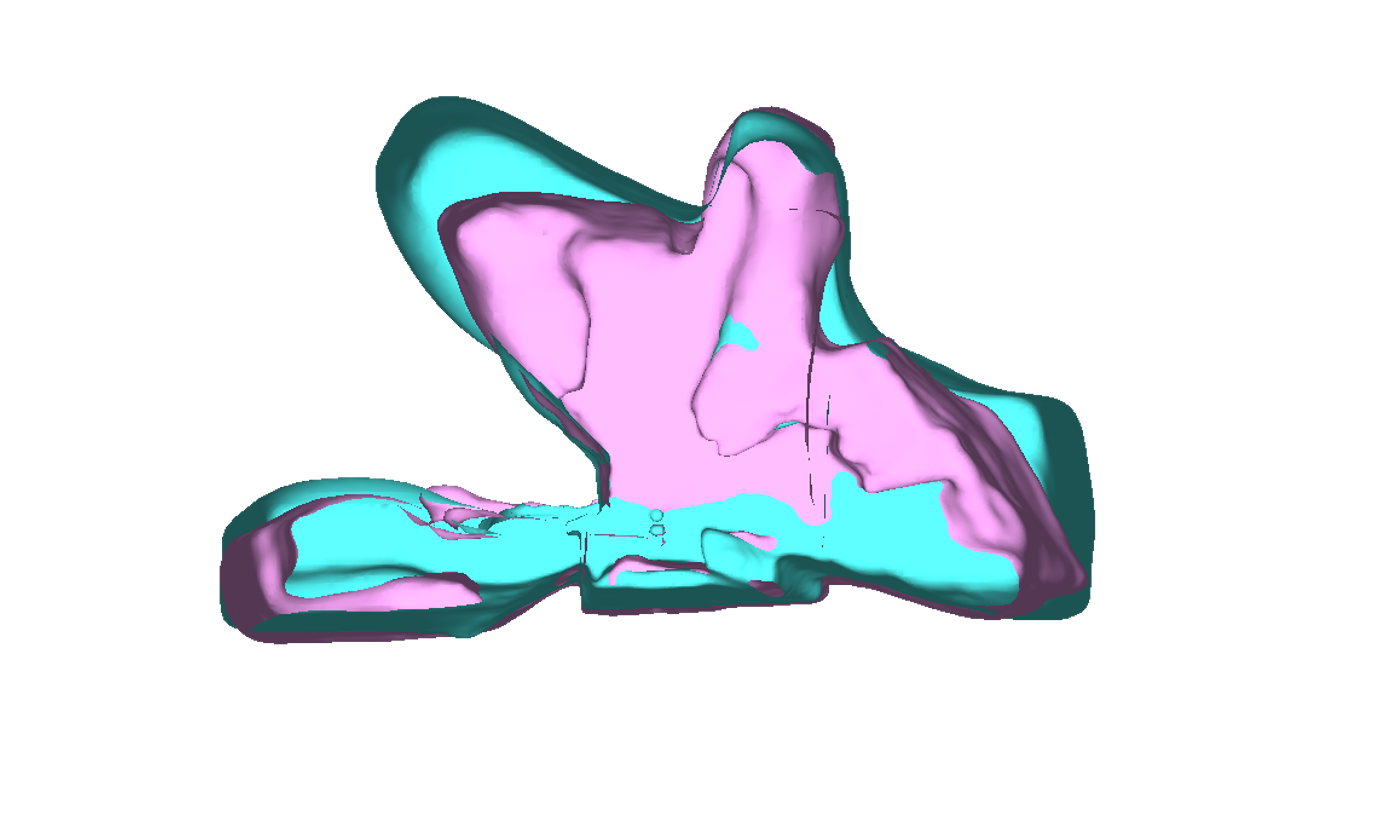}}
\subfloat{\includegraphics[width=0.24\linewidth]{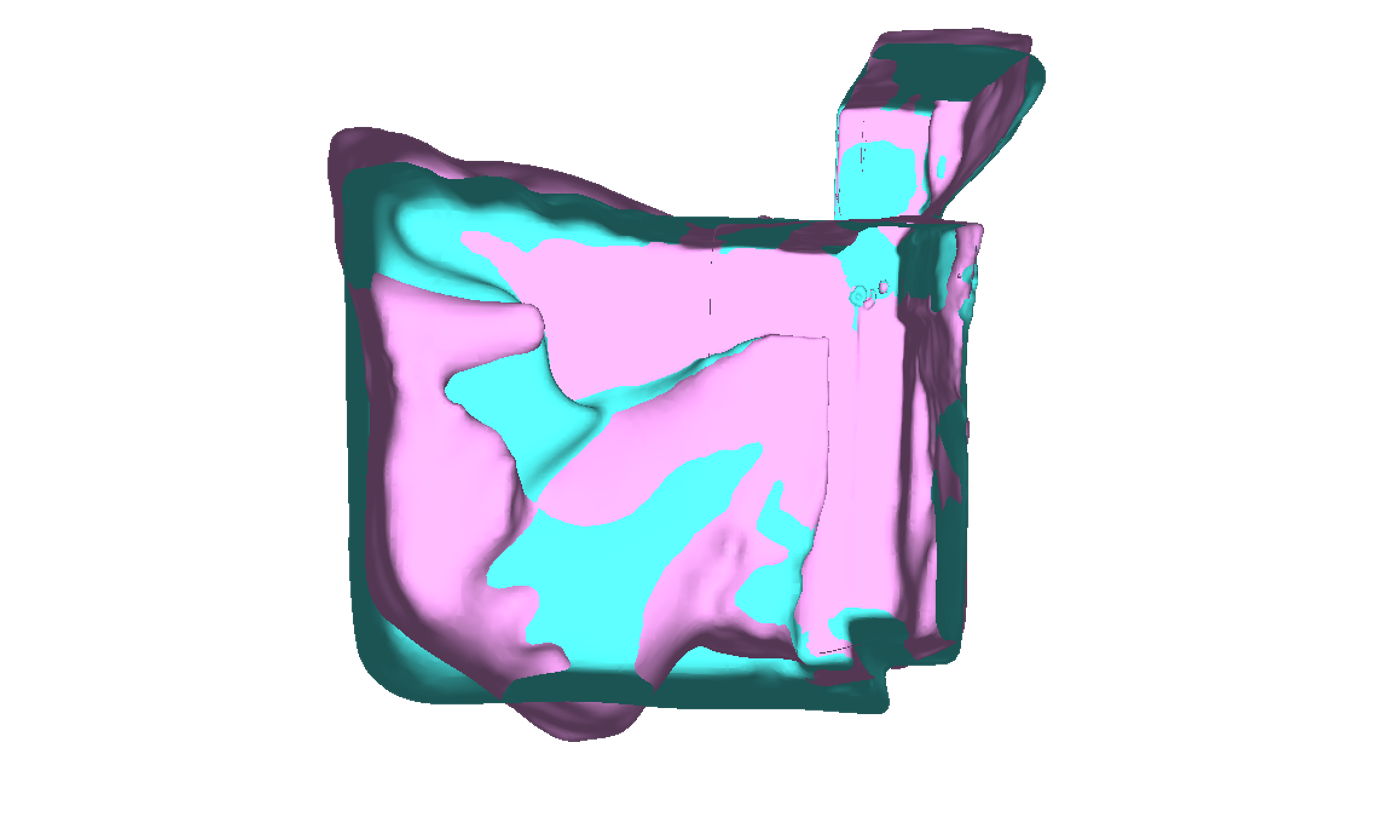}}

\setcounter{subfigure}{0}%

\caption{
Qualitative comparison of the Pnas and UNet models in surface reconstruction.
From top to bottom: \textbf{i)} input color panorama, \textbf{ii)} Pnas normal map from the estimated depth map, \textbf{iii)} UNet normal map, \textbf{iv)} Pnas Screened Poisson Surface Reconstruction \cite{kazhdan2013screened} 3D surface reconstruction, \textbf{v)} UNet 3D surface reconstruction, \textbf{vi)} overlaid Pnas (\textcolor{cyan}{cyan}) and UNet (\textcolor{candypink}{pink}) 3D surface reconstructions from birds eye view.
}
\label{fig:unet_mesh}
\end{figure*}

%% file: Figures/supp/inthewild.tex
\begin{figure*}[!htbp]

\centering

\includegraphics[]{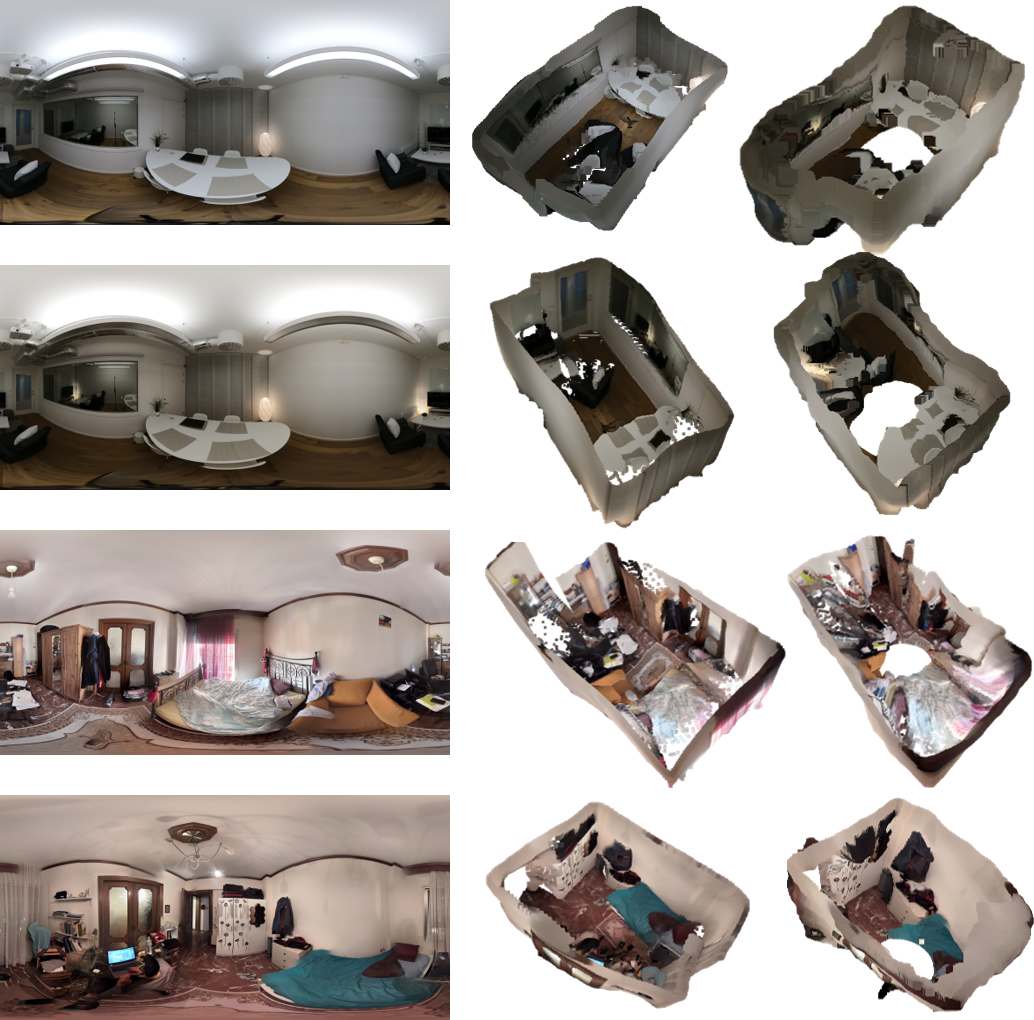}

\caption{
Qualitative results using in-the-wild data. 
On the left the input color panoramas are depicted.
The two top rows are captured with a \360 camera, while the bottom two rows are stitched panoramas from a mobile phone.
The colored point clouds of the predicted depths from our UNet model (middle) and BiFuse \cite{wang2020bifuse} (right).
Ceilings have been removed for visualization purposes. 
}

\label{fig:inthewild}

\end{figure*}